\documentclass[iicol]{sn-jnl}

\usepackage[dvipsnames,table,xcdraw]{xcolor}
\usepackage{color}
\usepackage{pgfplots}
\usepackage{adjustbox}
\usepackage{tikz}
\usetikzlibrary{patterns,shapes.arrows}
\usetikzlibrary{shapes, arrows.meta, positioning}

\usepgfplotslibrary{groupplots}
\usepgfplotslibrary{fillbetween}
\usetikzlibrary{arrows,decorations.pathmorphing,positioning,fit,trees,shapes,shadows,automata,calc} 
\usetikzlibrary{patterns,arrows,arrows.meta,calc,shapes,shadows,decorations.pathmorphing,decorations.pathreplacing,automata,shapes.multipart,positioning,shapes.geometric,fit,circuits,trees,shapes.gates.logic.US,fit, matrix,arrows.meta, quotes}
\usetikzlibrary{backgrounds,scopes}% \usetikzlibrary{shapes.geometric, arrows.meta, positioning, fit, backgrounds}
\usepackage{colortbl}
\usepackage{mathrsfs}
\usepackage[nocompress]{cite}
\newcolumntype{a}{>{\columncolor{green}}c}
\newcolumntype{b}{>{\columncolor{green!65}}c}
\newcolumntype{d}{>{\columncolor{green!40}}c}
\newcolumntype{e}{>{\columncolor{green!15}}c}
\usepackage{xspace}
\usepackage{multicol,multirow}

\usepackage{amsthm}
\usepackage{soul}
\usepackage[]{mdframed}
\usepackage{threeparttable}
\usepackage{pifont}% http://ctan.org/pkg/pifont
\newcommand{\cmark}{\ding{51}}%
\usepackage{wrapfig}
\usepackage{amsmath, amsfonts, amssymb}   
\usepackage{mathtools}
\usepackage{pifont}% 
\newcommand{\xmark}{\ding{55}}%
\let\labelindent\relax
\usepackage{enumitem}
\usepackage{flushend}
\usepackage{booktabs}
\usepackage{float}
\usepackage[linesnumbered,ruled,vlined]{algorithm2e}
\usepackage{bm}
\usepackage{apxproof}
\usepackage{subcaption}
\usepackage[font=small]{caption}
\newtheorem{theorem}{Theorem}

\newtheorem{corollary}{Corollary}
\newtheorem{lemma}[theorem]{Lemma}
\newtheorem{definition}{Definition}

\usepackage{soul}
\definecolor{sg}{HTML}{00ff7f}
\definecolor{lb}{HTML}{b0e0e6} % Adjusted to a recognizable code
\definecolor{lg}{HTML}{9bfaa8}
\begin{document}
\title[Deadlock-free, Safe, and Decentralized Multi-Robot Navigation in Social Mini-Games via Discrete-Time Control Barrier Functions]{Deadlock-free, Safe, and Decentralized Multi-Robot Navigation in Social Mini-Games via Discrete-Time Control Barrier Functions}

%%=============================================================%%
%% GivenName	-> \fnm{Joergen W.}
%% Particle	-> \spfx{van der} -> surname prefix
%% FamilyName	-> \sur{Ploeg}
%% Suffix	-> \sfx{IV}
%% \author*[1,2]{\fnm{Joergen W.} \spfx{van der} \sur{Ploeg} 
%%  \sfx{IV}}\email{iauthor@gmail.com}
%%=============================================================%%

\author*[1]{\fnm{Rohan} \sur{Chandra}}\email{rohanchandra@virginia.edu}

\author[2]{\fnm{Vrushabh} \sur{Zinage}}\email{vrushabh.zinage@utexas.edu}
% \equalcont{These authors contributed equally to this work.}
\author[2]{\fnm{Efstathios} \sur{Bakolas}}\email{bakolas@austin.utexas.edu}

\author[3,4]{\fnm{Peter} \sur{Stone}}\email{pstone@utexas.edu}

\author[3]{\fnm{Joydeep} \sur{Biswas}}\email{joydeepb@utexas.edu}

\affil*[1]{\orgdiv{Dept. of Computer Science}, \orgname{University of Virginia}}
\affil[2]{\orgdiv{Dept. of Aerospace Engineering \& Engg. Mechanics}, \orgname{The University of Texas at Austin}}
\affil[3]{\orgdiv{Dept. of Computer Science}, \orgname{The University of Texas at Austin}}
\affil[4]{\orgdiv{Executive Director}, \orgname{Sony AI, America}}

\abstract{
We present an approach to ensure safe and deadlock-free navigation for decentralized multi-robot systems operating in constrained environments, including doorways and intersections. Although many solutions have been proposed that ensure safety and resolve deadlocks, optimally preventing deadlocks in a minimally invasive and decentralized fashion remains an open problem. We first formalize the objective as a non-cooperative, non-communicative, partially observable multi-robot navigation problem in constrained spaces with multiple conflicting agents, which we term as social mini-games. Formally, we solve a discrete-time optimal receding horizon control problem leveraging control barrier functions for safe long-horizon planning. Our approach to ensuring liveness rests on the insight that \textit{there exists barrier certificates that allow each robot to preemptively perturb their state in a minimally-invasive fashion onto liveness sets i.e. states where robots are deadlock-free}. We evaluate our approach in simulation as well on physical robots using F$1/10$ robots, a Clearpath Jackal, as well as a Boston Dynamics Spot in a doorway, hallway, and corridor intersection scenario. Compared to both fully decentralized and centralized approaches with and without deadlock resolution capabilities, we demonstrate that our approach results in safer, more efficient, and smoother navigation, based on a comprehensive set of metrics including success rate, collision rate, stop time, change in velocity, path deviation, time-to-goal, and flow rate. }


\maketitle
% \pagenumbering{arabic}

% \thispagestyle{plain}
% \pagestyle{plain]}
\section{Introduction}
\label{sec: introduction}

We consider the task of multi-robot navigation in constrained environments such as passing through narrow doors and hallways, or negotiating right of way at corridor intersections. We refer to these types of scenarios as \emph{social mini-games}. Unlike humans, robots often collide or end up in a deadlock due to several challenges arising in social mini-games. Two key challenges, in particular, demand attention. First, without some form of cooperation, decentralized systems, even with perfect local sensing, result in deadlocks, collisions, or non-smooth trajectories~\cite{chandra2022gameplan,chandra2022socialmapf}. Second, humans are adept at avoiding collisions and deadlocks without having to deviate too much from their preferred walking speed or trajectory~\cite{francis2023principles, raj2024rethinking,gouru2024livenet, chandra2024towards, wu2023intent}. For instance, when two individuals go through a doorway together, one person modulates their velocity by just enough to enable the other person to pass through first, while still adhering closely to their preferred speed. This type of behavior presents a significant challenge for robots, which struggle to emulate such socially adaptive maneuvers while maintaining a consistent preferred velocity. The objective of this paper is to propose a safe and deadlock-free navigation algorithm for multiple robots in social mini-games. 
% \begin{figure}[t]
%     \centering
%     \includegraphics[width=\columnwidth]{images/base_doorway_profiling.png}
%     \caption{Caption}
%     \label{fig:enter-label}
% \end{figure}
% \input{images/cover/cover.tex}
\begin{table*}[t]
    \centering
    \resizebox{\textwidth}{!}{
    \begin{threeparttable}
    \begin{tabular}{lccc|cccc}
    \toprule
    &\multicolumn{3}{c}{Necessary Algorithmic Conditions} & \multicolumn{3}{c}{Assumptions}\\
    \midrule
       \multirow{2}{*}{Approach}   & \multirow{2}{*}{Collision-free}& \multirow{2}{*}{Liveness$^\dagger$} & \multirow{2}{*}{Dynamics}  & \multirow{2}{*}{Control}  & Non$^\mathsection$ & \multirow{2}{*}{Observation}  \\
        &   &  &   && Cooperative &    \\
     \midrule
DRL~\cite{long2018towards,cadrl,cadrl-lstm,ga3c-cadrl}&\xmark&\xmark &Differential drive&CTDE &\xmark&Partial\\   
ORCA-MAPF~\cite{orcamapf}&\xmark&\cmark &Single-integrator&Centralized &\xmark&Full\\  
Prediction + Planning~\cite{kamenev2022predictionnet} &\xmark&\xmark &$-$&CTDE &\xmark&Full\\
Game-theoretic Dist. Opt.~\cite{le2022algames}&\cmark&\xmark &Unicycle&Distributed &\cmark&Full\\
NH-TTC~\cite{davis2019nh}&\cmark&\xmark&Differential drive&Decentralized&\xmark&Full\\
Auction-based~\cite{snupi}&\xmark&\cmark$^*$ &Ackermann&Distributed &\cmark&Partial\\
Buffered Voronoi Cells~\cite{impc}&\xmark&\cmark$^*$ &Double-integrator&Distributed &\cmark&Full\\
% \midrule
NH-ORCA~\cite{nh-orca}&\cmark&\xmark&Differential drive&Decentralized&\cmark&Partial\\
CBFs~\cite{wang2017safety}&\cmark&\xmark &Double-integrator&Decentralized &\cmark&Partial\\
CBFs+KKT~\cite{grover2016before}&\cmark&\cmark$^*$ &Single-integrator&Decentralized &\cmark&Partial\\
DS-MPEPC~\cite{arul2023ds}&\cmark&\cmark$^*$ &Differential drive&Decentralized &\cmark&Partial\\
RLSS~\cite{csenbacslar2023rlss}&\cmark&\xmark &Differentially flat&Decentralized &\xmark&Partial\\
Humans &\cmark&\cmark&$-$&Decentralized &\cmark&Partial\\   
\rowcolor{gray!25}\textbf{This paper }&\textbf{\cmark}&\textbf{\cmark}&\textbf{Double-integrator}&\textbf{Decentralized} &\textbf{\cmark}&Partial\\    
     \bottomrule 
    \end{tabular}
    
{\footnotesize\begin{tablenotes}
\item[$\star$] Under certain conditions.
\item[$\mathsection$]Non-cooperative robots: robots optimize their individual objective functions~\cite{gt-marl-survey}.
\item[$\dagger$] Deadlock-free.
% \item[$\dagger\dagger$] Each robot performs decision making individually relying on local sensor observations.
\end{tablenotes}}
\end{threeparttable}
    }
    \caption{Comparing various approaches for multi-robot navigation in social mini-games.}
    \label{tab: comparing_approaches}
    % \vspace{-10pt}
\end{table*}

The goal is for robots to navigate in such social mini-games as humans do as much as possible. More formally, we outline a list of \textit{necessary} conditions along with some assumptions of the multi-robot navigation problem that can shift the nature of the problem toward more or less human-like behavior:

\begin{itemize}
\item Necessary conditions:
\begin{enumerate}
\item \textit{Ensure collision-free controls:} Navigation algorithms must produce controls that guarantee robots traverse their environment without collisions. 
% avoid moving in ways that result in collisions, navigation algorithms must also generate controls that ensure robots can navigate without collisions.
\item \textit{Demonstrate liveness:} Navigation algorithms must detect and prevent deadlocks in a minimally invasive fashion.
\item \textit{Obey kinodynamic constraints:} Humans operate under physical and dynamic constraints.
% , such as the limitations of our bodies. 
Navigation algorithms must both operate under complex kinodynamic constraints of the robot such as speed and acceleration limits, as well as be deployable on real robots.
% \item \textit{Real world deployability:} Humans navigate in the real world, where they must adapt to changing environments and make decisions in uncertain situations. For robots to navigate like humans do, they must be able to operate in the real world.
\end{enumerate}
\item Navigation algorithms can make several assumptions about the robot's behavior. We enumerate the possible assumptions as follows:
\begin{enumerate}[noitemsep]
\item \textit{Control Operation:} Robots can operate in a \ul{centralized}, \ul{decentralized}, or \ul{distributed} manner, which can affect the level of coordination required for successful navigation.
\item \textit{Non-cooperative agents:} Robots can be \ul{non-cooperative} where each robot optimizes its own objective function or \ul{cooperative} where each robot optimizes a global system objective.
\item \textit{Observability:} The level of observability can vary for robots, with some agents relying on \ul{partial} or \ul{full} observability of their surroundings, impacting their ability to make decisions and navigate successfully.
\end{enumerate}
\end{itemize}

\noindent We compare different classes of methods according to these desiderata and assumptions in Table~\ref{tab: comparing_approaches}. Although multi-robot navigation encompasses a vast range of algorithms, we narrow our focus to algorithms that have been applied to social mini-games either on real robots or in simulation. These include methods based on deep reinforcement learning~\cite{long2018towards,cadrl,cadrl-lstm,ga3c-cadrl, mavrogiannis2022b, sathyamoorthy2020densecavoid}, multi-agent path finding~\cite{orcamapf}, trajectory prediction~\cite{kamenev2022predictionnet, chandra2019densepeds, chandra2020roadtrack, chandra2019traphic, chandra2020forecasting}, game-theoretic distributed optimization\cite{le2022algames}, auctions~\cite{chandra2022socialmapf}, geometric planning~\cite{nh-orca, orca, orcamapf}, behavior modeling approaches~\cite{chandra2020cmetric, chandra2020graphrqi, chandra2021using, chandra2022towards}, and other optimization-based methods\cite{grover2016before, arul2023ds, wang2017safety, impc, davis2019nh}. From Table~\ref{tab: comparing_approaches}, we note that none of these methods satisfy all of the necessary conditions for optimal multi-robot navigation in social mini-games. 
% In the literature on multi-robot navigation for social mini-games, typically, the problem is that 
This paper addresses the following open research question: 

\begin{quote}
\begin{mdframed}
\textit{\textbf{Research Question:} how can we design an algorithm that can satisfy all of the necessary specifications for optimal multi-robot navigation in social mini-games?}
\end{mdframed}
\end{quote}

\textbf{Main Contributions:} We present the first approach for optimal, safe, and (preventive) deadlock-free receding-horizon navigation for robots with double-integrator dynamics in social mini-games, such as navigating through a narrow door or negotiating right of way at a corridor intersection. Our algorithm is minimally invasive, fully decentralized\footnote{while our approach is decentralized in most practical scenarios, we utilize a centralized auction protocol for tie-breaking in the rare case when robots speeds are exactly identical \textit{up to numeric precision}. Our approach then switches to a centralized mode for \textit{only as long as the auction is active}, and returns to decentralized mode when the auction is switched off. The auction, however, is rarely used in practice.}, and works in realtime both in simulation as well as on physical robots. Our main contributions include:

\begin{enumerate}

    \item We formally define social navigation in geometrically constrained environments through the notion of social mini-games. Prior research~\cite{grover2020does, impc} has only identified social mini-games as \textit{causes} of deadlocks, which they can then only resolve once the deadlock occurs. In this work, we leverage the geometrical properties of social mini-games to detect and \textit{prevent} deadlocks from occurring in an online fashion.
    
    \item We present a new class of multi-agent, realtime, decentralized controllers that perform safe and deadlock-free navigation via a discrete-time, finite-horizon MPC formulation. Safety is modeled via control barrier functions and liveness is ensured by a geometrically motivated detection-and-prevention strategy. Both safety and liveness are guaranteed in realtime by integrating them as constraints to the optimal control problem.

    \item We evaluate our proposed navigation algorithm in simulation as well in the real world using F$1/10$ robots, a Clearpath Jackal, and a Boston Dynamics Spot in a doorway, corridor intersection, and hallway scenario. We show that our approach results in safer and more efficient navigation compared to local planners based on geometrical constraints, optimization, multi-agent reinforcement learning, and auctions. Specifically, we show that our deadlock resolution is the smoothest in terms of smallest average change in velocity as well as path deviation. We demonstrate efficiency by measuring makespan and showing that the throughput generated by robots using our approach to navigate through these constrained spaces is identical to when humans navigate these environments naturally.

    % \item These controllers can generally be tacked on to any constrained optimization-based local trajectory planner such as model predictive control (MPC) or dynamic window approach (DWA), by simply adding our controller as an additional constraint.
    
    % \item Our approach to ensuring liveness rests on the novel insight that there exists barrier certificates that allow each robot to independently perturb their state in a minimally-invasive fashion onto liveness sets i.e. states where robots are deadlock-free

\end{enumerate}

% Additionally, our approach yields a flow rate of $2.8 - 3.3$ (ms)$^{-1}$ which is comparable to flow rate in human navigation at $4$ (ms)$^{-1}$.

In the remainder of this paper, we discuss related work in Section~\ref{ref: related_work} and formulate the problem in Section~\ref{sec: prob_formulation}. We present our controller in Section~\ref{sec: GT-resolution} and evaluate its performance in Section~\ref{sec: experiments}. We conclude the paper in Section~\ref{sec: conc}.

\section{Related Work}
\label{ref: related_work}

In this section, we review the existing approaches for multi-robot navigation in social mini-games. We categorize the approaches based on their mode of operation which could be decentralized, centralized, distributed, or a special class pertaining specifically to learning-based methods that rely on centralized training and decentralized execution (CTDE). Table \ref{tab: comparing_approaches} summarizes the comparison of these approaches based on mode of operation along with safety, liveness, real-world deployment, decentralized decision-making, self-interested agents, and observation.

\subsection{Collision Avoidance}
\label{subsec: collision_avoidance} 

Provable safety can be achieved by single-integrator systems \textit{e.g.} ORCA framework from Van Den Berg et al.~\cite{orca} and its non-holonomic variant~\cite{nh-orca}, which are effective for fast and exact multi-agent navigation. ORCA conservatively imposes collision avoidance constraints on the motion of a robot in terms of half-planes in the space of velocities. The optimal collision-free velocity can then be quickly found by solving a linear program. The original framework limits itself to holonomic systems but has been extended in~\cite{nh-orca} to model non-holonomic constraints with differential drive dynamics. ORCA also generates collision-free velocities that deviate minimally from the robots' preferred velocities. The major limitation of the ORCA framework is that the structure of the half-planes so constructed often results in deadlocks~\cite{orcamapf}.

Proving safety is harder for systems with double-integrator dynamics, therefore safety in these systems depends on the planning frequency of the system. For example, the NH-TTC algorithm~\cite{davis2019nh} uses gradient descent to minimize a cost function comprising a goal reaching term and a time-to-collision term, which rises to infinity as the agent approaches immediate collision. NH-TTC guarantees safety in the limit as the planning frequency approaches infinity. Other optimization-based approaches use model predictive control (MPC)~\cite{impc}; in such approaches, safety depends not only on the planning frequency but also on the length of the planning horizon.

Finally, Control Barrier Functions (CBFs)~\cite{grover2016before, wang2017safety} can be used to design controllers that can guarantee safety via the notion of forward invariance of a set \textit{i.e.} if an agent starts out in a safe set at the initial time step, then it remains safe for all future time steps, that is, it will never leave the safe set.
% and uses gradient-based optimization to compute the minima of the cost function. The advantages of NH-TTC over ORCA include application to complex dynamics and deployment in real robots. The disadvantages are that it requires full observability of the state space and does not model self-interested agents. Similar to the ORCA framework, NH-TTC also results in deadlocks in constrained environments.

\subsection{Deadlock Resolution Methods}

Deadlocks among agents arise due to symmetry in the environment that may cause conflicts between the agents~\cite{grover2020does, impc, grover2016before}. To break the symmetry, and escape the deadlock, agents must be perturbed, which can be done in several ways. The most naive, and easiest, way is to randomly perturb each agent~\cite{wang2017safety}. Random perturbations can be implemented in decentralized controllers and can generalize to many agents, but are sub-optimal in terms of path deviation and overall cost. Next, there are several recent efforts to choreograph the perturbation according to some set rules such as the right-hand rule~\cite{impc, zhou2017fast} or clockwise rotation~\cite{grover2016before}. These strategies improve performance over random perturbation and even give formal guarantees of optimality, but the imposed pre-determined ordering limits their generalizability; many cannot generalize to more than $3$ agents. Another line of research aims towards deadlock \textit{prevention} rather than resolution where an additional objective is to identify and mitigate potential deadlocks, even before they happen, such as in~\cite{impc}.

Another class of deadlock resolution methods rely on priority protocols and scheduling algorithms similar to those used in the autonomous intersection management literature~\cite{aim_survey_1}. Some prominent protocols include first come first served (FCFS), auctions, and reservations. FCFS~\cite{fcfs} assigns priorities to agents based on their arrival order at the intersection. 
% The first vehicle to arrive at the intersection is given the highest priority and is allowed to cross the intersection first, followed by the next vehicle, and so on. 
It is easy to implement but can lead to long wait times and high congestion if multiple vehicles arrive at the intersection simultaneously. In auctions~\cite{carlino2013auction, suriyarachchi2022gameopt}, agents bid to cross the intersection based on a specific bidding strategy. 
% Auctions are more complex then FCFS, but allow agents to express their private priorities via decentralized bidding unlike FCFS, which is centralized. 
Reservation-based systems~\cite{reservation} are similar to the auction-based system in which agents reserve slots to cross the intersection based on their estimated arrival and clearance times.

As noted by recent researchers~\cite{impc,zhou2017fast}, developing a provably optimal, decentralized, and general deadlock resolution technique is currently an open problem. In this work, we take a large step forward towards a solution.

\subsection{Learning-based Approaches}
\label{subsec: learning_based}

Coupling classical navigation with machine learning is rapidly growing field and we refer the reader to Xiao et al.~\cite{motion_survey} for a recent survey on the state-of-the-art of learning-based motion planning. Here, we review two categories of approaches that have been reasonably succesful in multi-agent planning and navigation. These are methods based on deep reinforcement learning (DRL) and trajectory prediction. DRL has been used to train navigation policies in simulation for multiple robots in social mini-games. Long et al.~\cite{long2018towards} presents a DRL approach for multi-robot decentralized collision avoidance, using local sensory information. They present simulations in various group scenarios including social mini-games. CADRL~\cite{cadrl}, or Collision Avoidance with Deep Reinforcement Learning, is a state-of-the-art motion planning algorithm for social robot navigation using a sparse reward signal to reach the goal and penalizes robots for venturing close to other robots. A variant of CADRL uses LSTMs to select actions based on observations of a varying number of nearby robots~\cite{cadrl-lstm}. Planning algorithms that use trajectory prediction models~\cite{kamenev2022predictionnet} estimate the future states of the robot in the presence of dynamic obstacles, and plan their actions accordingly. However, DRL-based methods are generally hard to train requiring tens of thousands of data samples, hours of training time, do not generalize to out of distribution environments, and lastly, they impose only soft constraints (collision checking) on safety, and do not provide hard guarantees.
\begin{table}[t]
    \centering
    \resizebox{\columnwidth}{!}{
    \begin{tabular}{cc}
    \toprule
       Symbol  & Description  \\
       \midrule
       \multicolumn{2}{c}{\textit{Problem formulation (Section~\ref{subsec: problem_formulation}})}\\
       \midrule
       $k$ & Number of agents\\
       $T$ & planning horizon\\
       $\mathcal{X}$ & general continuous state space\\
       $\mathcal{X}_I$ & set of initial states\\
       $\mathcal{X}_g$ &set of final states\\
       $x^i_t$ & state of agent $i$ at time $t$\\
       $\bar x^i_t$ & observable state of agent $i$ to other agents\\
       $\Omega^i$ & observation set of agent $i$\\
       $o^i_t$ & observation of agent $i$ at time $t$\\
       $\mathcal{O}^i$ & observation function ($\mathcal{O}^i: \mathcal{X}\rightarrow \Omega^i$)\\
       $\mathcal{N}^i\left( x^i_t\right)$ &  set of robots detected by $i$\\
       $\Gamma^i$ & agent $i$'s trajectory\\
       $\widetilde\Gamma^i$ & set of preferred trajectories\\
       $\mathcal{T}$ & transition dynamics (Equation~\ref{eq: control_affine_dynamics})\\
       $\mathcal{U}^i$ & action space for agent $i$\\
       $\mathcal{J}^i$ & running cost for agent $i$ ($\mathcal{J}^i_t:\mathcal{X} \times  \mathcal{U}^i \rightarrow \mathbb{R}$)\\
       $\mathcal{J}^i_f$ & running cost at time $T$\\
       $\mathcal{C}^i\left( x^i_t \right) \in \mathcal{X}$ & convex hull of agent $i$\\
       $\pi^i \in \mathcal{K}$ &controller belonging to set of controllers\\
       \midrule
       \multicolumn{2}{c}{\textit{Control Barrier Function (Section~\ref{subsec: cbf}})}\\
       \midrule
       $h^i:\mathcal{X} \longrightarrow \mathbb{R}$ & control barrier function\\
       $\mathscr{C}^i$ & safe set\\
       $\partial\mathscr{C}^i$ &  boundary of $\mathscr{C}^i$\\
       $ L_f h^i\left(x^i_t\right),  L_g h^i\left(x^i_t\right)$ & lie derivatives of $h^i\left(x^i_t\right)$ w.r.t $f$ and $g$.\\
       \midrule
       \multicolumn{2}{c}{\textit{Deadlock Prevention (Section~\ref{sec: GT-resolution}})}\\
       \midrule
       $[1;N]$ & set of integers $\{1,\;2,\dots,\;N\}$\\
       $\mathscr{C}_\ell(t)$ &liveness set\\
       $v_t^i$ & linear velocity of agent $i$\\
       $v_t$ & joint velocity of all agents\\
       $\widetilde v_t$ & perturbed joint velocity\\ 
       $h_v\left( x_t\right)$ & CBF of $\mathscr{C}_\ell(t)$\\
       $p_t^i$ & position of agent $i$\\
       $\theta_t^i$ & angle of agent $i$\\
       $\ell_j\left (p_t^i,v_t^i\right )$ & liveness function for agent $i$\\
       $\sigma$ & priority ordering \\
       $\sigma_{\textsc{opt}}$ & optimal ordering \\
       $\alpha_q$ & time-based reward for receiving an order position $q$\\
       $b^i$ & bid made by agent $i$\\
       $\left( r^i, p^i\right)$ & auction specified by allocation and payment rule\\
       $\zeta^i$ & private priority incentive parameter\\
         \bottomrule
    \end{tabular}
    }
    \caption{Summary of notation used in this paper.}
    \label{tab: notation}
\end{table}

On the other hand, Imitation Learning (IL) corresponds to a machine learning paradigm where an autonomous agent strives to learn a behavior by emulating an expert's demonstrations. IL-based approaches are much faster than traditional optimization-based approaches. Typically, the demonstrations comprise state-input pairs, generated by an expert policy during real-world execution. 
Addressing the imitation learning problem has led to three major approaches, namely, Behavior Cloning (BC)~\cite{dagger, Ross2011ReductionOfImitationLearning, Daftry2016LearningTransferablePolicies}, which involves direct learning of an imitative policy through supervised learning, Inverse Reinforcement Learning (IRL)~\cite{maxentirl, mairl, singleirl}, wherein a reward function is first inferred from the demonstrations, subsequently used to guide policy learning via Reinforcement Learning (RL)~\cite{Sutton2018ReinforcementLearning}, and generative models~\cite{Ho2016GenerativeAdversarialIL, Kostrikov2019ImitationLearningViaOffPolicyDistributionMatching, Dadashi2020PrimalWassersteinImitationLearning}.  We refer the reader to Zhang et al.~\cite{zheng2022imitation} for a more comprehensive background on IL. 
However, a shared drawback among the IL algorithms \cite{Kostrikov2019ImitationLearningViaOffPolicyDistributionMatching, Dadashi2020PrimalWassersteinImitationLearning,zheng2022imitation,bco,ifovideo} is their inability to encode state/safety and input constraints
underlying assumption of having access to the expert's action data
during the demonstration phase. 
Additionally, these methods do not transfer to domains that are outside the distribution of the expert demonstrations.

\subsection{Game-theoretic Distributed Optimization}
\label{subsec: related_trajectory_planning}

Another class of methods for multi-agent planning for self-interested agents includes distributed optimization in general-sum differential games. The literature on general-sum differential games classify existing algorithms for solving Nash equilibria in robot navigation into four categories. First, there are algorithms based on decomposition~\cite{wang2021game, britzelmeier2019numerical}, such as Jacobi or Gauss-Siedel methods, that are easy to interpret and scale well with the number of players, but have slow convergence and may require many iterations to find a Nash equilibrium. The second category consists of games~\cite{fisac2019hierarchical}, such as Markovian Stackelberg strategy, that capture the game-theoretic nature of problems but suffer from the curse of dimensionality and are limited to two players. The third category consists of algorithms based on differential dynamic programming~\cite{schwarting2021stochastic,sun2015game,sun2016stochastic,morimoto2003minimax,fridovich2020efficient,di2018differential} that scale polynomially with the number of players and run in real-time, but do not handle constraints well. Lastly, the fourth category contains algorithms based on direct methods in trajectory optimization~\cite{le2022algames, di2019newton, di2020first}, such as Newton's method, that are capable of handling general state and control input constraints, and demonstrate fast convergence. The algorithms described above give an analytical, closed-form solution that guarantees safety but not liveness. Additional limitations include the lack of deployability in the real world and the requirement of full observation.

The key distinction between our approach and traditional game-theoretic approaches is that unlike in the latter case, our approach does not explicitly enforce a Nash equilibrium which requires direct knowledge of the cost functions of other agents. Instead, our approach implicitly and indirectly achieves a solution that resembles a Nash equilibrium. So we essentially perform a tradeoff - we deviate slightly from a precise Nash solution to something that looks like a Nash solution, but our approach works in realtime on real robots in a fully decentralized manner.

% The algorithms described above give an analytical, closed-form solution that guarantees safety but not liveness. Additional limitations include the lack of deployability in the real world and the requirement of full observation.

\section{Problem Formulation and Background}
\label{sec: prob_formulation}
In this section, we begin by formulating social mini-games followed by stating the problem objective. Notations used in this paper are summarized in Table~\ref{tab: notation}.

\subsection{Problem Formulation}
\label{subsec: problem_formulation}
We formulate a social mini-game by augmenting a partially observable stochastic game (POSG)~\cite{posg}: $\left \langle k, \mathcal{X}, \{\Omega^i\}, \{\mathcal{O}^i\},\{\mathcal{U}^i\}, \mathcal{T},  \{\widetilde\Gamma^i\}, \{\mathcal{J}^i\}\right\rangle$ where $k$ denotes the number of robots. Hereafter, $i$ will refer to the index of a robot and appear as a superscript whereas $t$ will refer to the current time-step and appear as a subscript. The general state space $\mathcal{X}$ (\textit{e.g.} SE(2), SE(3), etc.) is continuous; the $i^\textrm{th}$ robot at time $t$ has a state $x^i_t\in \mathcal{X}$. A state $x^i_t$ consists of both visible parameters (\textit{e.g.} current position, linear and angular velocity and hidden (to other agents) parameters which could refer to the internal state of the robot such as preferred speed, preferred heading, etc. We denote the set of observable parameters as $\overline{x}^i_t$. On arriving at a current state $x^i_t$, each robot generates a local observation,  $o^i_t \in \Omega^i$, via $\mathcal{O}^i: \mathcal{X}\longrightarrow \Omega^i$, where $ \mathcal{O}^i\left(x^i_t\right) = \left\{x^i_t\right\} \cup \left\{\overline{x}^j_t: j\in \mathcal{N}^i\left( x^i_t\right)\right\}$, the set of robots detected by $i$'s sensors. Over a finite horizon $T$, each robot is initialized with a start state $x^i_{0} \in \mathcal{X}_I$, a goal state $x^i_N\in \mathcal{X}_g$ where $\mathcal{X}_I$ and $\mathcal{X}_g$ denote subsets of $\mathcal{X}$ containing the initial and final states. The transition function is given by $\mathcal{T}:\mathcal{X}\times \mathcal{U}^i \longrightarrow \mathcal{X}$, where $\mathcal{U}^i$ is the continuous control space for robot $i$ representing the set of admissible inputs for $i$. A discrete trajectory is specified as $\Gamma^i = \left( x^i_{0}, x^i_1, \ldots, x^i_T  \right)$ and its corresponding input sequence is denoted by $\Psi^i = \left( u^i_{0}, u^i_1, \ldots, u^i_{T-1}  \right)$. Robots follow the discrete-time control-affine system,
% \textcolor{blue}{(1) is continous time and (3b) is discrete time. However both have same $f$ and $g$}

\begin{equation}
x^i_{t+1} = f\left(x^i_t\right) + g\left(x^i_t\right)u^i_t
    \label{eq: control_affine_dynamics}
\end{equation}
which is obtained after applying a Runge-Kutta discretization scheme to a continuous-time system describing their motion (derived from first principles) which takes the following form:
\begin{equation}
% x^i(t) = f_c\left(x^i(t)\right) + g_c\left(x^i(t)\right)u^i(t),
x^i(t) = f_c\left(x^i(t)\right) + g_c\left(x^i(t)\right)u^i(t),
    \label{eq:CTcontrol_affine_dynamics}
\end{equation}
where $f_c$ and $g_c$ are locally Lipschitz continuous functions (note that in general, the state and input vectors of the continuous-time and the discrete-time systems that correspond to the same time instant are not  the same due to discretization induced errors). In this paper, we will mainly utilize the discrete-time model~\eqref{eq: control_affine_dynamics}, especially for control design purposes (these methods will rely on optimization techniques), but we will also refer to the continuous-time model \eqref{eq:CTcontrol_affine_dynamics} for analysis purposes. We will also assume that the continuous-time system~\eqref{eq:CTcontrol_affine_dynamics} is small-time controllable, which means that set of points reachable from $x^i(t)$ within the time interval $[t,t^\prime]$ will contain a neighborhood of $x^i(t)$ for any $t^\prime>t$~\cite{laumond2005guidelines}. Small-time controllability allows the following result to hold true:

\begin{theorem}[Laumond et al.\cite{laumond2005guidelines}]
    For symmetric small-time controllable systems the existence of
an admissible collision-free path between two given configurations is equivalent
to the existence of any collision-free path between these configurations.
\label{thm: laumond}
\end{theorem}

We denote by $\widetilde \Gamma^i$ as the set of \textit{preferred} trajectories for robot $i$ that solve the two-point boundary value problem. A preferred trajectory, as defined by existing methods~\cite{impc, orca, nh-orca}, refers to a collision-free path a robot would follow in the absence of dynamic or static obstacles and is generated via a default planner according to some predefined criteria such as shortest path, minimum time, etc. A collision is defined as follows. Let $\mathcal{C}^i\left( x^i_t \right) \subseteq \mathcal{X}$ represent the space occupied by robot $i$ (as a subset of the state space $\mathcal{X}$) at any time $t$ which can be approximated by the convex hull of a finite number of points that determine the boundary of the robot (e.g., vertices of a polytopic set). Then, robots $i$ and $j$ are said to collide at time $t$ if $\mathcal{C}^i\left(x^i_t \right) \cap \mathcal{C}^j\left(x^j_t \right) \neq \emptyset$.

% for every state $x^i_t\in\widetilde \Gamma^i$, there exists $t^\prime \leq T$ such that $x^i_{t+t^\prime} \in \mathcal{R}\left( x^i_t, \mathcal{U}^i, t^\prime\right) \cap \widetilde \Gamma^i$, where $\mathcal{R}\left( x^i_t, \mathcal{U}^i, t^\prime\right)$ is the set of reachable states from $x^i_t$ traveling for time $t^\prime$.   %% smooth vector fields.
% We define a controller or policy as $\Gamma^i: \Omega^i \longrightarrow \mathcal{U}^i$ that takes in an observation $o^i_t$ and outputs a control $u^i_t$ such as velocity or acceleration. 
% \input{algorithms/flowchart}
Each robot has a running cost $\mathcal{J}^i:\mathcal{X} \times  \mathcal{U}^i \longrightarrow \mathbb{R}$ that assigns a cost to a state-input pair $\left(x^i_t, u^i_t\right)$ at each time step based on user-defined variables \textit{e.g.} distance of the robots current position from the goal, change in the control across subsequent time steps, and distance between the robots preferred and actual paths. Finally, we define a social mini-game as follows,

\begin{definition}
A \textbf{social mini-game} occurs if for some $\delta > 0$ and integers $a,b \in (0,T)$ with $ b-a > \delta$, there exists at least one pair $i,j, i\neq j$ such that for all $\Gamma^i \in \widetilde \Gamma^i, \Gamma^j \in \widetilde \Gamma^j$, we have $\mathcal{C}^i\left( x^i_t \right) \cap \mathcal{C}^j\left( x^j_t \right)\neq \emptyset \ \forall \ t \in [a,b]$, where $x^i_t, x^j_t$ are elements of $\Gamma^i$ and $\Gamma^j$.
\label{def: social_minigame}
\end{definition}

We depict several examples and non-examples of social mini-games in Figure~\ref{fig: social_minigame}. The first scenario can be characterized as a social mini-game due to the conflicting preferred trajectories of agents $1$ and $2$ within a specific time interval $[a, b]$, where the duration $b-a \geq \delta$. In this case, the agents' trajectories intersect, generating a conflict. However, the second and third scenarios, in contrast, do not qualify as social mini-games since no conflicts arise between the agents. In the second scenario, there is no common time duration where agents intersect each other, and their trajectories remain independent in time. In the third scenario, agent $2$ possesses an alternative preferred trajectory that avoids conflicts during any time duration, allowing for a seamless transition to a conflict-free path. A robot has the following best response in a social mini-game:

\begin{figure}[t]
\centering
\includegraphics[width=\columnwidth]{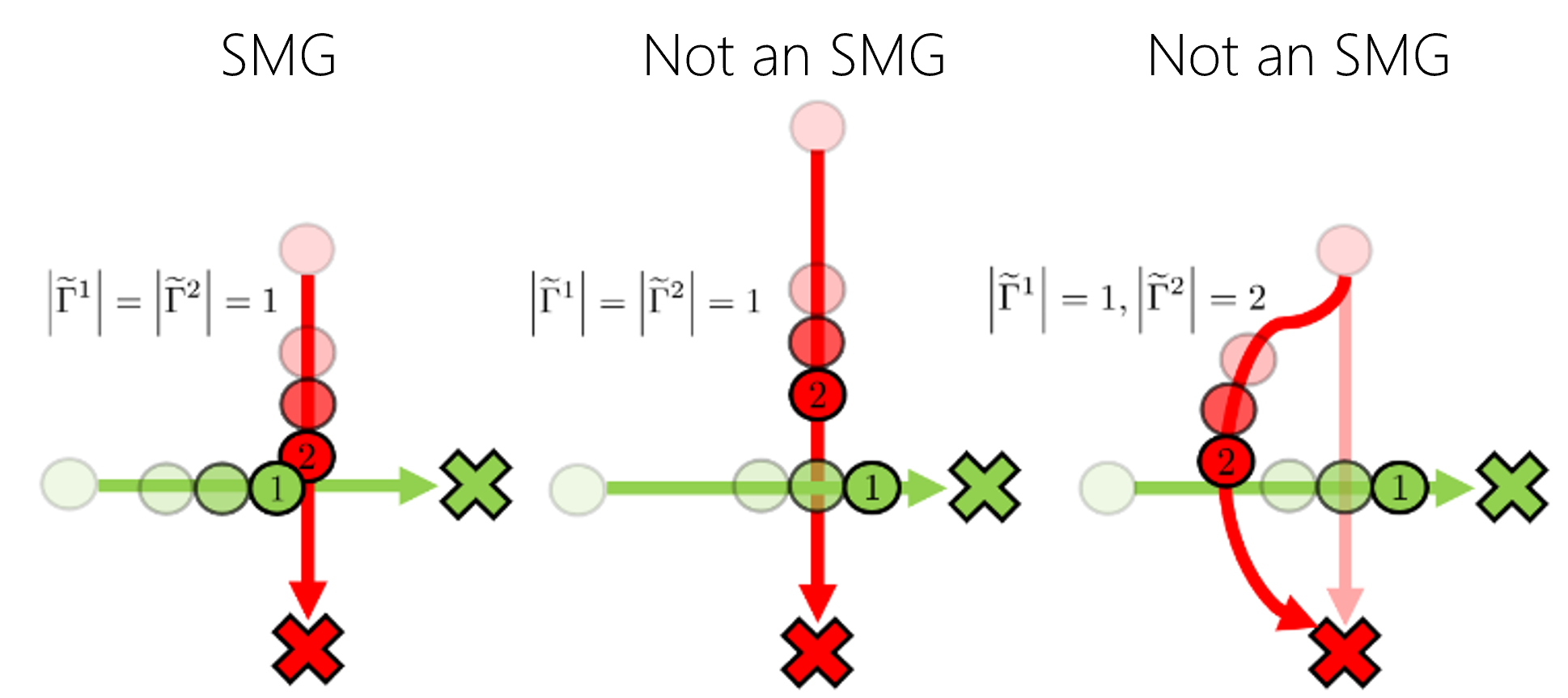}
\caption{\textbf{Examples/Counter-examples of social mini-games:} Arrows indicate the direction of motion for two agents $1$ and $2$ toward their goals marked by the corresponding cross. The first scenario is a social mini-game since both the preferred trajectories of agents $1$ and $2$ are in conflict from some $t=a$ to $t=b$ where $b-a \geq \delta$. The second and third scenarios are \textit{not} social mini-games as there are no conflicts. In the second scenario, there is no duration where agents intersect one another. In the third scenario, agent $2$ has an alternate conflict-free preferred trajectory to fall back on.}
  \label{fig: social_minigame}
  % \vspace{-10pt}
\end{figure} 

\begin{definition}
\textbf{Optimal Solution for a Robot in Social Mini-Games:} For the $i^\textrm{th}$ robot, given its initial state $x^i_0$, an optimal trajectory $\Gamma^{i,*}$ and corresponding optimal input sequence $\Psi^{i,*}$ are given by,
\begin{subequations}
\begin{align}
\left( \Gamma^{i,*}, \Psi^{i,*}\right) =& \arg\min_{(\Gamma^i,  \Psi^{i})} \sum_{t={0}}^{T-1} \mathcal{J}^i\left(x^i_t, u^i_t\right) + \mathcal{J}^i_T\left(x^i_N\right) \label{eq: solving_3}\\
\text{s.t}\;\;  x^i_{t+1}=& f\left(x^i_t\right) + g\left(x^i_t\right)u^i_t,\quad\forall t\in[1;T-1]\label{eq: discrete_dynamics}\\
\mathcal{C}^i\left( x^i_t \right) &\cap \mathcal{C}^j\left( x^j_t \right)= \emptyset \ \forall j \in \mathcal{N}^i\left(x^i_t \right) \ \forall t \label{eq: sub_coll}\\
% &x^i_0\in \mathcal{X}_i\\ 
&x^i_T\in \mathcal{X}_g\label{eq: libe_constraint}
\end{align}
\label{eq: best_response_example}
\end{subequations}
\label{def: best_response}
\end{definition}
\vspace{-0.5cm}
where $\mathcal{J}^i_T$ is the terminal cost.  A solution to optimal navigation in social mini-games is a (finite) sequence of state-input pairs $\left(\left(\Gamma^{1,*}, \Psi^{1,*}\right),\left(\Gamma^{2,*}, \Psi^{2,*}\right),\ldots, \left(\Gamma^{k,*}, \Psi^{k,*}\right)\right )$. 

% The objective function captures the cost of a robot deviating from its preferred trajectory between the robot $i$'s current position and the position in the preferred trajectory at time $t \in [0,T]$. An example of such a cost function is $\mathcal{J}^i\left(x^i_t\right) = \left\lVert x^i_t - \widetilde x^i_t  \right\rVert^2$.

\subsection{Algorithm Overview for Solving Social Mini-games}
As solving~\eqref{eq: best_response_example} jointly is computationally intensive~\cite{chen2010sequential}, we solve the multi-robot navigation problem~\eqref{eq: best_response_example} for the discrete-time system~\eqref{eq: control_affine_dynamics} in a decentralized fashion. Each robot solves~\eqref{eq: best_response_example} in a receding horizon fashion using an MPC asynchronously; on the $i^\textrm{th}$ robot's turn, it treats all other robots as static obstacles that are incorporated into the safety constraint~\eqref{eq: sub_coll}. At each simulation iteration for the $i^\textrm{th}$ robot, the MPC controller solves the constrained finite-time optimal control problem~\eqref{eq: best_response_example} with horizon $T$. The optimal control sequence to this optimization problem for the $i^\textrm{th}$ robot is a sequence of inputs $\Psi^{i,*} = \left(u^{i,*}_{t},u^{i,*}_{t+1} \ldots, u^{i,*}_{t+T-1}\right )$, $u^{i,*}_{t} \in \mathscr{U}^i_t$. Then, the first element of the solution, $u^i_t \gets u^{i,*}_t$, is executed by robot $i$. The optimization is repeated at the next iteration based on the updated state for all robots. 

Solving~\mbox{\eqref{eq: solving_3}} asynchronously, however, may lead to different agents making decisions based on different sets of information at different times. This inconsistency can lead to a lack of coordination among agents, resulting in safety and deadlock issues. We address both these issues with a combination of using a sufficiently small sampling time as well as using control barrier functions in the following manner:

\begin{itemize}
    \item \textit{Safety (Constraint~\mbox{\ref{eq: sub_coll}}):} We model static and dynamic safety in Equation~\mbox{\ref{eq: sub_coll}} in the context of set invariance using Control Barrier Functions~\mbox{\cite{ames2019control}} (CBFs).
    \item \textit{Deadlock Prevention (Constraint~\mbox{\ref{eq: libe_constraint}}):} We detect deadlocks by exploiting the geometrical symmetry of social mini-games~\mbox{\cite{grover2020does}} in an online fashion, and design a minimally invasive perturbation strategy to prevent the deadlock from occurring. The deadlock prevention algorithm can be integrated with CBFs for double-integrator (or higher order) dynamical systems.

\end{itemize}

In the remainder of this article, we describe our approaches for addressing safety via CBFs in Section~\mbox{\ref{subsec: cbf}} followed by our deadlock prevention algorithm in Section~\mbox{\ref{sec: GT-resolution}}. We present results in Section~\mbox{\ref{sec: experiments}} and conclude in Section~\mbox{\ref{sec: conc}}.

\section{Safety via Control Barrier Functions (Constraint~\ref{eq: sub_coll})}
\label{subsec: cbf}

Control Barrier Functions (CBFs) are a powerful tool used in control theory for designing controllers that guarantee safety of controlled nonlinear systems~\cite{ames2019control}. CBF-based controllers constrain the behavior of a robot system to enable collision-free trajectories. CBF-based controllers also guarantee safety and robustness in multi-agent systems and scale to a large number of robots, while easily adapting to changes in the environment or robot dynamics. Additionally, they have been combined with traditional control techniques, such as Model Predictive Control (MPC) \cite{zeng2021safety_cbf_mpc} and PID, for improved safety and performance. In this background, we will discuss the basic theory of CBFs and their mathematical formulation.

The safety of a set $\mathscr{C}^i$ for a given system is closely related to its forward invariance which is a property that requires the system starting from a given set of initial conditions inside $\mathscr{C}^i$ to remain in $\mathscr{C}^i$ for all times. We first describe CBFs for the continuous-time case and then specify the modified variant for the discrete-time case. Consider a scalar valued function, $h^i:\mathcal{X} \longrightarrow \mathbb{R}$, where $\mathcal{X}$ denotes the set of admissible states $x^i_t\in \mathcal{X}$ such that the following conditions hold:
% for a dynamical system that ensures \textit{forward invariance} of that set. This means that if $x^i_{0} \in \mathscr{C}^i$, then $x^i_t \in \mathscr{C}^i$ for all $ t \geq 0$. In other words, if $\mathscr{C}^i$ is guaranteed to be collision free, then as long as the robot exists in $\mathscr{C}^i$, it will remain collision-free. $\mathscr{C}^i$ is defined as the superlevel\footnote{A superlevel set of a function $q(x), x \in \mathcal{D}$ is the set of points $x \in \mathcal{D}$ such that $q(x) \geq 0$} set of $h^i$:
\begin{subequations}
    \begin{align}
          &\mathscr{C}^i = \left\{x^i_t\in \mathbb{R}^n | h^i\left(x^i_t\right) \geq 0\right\}\\
    &h^i\left(x^i_t\right) = 0 \ \forall\; x^i_t \in \partial\mathscr{C}^i\\
       &h^i\left(x^i_t\right) < 0 \ \forall\; x^i_t \in \mathbb{R}^n\setminus\mathscr{C}^i  
    \end{align}
    \label{eq: C_set}
\end{subequations}
% \begin{equation}
% \begin{split}
%     \mathscr{C}^i &= \left\{x^i_t\in \mathbb{R}^n | h^i\left(x^i_t\right) \geq 0\right\}\\
%     &h^i\left(x^i_t\right) = 0 \ \forall\; x^i_t \in \partial\mathscr{C}^i\\
%        &h^i\left(x^i_t\right) < 0 \ \forall\; x^i_t \in \mathbb{R}^n\setminus\mathscr{C}^i\\
% \end{split}
%     \label{eq: C_set}
% \end{equation}
where $\mathscr{C}^i$ is the safe set and $\partial\mathscr{C}^i$ denotes its boundary. The time derivative of $h^i\left(x^i_t\right)$ along the state trajectory of agent $i$ is given as

% \begin{equation}
%     \frac{d\left(h^i\left(x^i_t\right)\right)}{dt} = \frac{\partial h^i\left(x^i_t\right)}{\partial x} \dot x^i_t =  \frac{\partial h^i\left(x^i_t\right)}{\partial x} \left(f\left(x^i_t\right) + g\left(x^i_t\right)u^i_t\right)
%     \label{eq: h_time_derivative}
% \end{equation}

\begin{equation}
 \frac{d\left(h^i\left(x^i_t\right)\right)}{dt} = L_f h^i\left(x^i_t\right) + L_g h^i\left(x^i_t\right)u^i_t
          \label{eq: h_time_derivative}
\end{equation}
where $L_f h^i\left(x^i_t\right)$ and $L_g h^i\left(x^i_t\right)$ denote the Lie derivatives of $ h^i\left(x^i_t\right)$ along $f$ and $g$, respectively.
 % Since $h^i\left(x^i_t\right)$ is a scalar-valued function, the right hand side in Equation~\eqref{eq: h_time_derivative} can be written equivalently as,
% \begin{equation}
%        \frac{\partial h^i\left(x^i_t\right)}{\partial x} \left(f\left(x^i_t\right) + g\left(x^i_t\right)u^i_t\right)
%     \label{eq: safety_barrier_certificate_scalar}
% \end{equation}
Then, $h^i\left(x^i_t\right)$ is a CBF if there exists a class $\mathcal{K}_{\infty}$\footnote{A function $\alpha(\cdot):\mathbb{R}\rightarrow\mathbb{R}$ belongs to the class of $\mathcal{K}_\infty$ functions if it is strictly increasing and in addition, $\alpha(0)=0$ and $\underset{r\rightarrow \infty}{\lim} \alpha(r) = \infty$} function $\kappa$ such that the following holds true
\begin{align}
    \underset{u^i_t\in U}{\sup}\;       L_f h^i\left(x^i_t\right) + L_g h^i\left(x^i_t\right) u^i_t + \kappa\left(h^i\left(x^i_t\right)\right)\geq 0
    \label{eqn:sup_cbf_condition}
\end{align}

We define the safe or collision-free control space $\mathscr{U}^i$ over $x^i_t \in \mathcal{X}$ to be the set of control inputs $u^i_t \in  \mathcal{U}^i$ such that the following inequality holds:
\begin{equation}
       L_f h^i\left(x^i_t\right) + L_g h^i\left(x^i_t\right) u^i_t + \kappa\left(h^i\left(x^i_t\right)\right)\geq 0
    \label{eq: safety_barrier_certificate}
\end{equation}
% The set $\mathscr{U}^i$ that guarantees safety is given by $u^i_t\in \mathcal{U}^i$ such that Equation~\eqref{eq: safety_barrier_certificate} holkds
% \begin{align}
%          \{u^i_t\in \mathcal{U}^i\;| Equation~\label{eq: safety_barrier_certificate}\geq 0\}
%           % \mathscr{U}^i=\{u^i_t\in \mathcal{U}^i\;| L_f h^i\left(x^i_t\right) + L_g h^i\left(x^i_t\right) u^i_t + \kappa\left(h^i\left(x^i_t\right)\right)\geq 0\}
%           \label{eqn:control_input_set_safe}
% \end{align}\
% \input{images/geom/geom}

% where $\kappa$ is a positive-valued monotonic function.
% and $ L_f h^i\left(x^i_t\right),  L_g h^i\left(x^i_t\right)$ are the Lie derivatives of $h^i\left(x^i_t\right)$ with respect to $f$ and $g$.

Equation~\eqref{eq: safety_barrier_certificate} is known as the safety barrier constraint or safety barrier certificate.
% \begin{theorem}
In summary, the set $\mathscr{C}^i \subseteq \mathcal{X}$ is guaranteed to be safe if $\mathscr{U}^i$ is non-empty and $u^i_t \in \mathscr{U}^i$.
% \mathscr{C}^i \subseteq \mathcal{X}$ is the superlevel set of a CBF $h^i\left(x^i_t\right) ,x^i_t \in \mathcal{X}$ as defined in Equation~\eqref{eq: C_set} and Equation~\eqref{eq: safety_barrier_certificate} holds, then $\mathscr{C}^i$ is forward invariant if $u^i_t \in \mathscr{U}^i$.
The formulation of safety barrier certificates presented above is for a single robot, but it can be extended trivially to multi-robot scenarios. The only difference is that the barrier function $h^i\left(x^i_t\right)$ will change to $h(x_t)$; that is, it will lose the indexing on $i$ and will become a function of $\mathcal{X}^k \longrightarrow \mathbb{R}$, which denotes the aggregate states of $k$ robots. Similarly, $\mathscr{C}^i$ will become simply $\mathscr{C} \subset \mathcal{X}^k$. As the system we consider in this paper is discrete-time, due to our MPC-based formulation, instead of~\eqref{eq: safety_barrier_certificate}, we will be using the discrete-time version of the CBF which is given by,

\begin{equation}
\Delta h^i_s\left(x^i_t,  u^i_t\right) \geq -\gamma h_s^i\left(x^i_t\right)
    \label{eq: discrete_barrier_func}
\end{equation}

\noindent where $h_s^i\left(x^i_{t}\right)$ is chosen as an obstacle avoidance barrier function (more details in Section~\ref{subsec: exp_setup}), $\gamma > 0$, and $\Delta h_s^i\left(x^i_{t},  u^i_t\right)$ is given by,
\begin{equation}
    \Delta h_s^i\left(x^i_{t},  u^i_t\right) = h_s^i\left(x^i_{t+1}\right) - h_s^i\left(x^i_{t}\right)
    \label{eq: discrete_del_h}
\end{equation}

The set $\mathscr{U}^i$ that guarantees safety is thus given by,
\begin{align}
          \mathscr{U}^i = \{u^i_t\in \mathcal{U}^i\;|   \Delta h^i_s\left(x^i_t,  u^i_t\right) + \gamma h_s^i\left(x^i_t\right)\geq 0\}
          \label{eqn:control_input_set_safe_discrete}
\end{align}

CBFs offer a stronger guarantee of safety than directly imposing collision avoidance constraints in an MPC formulation. In~\mbox{\cite{zeng2021safety_cbf_mpc}}, it was conclusively found that the problem with directly imposing collision avoidance constraints into an MPC, as opposed to using CBFs, is that the constraint will take effect \textit{only} when the planning horizon intersects with an obstacle. In other words, the robot will not take any action to avoid obstacles until it is close to them (refer to the Figure 2 in Zeng et al.~\mbox{\cite{zeng2021safety_cbf_mpc}}). One way to solve this problem is to use a larger horizon, but that will increase the computational complexity in the optimization.
We refer the reader to Zeng et al.~\cite{zeng2021safety_cbf_mpc} for more details.

\section{Deadlock Prevention (Constraint~\ref{eq: libe_constraint})}
\label{sec: GT-resolution}

In social mini-games, however, $\mathscr{U}^i_t$ often ends up being an empty set, \textit{i.e.} there is no input for which~\eqref{eq: discrete_barrier_func} is satisfied, resulting in deadlocks~\cite{wang2017safety, zhou2017fast, impc} due to the symmetry in the environment configuration~\cite{grover2020does}. As in previous work~\cite{wang2017safety, grover2016before, grover2020does}, we define a deadlock as follows,

\begin{definition}
    \textbf{Deadlock:} A system of $k$ robots executing the controller given by Equation~\eqref{eq: best_response_example} enters a deadlock if, starting at time $t$, there exists at least one agent $i$ such that $ u^{i}_t = 0$ for some threshold $\beta>0$, typically measured in the order of a few seconds, while $x^i_t \notin \mathcal{X}_g$, for each robot $i \in [1,k]$. Mathematically, if there exists at least one pair $\left(x^i_t,u^i_t\right)\in\mathcal{D}^i(t)$ where the set $\mathcal{D}^i(t)$ is defined as follows
    \begin{align}
        \mathcal{D}^i(t)=\left\{\left(x^i_t,u^i_t\right):x^i_t\notin\mathcal{X}_g,\;\;u^i_t=0 \  \textnormal{for some} \ \beta > 0\right\}
        \label{eqn:deadlock_set}
    \end{align}
    then the system of these $k$ robots is said to be in deadlock. Furthermore, a trajectory that contains such $u^{i}_t$ that satisfy the deadlock condition \eqref{eqn:deadlock_set} is said to be \textit{inadmissible}.
    \label{def: deadlock}
\end{definition}

Here, $\beta$ is typically in the order of a few seconds~\cite{csenbacslar2023rlss}. Resolving or preventing deadlocks involves a two-stage procedure: $(i)$ detecting the deadlock and $(ii)$ resolving or preventing the deadlock. So far, the literature~\cite{wang2017safety, arul2023ds, zhou2017fast, grover2016before, impc} has focused on resolving deadlocks after they happen, which then require resolution strategies that are quite invasive. Commonly used strategies include perturbing the position of a robot according to the right-hand rule causing it to deviate from its preferred trajectories in a clockwise direction resulting in a sub-optimal controller. In the following section, we formalize an approach to detect, and the prevent, deadlocks via a minimally invasive perturbation. To aid readability, we visualize our theoretical framework in Figure\mbox{~\ref{fig: flowchart}}. We begin by defining minimally invasive deadlock prevention as follows:
\begin{figure*}[t]
    \centering
    \begin{subfigure}[h]{0.345\linewidth}
\includegraphics[width = \linewidth]{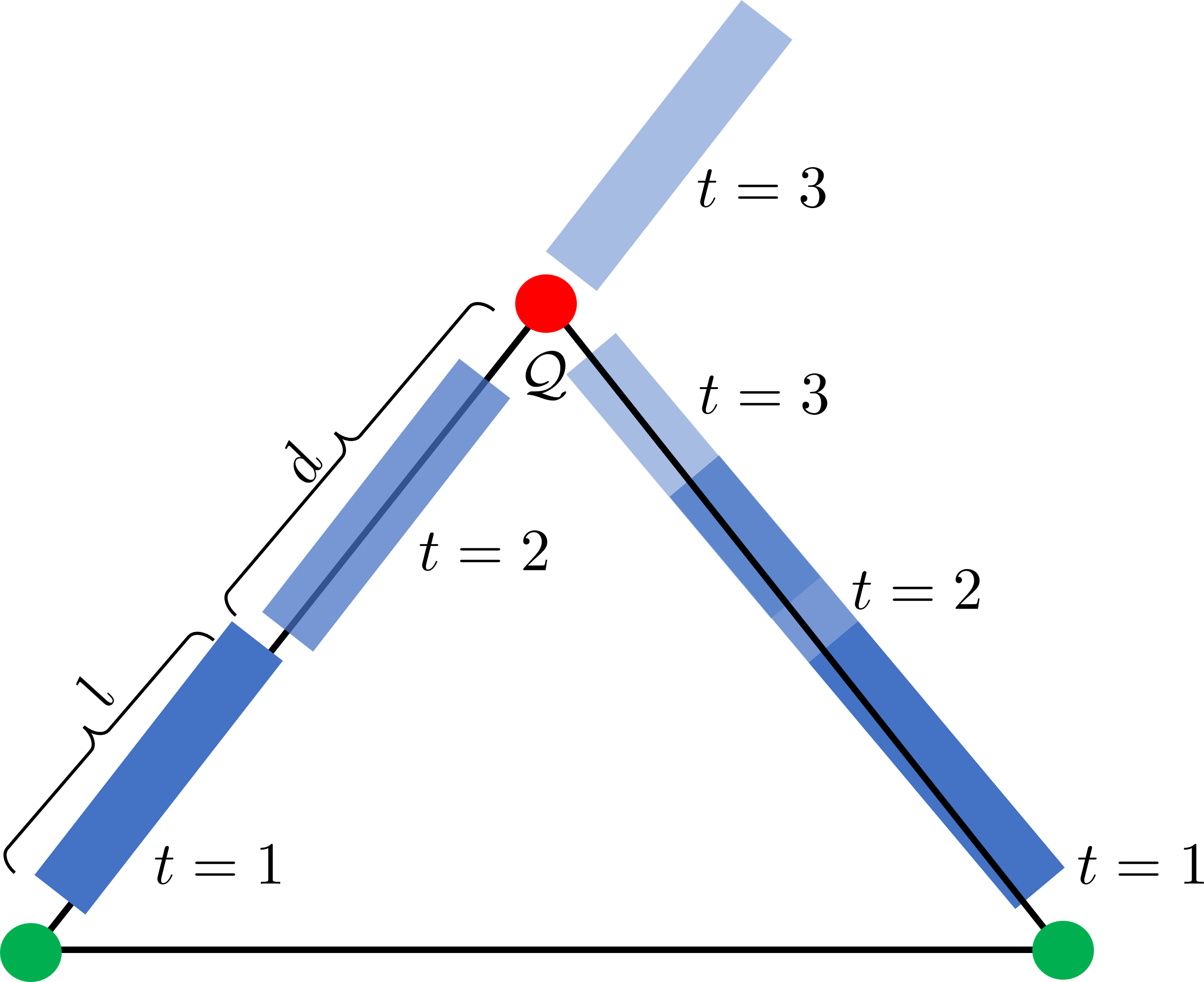}    
    \caption{}
    \label{fig: geom1}
    \end{subfigure}
    \begin{subfigure}[h]{0.315\linewidth}
\includegraphics[width = .95\linewidth]{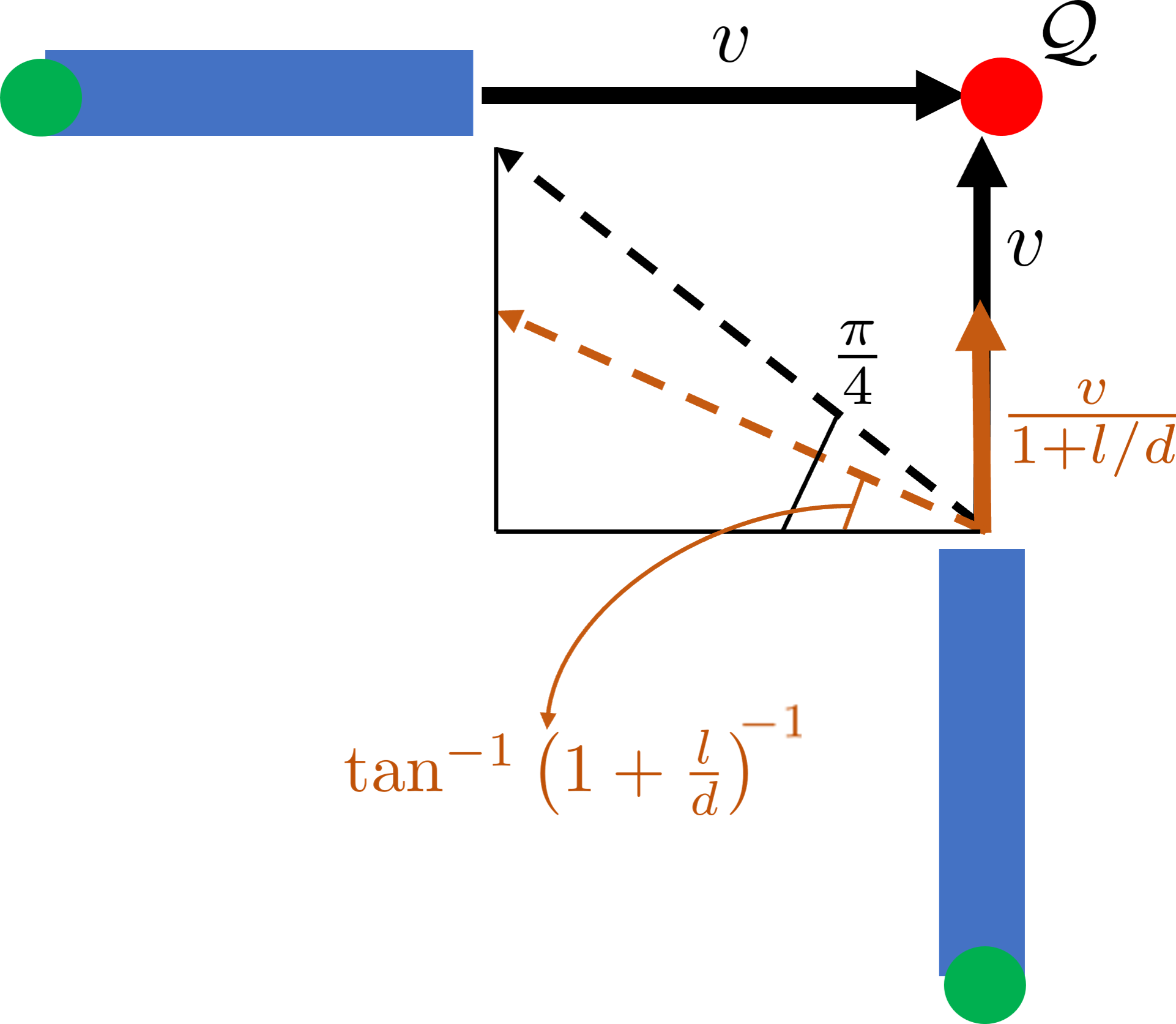}    
    \caption{}
    \label{fig: geom2}
    \end{subfigure}
    \begin{subfigure}[h]{0.315\linewidth}
\includegraphics[width = .8\linewidth]{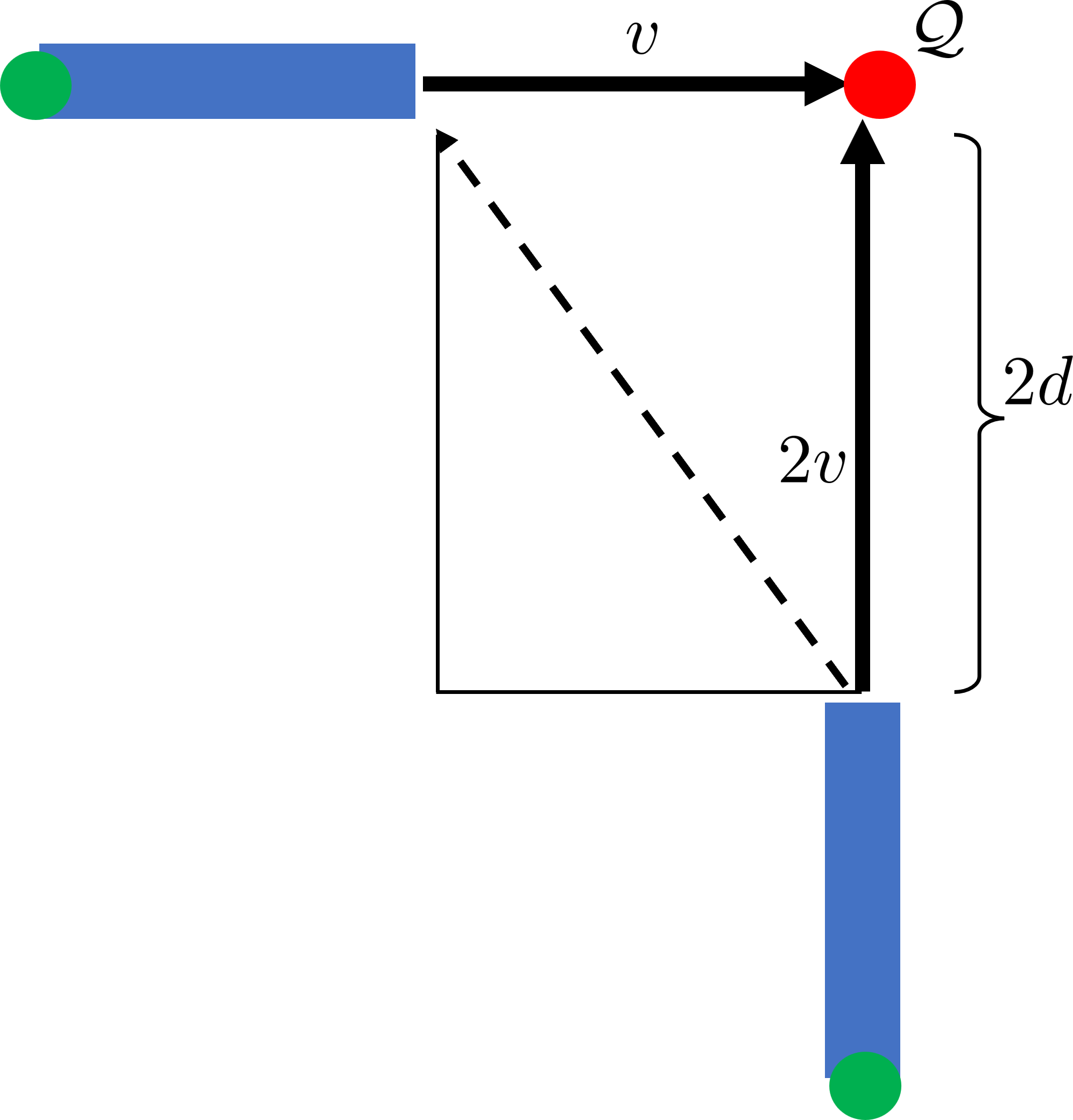}   
    \caption{}
    \label{fig: geom3}
    \end{subfigure}
    \caption{Deriving the threshold, $\ell_\textnormal{thresh}$, in a symmetrical social mini-game when $\ell_j(p_t^i,v_t^i) = 0$. Robots are blue car-like systems of length $l$. Green and red circles represent starting and goal positions, respectively. \textit{(left)} Step $1$: construct the worst case scenario when $\ell_j(p_t^i,v_t^i) = 0$. The second robot on the right must reach $\mathcal{Q}$ when the robot on the left passes fully through $\mathcal{Q}$. \textit{(middle) Step $2$:} Assuming the second robot slows down by $\frac{v}{1 + \frac{l}{d}}$, $\ell_\textnormal{thresh}$ is obtained via Equation~\eqref{eq: l_thresh}. \textit{(right)} Symmetric SMGs with arbitrary angles with the relative position vector configurations can be reduced to the SMG in Figure~\mbox{\ref{fig: geom2}}.}
    \label{fig: geom}
    % \vspace{-10pt}
\end{figure*}

\begin{definition}
    \textbf{Minimally Invasive Deadlock Prevention (or Resolution).} A deadlock preventive (or resolving) control strategy $u^i_t = \left[ v^i_t, \omega^i_t\right]^\top$ prescribed for robot $i$ at time $t$ with current heading angle $\theta^i_t$ is said to be minimally invasive if:
    \begin{enumerate}
        \item $\Delta \theta^i_t = \theta^i_{t+1} - \theta^i_t = 0$ (does not deviate from the preferred trajectory).
        \item $v_{t+1}^i = v_t^i +\delta_{\text{opt}}(t)$ where $\delta_{\text{opt}}(t) = \arg\min\left\Vert v_t^i + \delta \right\rVert, \delta \in \mathbb{R}$ such that a robot with speed $v_{t+1}^i$ prevents or resolves a deadlock.
        % In other words, $\delta_{\text{opt}}(t)$ can be computed by solving the following optimization problem
        % \begin{subequations}
        %         \begin{align}
        % \delta_{\text{opt}}(t)=&\underset{\delta\in\mathbb{R}}{\arg\min}\;\|v^i_t+\delta\|,\quad,\\
        % &v^i_{t+1}=v_t^i+\delta\\
        % &u^i_t\in\mathscr{U}^i,\;\;u^i_{t+1}\in\mathscr{U}^i\\
        % &\left(x^i_{t+1},u^i_{t+1}\right)\notin\mathcal{D}^i(t)
        % \label{eqn:minimally_invasive_optimization}
        % \end{align}
        %         \end{subequations}
        %         Note that the optimization problem \eqref{eqn:minimally_invasive_optimization} is a non-trivial problem to solve as the set $\mathcal{D}^i(t)$ is not known priori. Furthermore, even if the set $\mathcal{D}^i(t)$ is known apriori, finding $\delta_{\text{opt}}(t)$ could be computationally expensive especially when the set $\mathcal{D}^i(t)$ is non-convex.
        % \textcolor{red}{the earlier two points can be selling point for using heuristics. Also, you can highlight your earlier paper.}
        % \item $v_{t+1}^i = v_t^i +\delta$ where $\delta = \arg\min\left\Vert v_t^i + \delta \right\rVert, \delta \in \mathbb{R}$ such that $\ell_j(p_t^i,v_t^i +\delta) \geq \ell_\textnormal{thresh}$ (smallest change in speed that prevents a deadlock).
    \end{enumerate}
    \label{def: min_invasive}
\end{definition}
In other words, $\delta_{\text{opt}}(t)$ can be computed by solving the following optimization problem
        \begin{subequations}
                \begin{align}
        \delta_{\text{opt}}(t)=&\underset{\delta\in\mathbb{R}}{\arg\min}\;\|v^i_t+\delta\|,\\
        &v^i_{t+1}=v_t^i+\delta\\
        &u^i_t\in\mathscr{U}^i,\;\;u^i_{t+1}\in\mathscr{U}^i\\
        &\left(x^i_{t+1},u^i_{t+1}\right)\notin\mathcal{D}^i(t+1)
        \label{eqn:minimally_invasive_optimization}
        \end{align}
                \end{subequations}
                Note that the optimization problem \eqref{eqn:minimally_invasive_optimization} is a non-trivial problem to solve as the set $\mathcal{D}^i(t)$ is not known priori. Furthermore, even if the set $\mathcal{D}^i(t)$ is known apriori, finding $\delta_{\text{opt}}(t)$ could be computationally expensive especially when the set $\mathcal{D}^i(t)$ is non-convex.
A minimally invasive perturbation does not cause a robot to deviate from its preferred trajectory (condition ($1$)) only allowing it to speed up or slow down (condition ($2$)).

\begin{theorem}
Consider a symmetrical social mini-game, $\mathcal{S}$, as shown in Figure~\ref{fig: geom1}. We represent robots as single-integrator car-like objects with lengths $l^1 = l^2 = l$, speeds $v^1_t, v^2_t$, and the following system dynamics:

\[
\begin{bmatrix}
\dot{x} \\
\dot{y} \\
\dot{\theta}
\end{bmatrix}
=
\begin{bmatrix}
v \cos(\theta) \\
v \sin(\theta) \\
\omega
\end{bmatrix}
\]

\begin{figure*}[t]
    \centering
    \begin{subfigure}[t]{0.49\linewidth}
        \centering
        \includegraphics[width=\linewidth]{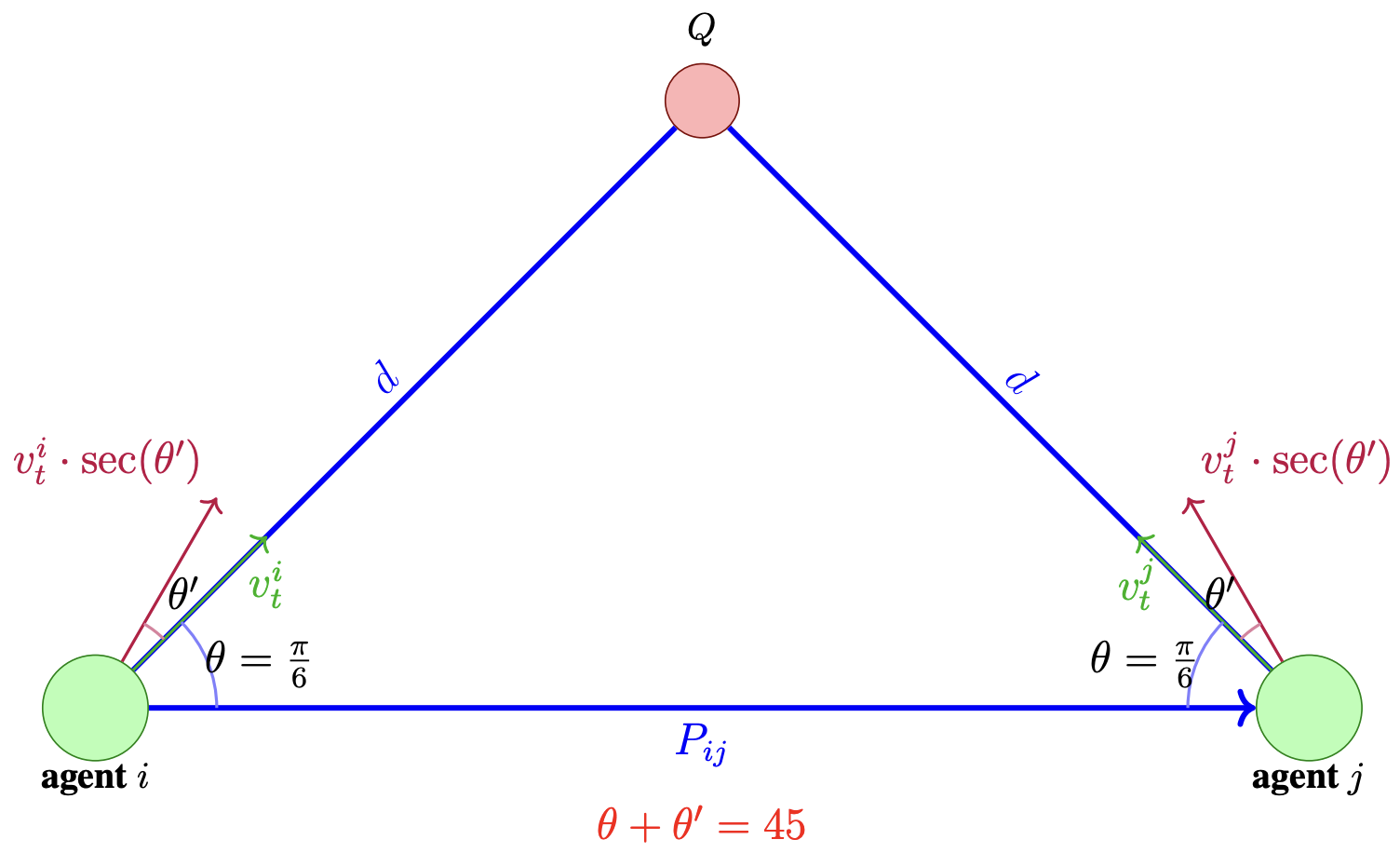}
        \caption{Velocity projection for $\theta = \frac{\pi}{6}$.}
        \label{fig: smg30}
    \end{subfigure}
    \begin{subfigure}[t]{0.49\linewidth}
        \centering
        \includegraphics[width=\linewidth]{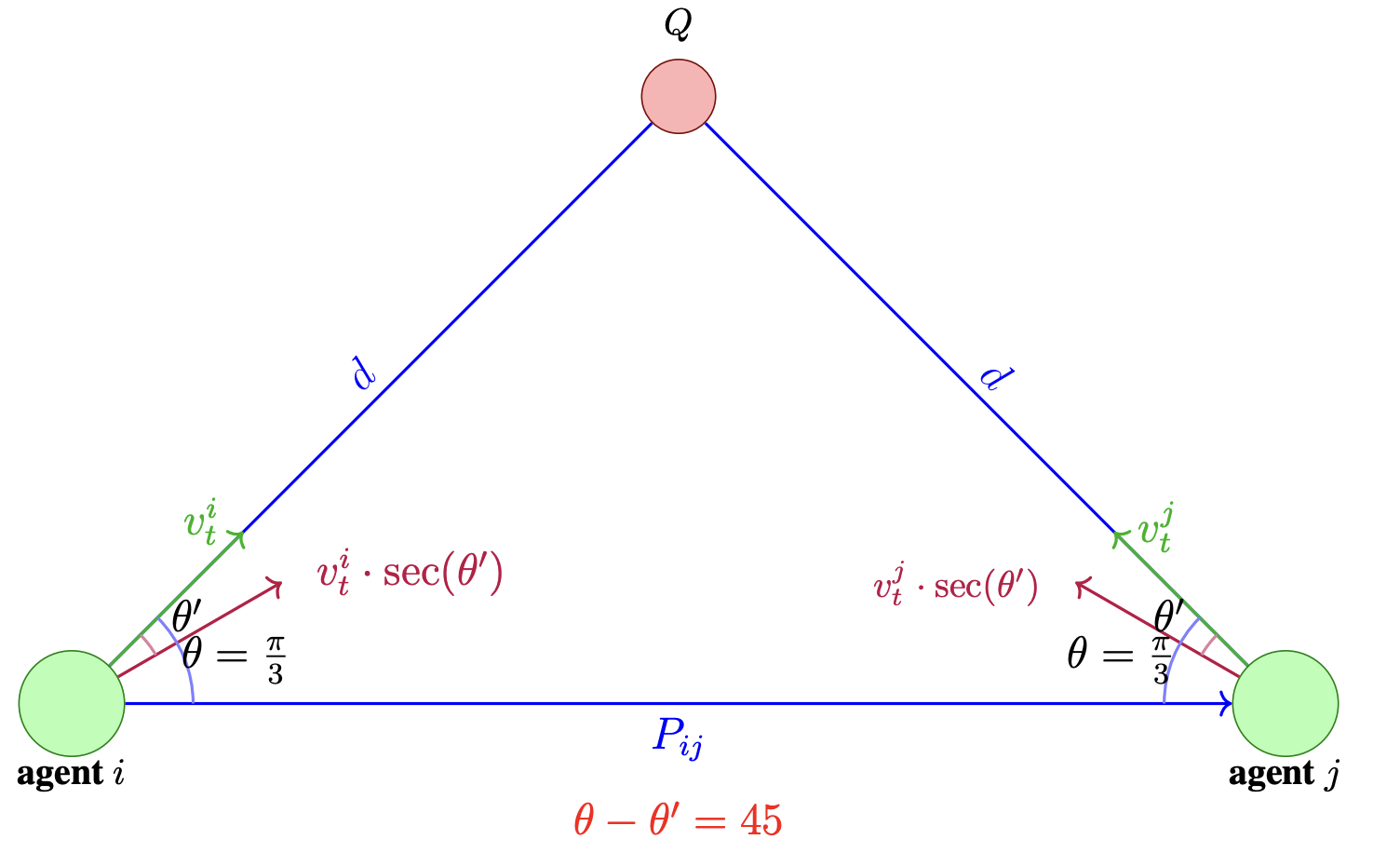}
        \caption{Velocity projection for $\theta = \frac{\pi}{3}$.}
        \label{fig: smg60}
    \end{subfigure}
    \caption{Velocity projection for two symmetrical SMGs with arbitrary angles $\theta$ with respect to the relative position vector. In these two examples, we examine when $\theta = \frac{\pi}{6}$ and $\theta = \frac{\pi}{3}$ (although any arbitrary $\left \lvert \theta \right\rvert < \frac{\pi}{2}$ may be chosen).}
    \label{fig: velocity_projection}
\end{figure*}

such that $0 < \left \lvert v^1_t - v^2_t \right\rvert \leq \epsilon$. Assume further that they are at distances $d^1 = d^2 = d$ from the point of collision $\mathcal{Q}$ (could be a doorway or intersection). For the robots to reach $\mathcal{Q}$ without colliding in a minimally invasive sense (Definition~\ref{def: min_invasive}), one of the robots must slow down (or speed up) by a factor of $\zeta \geq 2$ in the limit as $(d, \epsilon) \to (l, 0)$.
% such that $v^1_t \geq 2 v^2_t$ or $v^2_t \geq 2 v^1_t$.
    \label{thm: factor_of_2}
\end{theorem}

\begin{proof}
%     In order for robot $1$ to clear the distance $d$ in the time that robot $2$ reaches the $\mathcal{Q}$, robot $2$ must slow its speed by a factor of 
% \begin{equation}
% \zeta \geq \left(1 + \frac{l}{d}\right)\left(1 - \frac{\epsilon}{v^1_t}\right)    
% \end{equation}

In a symmetric SMG consisting of two robots, we assume that the speeds of the robots are nearly identical. In practice, two speeds will likely differ by some small amount, which we denote as $\epsilon$ giving $0 < \lvert v^1_t - v^2_t\rvert \leq \epsilon$. Suppose, without loss of generality, that robot $1$ is slightly faster than robot $2$ by $\epsilon$. Thus, $v^1_t - v^2_t = \epsilon \implies v^2_t = v^1_t -\epsilon$. In order for robot $1$ to clear the distance $d$ in the time that robot $2$ reaches the $\mathcal{Q}$, robot $1$ must increase its speed by a factor of $\zeta$. Therefore $v^1_{t+1} \gets \zeta v^1_{t}$ while $v^2_{t+1} = v^1_t -\epsilon$. We assume that speed changes are instantaneous as the agents enter the SMG.

Now time taken by robot 2 to reach Q is $t_2 = \frac{d}{v^2_{t+1}} = \frac{d}{v^1_t -\epsilon}$. And the time taken by robot 1 to \textit{clear} Q (that is, not just reach Q, but that the entire length $l$ should clear Q) is $t_1 = \frac{d + l}{v^1_{t+1}} = \frac{d + l}{\zeta v^1_t}$. Note that $t_2 \geq t_1$. Thus,

\begin{equation*}
    \frac{d}{v^1_t -\epsilon} \geq \frac{d + l}{\zeta v^1_t} 
\end{equation*}

Rearranging terms gives us Equation (13),

\begin{equation}
\zeta \geq \left(1 + \frac{l}{d}\right)\left(1 - \frac{\epsilon}{v^1_t}\right)     
\end{equation}

In the limit, $\zeta \to 2$ as $(d, \epsilon) \to (l, 0)$.
\end{proof}

\subsection{Deadlock Detection}

A state is represented as $x^i_t$ at time $t$ of which $\left(p^i_t, \theta^i_t,v^i_t, \omega^i_t\right) \in \mathbb{R}^2 \times \mathbb{S}^1\times \mathbb{R}^2\times \mathbb{R}$ represent the current position, heading, linear and angular velocities of the $i^\textrm{th}$ robot. Previous studies on deadlock detection~\cite{impc, grover2020does} show that deadlocks arise from symmetry in the environment configuration. Symmetry is geometrically defined in terms of the initial positions, goals, and velocities of the agents. Example of symmetrical configurations are given in Figure~\ref{fig: geom}. Figures~\ref{fig: geom1} and~\ref{fig: geom2} depict a symmetrical SMG with $\theta = \frac{\pi}{4}$ whereas Figure~\ref{fig: geom3} depicts a symmetrical SMG with arbitrary angles.

An important observation is that any symmetrical SMG with arbitrary angle $\theta$ can be reduced to a symmetrical SMG with angle $\frac{\pi}{4}$. In any symmetrical SMG with arbitrary angle $\theta, \left \lvert \theta \right \rvert < \frac{\pi}{2}$ (agents diverge away from each other for $\left \lvert \theta \right \rvert \geq \frac{\pi}{2}$), with respect to the relative position vector, suppose two robots $i$ and $j$ have nearly identical velocities $v^i_t \approx v^j_t$ (otherwise, it would not be a symmetrical SMG) or $\left \lvert v^i_t - v^j_t \right \rvert \leq \epsilon$. Then, without loss of generality, we can project a component of each robots velocity such that it subtends an $\theta^\prime$ with the original corresponding velocity vector. The advantage of such a projection is that it eliminates the dependency on the angle, with the only dependency remaining on the scaling factor $\zeta$. We set $\theta'$ as follows:

\begin{enumerate}
    \item \textbf{Case 1:} When the arbitrary angle $\theta < \frac{\pi}{4}$, then we set $\theta' = \frac{\pi}{4} - \theta$. 

    \item \textbf{Case 2:} When the arbitrary angle $\frac{\pi}{2}>\theta \geq \frac{\pi}{4}$, then we set $\theta' = \theta - \frac{\pi}{4}$. 

\end{enumerate}
\begin{figure*}[t]
    \centering
    \includegraphics[width = \linewidth]{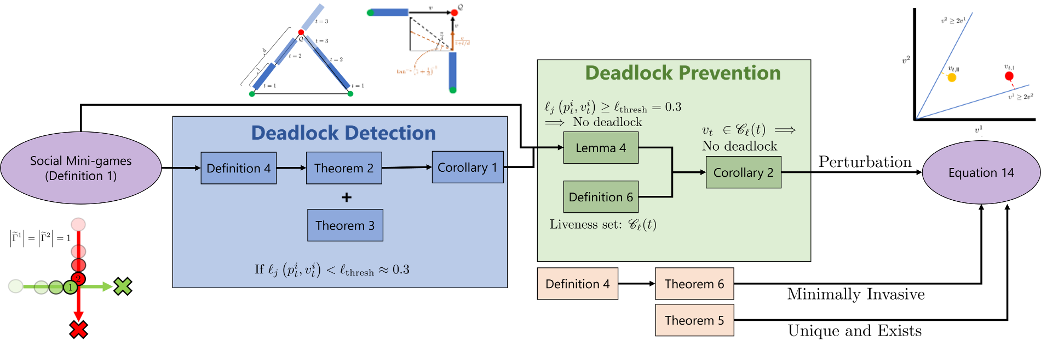}
    \caption{\textbf{Flowchart of the deadlock detect-and-prevent logic}. This diagram illustrates at a high level the process for detecting and preventing deadlocks in our approach. It includes the foundational definition of social mini-games, followed by the prevention strategies enforced by liveness conditions. The interaction between these elements is visualized through connections between theoretical results in Section~\ref{sec: GT-resolution}.}
    \label{fig: flowchart}
\end{figure*}

% \begin{figure*}
%     \centering
% \begin{tikzpicture}[
%     node distance=1cm and 1cm,
%     mynode/.style={draw, align=center, minimum width=2.5cm, minimum height=1cm, fill=blue!20},
%     myellipse/.style={draw, ellipse, align=center, minimum width=2.5cm, minimum height=1cm, fill=purple!20},
%     myarrow/.style={-Stealth},
%     mylabel/.style={font=\scriptsize, midway, fill=white}
% ]
%     % Nodes
%     \node[myellipse] (social) {Social Mini-games\\(Definition 1)};
%     \node[mynode, right=of social] (detection) {Deadlock Detection\\Theorem 1};
%     \node[mynode, below=of detection] (lemma) {Lemma 2};
%     \node[mynode, right=of lemma] (definition5) {Definition 5};
%     \node[mynode, right=of detection] (prevention) {Deadlock Prevention};
%     \node[myellipse, right=of prevention] (equation) {Equation 12};
%     \node[mynode, below=of definition5] (definition4) {Definition 4};
%     \node[mynode, right=of definition4] (theorem3) {Theorem 3};
    
%     % Edges
%     \draw[myarrow] (social) -- (detection);
%     \draw[myarrow] (detection) -- (prevention);
%     \draw[myarrow] (lemma) -- (prevention);
%     \draw[myarrow] (definition5) -- (lemma);
%     \draw[myarrow] (definition4) -- (theorem3);
%     \draw[myarrow] (theorem3) -- (prevention);
%     \draw[myarrow] (prevention) -- (equation);
    
%     % Labels (for the arrows, if needed)
%     % \draw[myarrow] (node1) -- node[mylabel] {label} (node2);
% \end{tikzpicture}
%     \caption{Caption}
%     \label{fig:enter-label}
% \end{figure*}

The new transformed velocities are given by $v^{i, \prime}_t = v^i_t \sec\theta^\prime$ and $v^{j, \prime}_t = v^j_t \sec \theta^\prime$. Figure~\ref{fig: velocity_projection} demonstrates the velocity projection for two symmetrical SMGs with arbitrary angles $\theta$ with respect to the relative position vector. In Figures~\ref{fig: smg30} and~\ref{fig: smg60}, we examine when $\theta = \frac{\pi}{6}$ and $\theta = \frac{\pi}{3}$, respectively, although any arbitrary $\left \lvert \theta \right\rvert < \frac{\pi}{2}$ may be chosen.
% The key idea is that these projected velocities also constitute a symmetrical SMG.  In fact, notice that this resulting SMG is precisely the one in Figure~\ref{fig: geom2}. 
We now define a liveness function as follows:

% Social mini-games often lead to geometrically symmetrical configurations, which in turn produces a strong likelihood of a deadlock~\cite{grover2020does}. This gives us a way of detecting deadlocks by leveraging the properties of social mini-games. Define a liveness function as follows:

% \begin{definition}
% A liveness function $\ell_j:\mathbb{R}^2\times\mathbb{R}^2\rightarrow\mathbb{R}$ between agents $i$ and $j$ ($i\neq j$) as follows:
% \begin{align}
% \ell_j\left(p_t^i,v_t^i\right)=\cos^{-1}\left(\frac{\left \langle \overrightarrow{p_t^i-p^j_t},\overrightarrow{v^i_t-v^j_t}\right \rangle}{\left \lVert \overrightarrow{p^i_t-p^j_t}\right \rVert \left \lVert \overrightarrow{v^i_t-v^j_t}\right \rVert+\epsilon}\right)
% \label{eqn:liveness_formula}
% \end{align}
% where $\left \langle a, b \right \rangle$ denotes dot product between vectors $a$ and $b$ and $\|.\|$ denotes the Euclidean norm, $\ell_j\in[0,\pi]$, and $\epsilon>0$ ensures that the denominator is positive. Furthermore, note that $\ell_j(p_t^i,v_t^i)=\ell_i(p_t^j,v_t^j)$ for $i\neq j$.
% \end{definition}

\begin{definition}
\textbf{Liveness Function}: A liveness function $\ell_j:\mathbb{R}^2\times\mathbb{R}^2\rightarrow\mathbb{R}$ between agents $i$ and $j$ ($i\neq j$) as follows:
\begin{align}
\ell_j(p_t^i,v_t^i)=\cos^{-1}\left(\frac{\left \langle \overrightarrow{p_t^i-p^j_t},\overrightarrow{{ v^{i, \prime}_t}- {v^{j, \prime}_t}}\right \rangle}{\left \lVert \overrightarrow{p^i_t-p^j_t}\right \rVert \left \lVert \overrightarrow{{ v^{i, \prime}_t}- { v^{j, \prime}_t}}\right \rVert+\epsilon}\right)
\label{eqn:liveness_formula}
\end{align}
where ${v^{i, \prime}_t} = v^i_t\cdot \sec(\theta'), {v^{j, \prime}_t} = v^j_t\cdot \sec(\theta')$ correspond to the velocity projection component of each robot's velocity such that it subtends an angle $\theta^\prime = \pm\left(\frac{\pi}{4} - \theta\right)$ with the original velocity vectors, $v^i_t$ and $v^j_t$, as defined by the velocity transformation procedure, where $\left\lvert\theta\right \rvert < \frac{\pi}{2}$ is the arbitrary SMG angle. Further, $\left \langle a, b \right \rangle$ denotes dot product between vectors $a$ and $b$ and $\|.\|$ denotes the Euclidean norm, $\ell_j\in[0,\pi]$, and $\epsilon>0$ ensures that the denominator is positive. Lastly, note that $\ell_j(p_t^i,v_t^i)=\ell_i(p_t^j,v_t^j)$ for $i\neq j$.
\label{def: liveness_function}
\end{definition}

% and $\ell_j\left( x^i_t, v^i_t\right) < \ell_\textnormal{thresh}$ implies that $i$ and $j$ are on a collision course and therefore have the potential for entering a deadlock. 
% This observation follows from the studies on symmetry as a cause of deadlocks~\cite{impc, grover2020does}.
% Note that $\ell_j(p_t^i,v_t^i)=\ell_i(p_t^j,v_t^j)$ for $i\neq j$. 

The liveness function is meant to geometrically capture the symmetry of an SMG (refer to Figure~\mbox{\ref{fig: geom}}). Formally, the liveness function measure the angle between the relative displacement ($\overrightarrow{p^{12}_t} = \overrightarrow{p^1_t - p^2_t}$) and the relative linear velocity ($\overrightarrow{v^{12,\prime}} = \overrightarrow{v^{1, \prime}_t - v^{2,\prime}_t}$) vectors. The angle between $\overrightarrow{p^{12}_t}$ and $\overrightarrow{v^{12,\prime}_t}$ determines the symmetry of the SMG. In the case of perfect symmetry ($v^{1,\prime}_t \approx v^{2,\prime}_t$), we can easily show using vector geometry that $\overrightarrow{p^{12}_t}$ and $\overrightarrow{v^{12,\prime}_t}$ will be nearly perfectly aligned (parallel) in which case the angle between them will be approximately $0$ implying the dot product will be nearly $1$. The dot product decreases as the symmetry decreases (as $v^{1,\prime}_t$ and $v^{2,\prime}_t$ diverge). The $\epsilon$ in the denominator is the same $\epsilon$ used in Theorem~\mbox{\ref{thm: factor_of_2}} which represents the practicality of the fact that there always exists a small difference between $v^{1,\prime}_t$ and $v^{2,\prime}_t$, no matter how close they may be. We now present the following theorem
\begin{theorem}
Consider the social mini-game, $\mathcal{S}$, defined in Theorem~\ref{thm: factor_of_2}. Then at time $t$,
% \begin{equation}
%     \ell_j(p_t^i,v_t^i)=\ell_i(p_t^j,v_t^j) = \frac{\pi}{4} - \tan^{-1}\frac{1}{\zeta}
% \end{equation}
\begin{equation}
    \mathcal{S} \ \textnormal{exists} \iff \ell_j(p_t^i,v_t^i)=\ell_i(p_t^j,v_t^j) \leq \ell_\textnormal{thresh}
\end{equation}

where $\ell_\textnormal{thresh} := \frac{\pi}{4} - \tan^{-1}\frac{1}{2}$.
    \label{thm: smg->l<tau}
\end{theorem}
\begin{proof}
We prove the forward direction first. Using basic trigonometry, the scaling factor $\zeta$ can be used to determine the threshold $\ell_\textnormal{thresh}$ as shown in Figure~\ref{fig: geom2}. Formally,

\begin{equation}
    \ell_j = \frac{\pi}{4} - \tan^{-1}\frac{1}{\zeta}
    \label{eq: l_thresh}
\end{equation}
Using Theorem~\ref{thm: factor_of_2}, in the limit $\lim_{d\to l, \epsilon\to 0} \zeta = 2 \implies \ell_j = \ell_i = \frac{\pi}{4} - \tan^{-1}\frac{1}{2}$. Now we prove the backward direction. Assume $\ell_j = \ell_i \leq \frac{\pi}{4} - \tan^{-1}\frac{1}{2}$. We can construct a scenario $S$ geometrically similar to $\mathcal{S}$. It will be equivalent to $\mathcal{S}$ when we can show that one robot will not be able to clear $\mathcal{Q}$ in the time when the other agent reaches $\mathcal{Q}$ (by Definition~\ref{def: social_minigame}) which happens when $\zeta \leq 2$ (Theorem~\ref{thm: factor_of_2}). As $\ell_j = \ell_i \leq 0.3 \implies \zeta \leq 2$ is true, $S$ is equivalent to $\mathcal{S}$.
\end{proof}

We empirically verify in simulation that this threshold, $\ell_\textnormal{thresh} \approx 0.3$, in fact, works well in practice, as shown in Figure~\ref{fig: vel}. Recall that social mini-games often yield a deadlock due to their geometrically symmetric configurations~\cite{grover2020does}. Theorem~\ref{thm: smg->l<tau} gives us a way to monitor for social mini-games and therefore, deadlocks. 

\begin{corollary}
    \textbf{Deadlock Detection:} A system of $2$ robots may result in a deadlock if $\ell_j(p_t^i,v_t^i)=\ell_i(p_t^j,v_t^j) \leq \ell_\textnormal{thresh}$, where $\ell_\textnormal{thresh}$ is defined as in Theorem~\ref{thm: smg->l<tau}.
\end{corollary}
\begin{proof}
    % Using Conjecture~\ref{conj: non_symetric} and Theorem~\ref{thm: smg->l<tau}.
    The proof follows from Theorem~\ref{thm: smg->l<tau}.
\end{proof}

\paragraph{Equivalence of Equations~\eqref{eq: l_thresh} and~\eqref{eqn:liveness_formula}:} It is important to observe that Theorem~\ref{thm: smg->l<tau} and Definition~\ref{def: liveness_function} are equivalent. Specifically, we will show that Equations~\eqref{eq: l_thresh} and~\eqref{eqn:liveness_formula} are equivalent. First, from Definition~\ref{def: liveness_function}, assuming W.L.O.G that $\theta < \frac{\pi}{4} \implies \theta + \theta' = \frac{\pi}{4}$, the liveness function can be equivalently written as,

{\small\begin{equation}
    \begin{split}
        \ell_j &= \cos^{-1}\left(\frac{\cos(\theta + \theta') \cdot \left(1 + \frac{1}{\zeta}\right)}{\sqrt{1 + \frac{1}{\zeta^2} + \frac{2}{\zeta}\cos\left(2\left(\theta + \theta'\right)\right)}}\right)\\
        &= \cos^{-1}\left( \sqrt{\frac{1}{2} + \frac{\zeta}{\zeta^2+1}} \right)
    \end{split}
\end{equation}}

% We aim to prove the identity:
% \[
% \cos^{-1}\left(\sqrt{\frac{1}{2} + \frac{1}{2\zeta}}\right) = \frac{\pi}{4} - \tan^{-1}\left(\frac{1}{\zeta}\right).
% \]

\noindent We would substitute $\cos(\theta - \theta')$ if $\frac{\pi}{4} \leq \theta < \frac{\pi}{2}$. Let \(\alpha = \cos^{-1}\left(\sqrt{\frac{1}{2} +\frac{\zeta}{\zeta^2+1}}\right)\), then
$\cos(\alpha) = \sqrt{\frac{1}{2} + \frac{\zeta}{\zeta^2+1}}$. Set $\beta = \tan^{-1}\left(\frac{1}{\zeta}\right)$. It follows that:

\begin{equation}
    \begin{split}
        \tan(\beta) &= \frac{1}{\zeta},\\
        \cos(\beta) &= \frac{\zeta}{\sqrt{\zeta^2 + 1}},\\
        \sin(\beta) &= \frac{1}{\sqrt{\zeta^2 + 1}}.
    \end{split}
\end{equation}

\noindent Further, using the trigonometric identity, $\cos\left(\frac{\pi}{4} - \beta\right) = \frac{\cos\beta + \sin\beta}{\sqrt{2}}$, we get:

\begin{equation}
    \begin{split}
        \cos\left(\frac{\pi}{4} - \beta\right) &= \frac{\frac{\zeta}{\sqrt{\zeta^2 + 1}} + \frac{1}{\sqrt{\zeta^2 + 1}}}{\sqrt{2}}\\
        &= \frac{\frac{\zeta + 1}{\sqrt{\zeta^2 + 1}}}{\sqrt{2}} \\
        &= \sqrt{\frac{1}{2} + \frac{\zeta}{\zeta^2+1}} \\
        &= \cos(\alpha).
    \end{split}
\end{equation}

\noindent or $\frac{\pi}{4} - \beta = \alpha$. Therefore,

\begin{equation}
    \begin{split}
&\cos^{-1}\left(\frac{\cos(\theta + \theta') \cdot \left(1 + \frac{1}{\zeta}\right)}{\sqrt{1 + \frac{1}{\zeta^2} + \frac{2}{\zeta}\cos\left(2\left(\theta + \theta'\right)\right)}}\right)= \frac{\pi}{4} - \beta\\
&=\frac{\pi}{4} - \tan^{-1}\left(\frac{1}{\zeta}\right)
    \end{split}
\end{equation}

\noindent And we already have that,

\begin{equation}
    \begin{split}
\ell_j(p_t^i,v_t^i)&=\cos^{-1}\left(\frac{\left \langle \overrightarrow{p_t^i-p^j_t},\overrightarrow{ v^{i, \prime}_t- v^{j, \prime}_t}\right \rangle}{\left \lVert \overrightarrow{p^i_t-p^j_t}\right \rVert \left \lVert \overrightarrow{{ v^{i, \prime}_t}- { v^{j, \prime}_t}}\right \rVert+\epsilon}\right)\\
&= \cos^{-1}\left(\frac{\cos(\theta + \theta') \cdot \left(1 + \frac{1}{\zeta}\right)}{\sqrt{1 + \frac{1}{\zeta^2} + \frac{2}{\zeta}\cos\left(2\left(\theta + \theta'\right)\right)}}\right)    
    \end{split}
\end{equation}

\noindent Therefore, finally,

\begin{equation}
    \begin{split}
\ell_j(p_t^i,v_t^i)&=\cos^{-1}\left(\frac{\left \langle \overrightarrow{p_t^i-p^j_t},\overrightarrow{ v^{i, \prime}_t- v^{j, \prime}_t}\right \rangle}{\left \lVert \overrightarrow{p^i_t-p^j_t}\right \rVert \left \lVert \overrightarrow{{ v^{i, \prime}_t}- { v^{j, \prime}_t}}\right \rVert+\epsilon}\right) \\
&= \frac{\pi}{4} - \tan^{-1}\left(\frac{1}{\zeta}\right)
    \end{split}
\end{equation}

\noindent In other words, we showed that Equation~\eqref{eqn:liveness_formula} in Definition~\ref{def: liveness_function} is equivalent to Equation~\eqref{eq: l_thresh} in Theorem~\ref{thm: smg->l<tau}.

\subsection{Deadlock Prevention} 

From Theorem~\ref{thm: smg->l<tau}, we can derive the following lemma,

\begin{lemma}
If at time $t$, $x^i_t \in \widetilde \Gamma^i$ and $\ell_j\left( x^i_t, v^i_t\right) \geq \ell_\textnormal{thresh}$, then agent $i$ is \ul{not} in a deadlock with agent $j$, where $j\in\mathcal{N}^i\left(x^i_t\right)$. 
    \label{lem: smg->deadlock}
\end{lemma}
\begin{proof}
From Theorem~\ref{thm: smg->l<tau}, a social mini-game implies $\ell_j\left( x^i_t, v^i_t\right) < \ell_\textnormal{thresh}$. Therefore, by the contrapositive, a preventive solution can be devised by perturbing both robots such that $\ell_j\left( x^i_t, v^i_t\right) \geq \ell_\textnormal{thresh}$, thereby resolving the social mini-game. If there is no social mini-game occurring at time $t$, then by Definition~\ref{def: social_minigame}, there exists some $\widetilde \Gamma^i$ that is not in conflict with any other agent $j$ and therefore, collision-free. By our assumption of small-time controllability of~\eqref{eq: control_affine_dynamics} and Theorem~\ref{thm: laumond}, there exists an admissible, and collision-free, path between configurations in $\mathcal{X}_I$ and $\mathcal{X}_G$.
% Also by definition of a preferred trajectory, for every state $x^i_t\in\widetilde \Gamma^i$, $\exists t^\prime \leq T$ such that $x^i_t \in \mathcal{R}\left( x^i_{t-t^\prime}, \mathcal{U}^i, t^\prime\right) \cap \widetilde \Gamma^i$. Then $x^i_T \in \mathcal{R}\left( x^i_{t-t^\prime}, \mathcal{U}^i, t^\prime\right) \cap \widetilde \Gamma^i$ for some $x^i_{t-t^\prime}, t^\prime$. System~\eqref{eq: control_affine_dynamics} is therefore small-time local controllable if agent $i$ follows $\widetilde \Gamma^i$, implying that a deadlock cannot occur.
\end{proof}

We propose a perturbation strategy that can be integrated as a constraint (CBF constraint for double-integrator or higher order dynamics) and combined with existing controllers, such as MPC or DWA, to resolve deadlocks. Our decentralized perturbation strategy is designed to be minimally invasive.
% in the following sense:

% \begin{definition}
%     \textbf{Minimally Invasive Deadlock Prevention (or Resolution).} A deadlock preventive (or resolving) control strategy $u^i_t = [v^i_t, \omega^i_t]^\top$ prescribed for robot $i$ at time $t$ is said to be minimally invasive if:
%     \begin{enumerate}
%         \item $\Delta \theta^i_t = 0$ (does not deviate from the preferred trajectory).
%         \item $v_{t+1}^i = v_t^i +\delta$ where $\delta = \arg\min\left\Vert v_t^i + \delta \right\rVert, \delta \in \mathbb{R}$ such that a robot with speed $v_{t+1}^i$ prevents or resolves a deadlock.
%         % \item $v_{t+1}^i = v_t^i +\delta$ where $\delta = \arg\min\left\Vert v_t^i + \delta \right\rVert, \delta \in \mathbb{R}$ such that $\ell_j(p_t^i,v_t^i +\delta) \geq \ell_\textnormal{thresh}$ (smallest change in speed that prevents a deadlock).
%     \end{enumerate}
    
%     \label{def: min_invasive}
% \end{definition}
In particular, we design a perturbation that acts on $v^i_t$, and not on $p^i_t$, through the notion of \emph{liveness sets}, which are analogous to the safety sets in CBFs. Formally,

\begin{definition}
At any time $t$, given a configuration of $k$ robots, $x^i_t \in \mathcal{X}$ for $i\in [1,k]$, a \textbf{liveness set} is defined as a union of convex sets, $\mathscr{C}_\ell(t) \subseteq \mathbb{R}^k$ of joint speed $v_t = \left[v^1_t, v^2_t, \ldots, v^k_t\right]^\top$ such that $v^i_t \geq \zeta v^j_t$ for all distinct pairs $i,j$, $ \zeta \geq 2$. 

    \label{def: liveness_sets}
\end{definition}
As $v^i_t \geq \zeta v^j_t \equiv v^j_t \geq \zeta v^i_t$, we can permute the order of agents as $k!$, each resulting in a different convex hull. We take the union of these convex hulls to generate $\mathscr{C}_\ell(t)$.
\begin{corollary}
If $x^i_t \in \widetilde \Gamma^i$, then $v_t \ \in \mathscr{C}_\ell(t)$ guarantees that the system of robots are deadlock-free.
\label{cor: v_t}
\end{corollary}

\begin{proof}
From Definition~\mbox{\ref{def: liveness_sets}}, since $v^i_{t} \in \mathscr{C}_\ell(t)$, we have $v^i_t \geq \zeta v^j_t$. According to Theorem\mbox{~\ref{thm: factor_of_2}}, we select $\zeta \geq 2$. Then, according to Theorem\mbox{~\ref{thm: smg->l<tau}}, $\zeta \geq 2 \implies \ell_j = \ell_i \geq \frac{\pi}{4} - \tan^{-1}\frac{1}{2}$. Finally, by Lemma\mbox{~\ref{lem: smg->deadlock}}, $\ell_j \geq \frac{\pi}{4} - \tan^{-1}\frac{1}{2}$ implies that robot $i$ is not in a deadlock, \textit{i.e.,} $\left \{  
x^i_{t}, u^i_{t} \not\in \mathcal{D}^i(t)\right \}$.
\end{proof}

% \begin{proof}
% If every agent at $t=0$ is collision-free, then perturb $u^i_t$ for all $i$ such that $\mathcal{R}\left( x^i_{t}, \mathcal{U}^i, \Delta t\right) \cap \widetilde \Gamma^i \neq \mathcal{R}\left( \widehat x^i_{t}, \mathcal{U}^i, \Delta t\right) \cap \widetilde \Gamma^i$ and $\mathcal{C}^i\left( \widehat x^i_t \right) \cap \mathcal{C}^j\left( \widehat  x^j_t \right)\neq \emptyset \ \forall \ t \in [t, t+\Delta t]$, where $\widehat x^i_{t+\Delta t} \in \mathcal{R}\left( \widehat x^i_{t}, \mathcal{U}^i, \Delta t\right) \cap \widetilde \Gamma^i$. 
% Trivially, one possible choice is $u^j_t = 0$ for all $j\neq i$ so that $\mathcal{R}\left( \widehat x^j_{t}, \mathcal{U}^j, \Delta t\right) \cap \widetilde \Gamma^j = \left \{ x^j_t \right \}$ and $\mathcal{C}^i\left( x^i_t \right) \cap \mathcal{C}^j\left( \widehat  x^j_t \right)\neq \emptyset \ \forall \ t \in [t, t+\Delta t]$ where $\Delta t < \beta$.

% \end{proof}

From Corollary~\ref{cor: v_t}, if each $v^i_t$ is such that the joint velocity $v_t \in \mathscr{C}_\ell(t)$, then there is no deadlock. If, however, $v_t \notin \mathscr{C}_\ell(t)$, then robot $i$ will adjust $v^i_t$ such that $v_t$ is projected on to the nearest point in $\mathscr{C}_\ell(t)$ via,

\begin{equation}
    \widetilde v_t = \arg\min_{\mu \in \mathscr{C}_\ell(t)} \left \lVert v_t - \mu\right \rVert_2
    \label{eq: perturbation}
\end{equation}

\begin{theorem}
    A solution to the optimization problem~\eqref{eq: perturbation} always exists and is unique if $v^i_t \neq v^j_t$ for any robots $i$ and $j$ with $i \neq j$.
\end{theorem}
\begin{proof}

The set $\mathscr{C}_\ell(t)$ can be described as the complement of an open convex polytopic set $P$ whose boundary $\partial P$ is a subset of $\mathscr{C}_\ell(t)$, that is, $\partial P \subseteq \mathscr{C}_\ell(t)$ and in particular, $\partial P \subseteq \partial \mathscr{C}_\ell(t)$. Existence of a minimum distance projection from any (interior) point of the (open) polytopic set $P$ to its boundary $\partial P \subseteq \partial \mathscr{C}_\ell(t)$ is trivially satisfied given as the set of edges, $\mathcal{K}$, that determine $\partial P$ is a finite set, which always yields a minimum. Formally, the Euclidean distance function $q(y) = \lVert x-y\rVert$, from a point $y \in \partial P$ to a given point $x \in \textrm{int}(P)$ is a continuous function, which always attains its minimum in $\partial P$ which is a compact (closed and bounded) set comprising a finite number of edges. 

Next, we will show that the minimal distance projection to $\partial P \subseteq \partial \mathscr{C}_\ell(t)$ is unique. First, we note that the projection of a point to an edge always exists and is unique (the latter projection either corresponds to the minimal distance projection to the line containing the line segment, if the latter projection belongs to the line segment, or one of the endpoints of the segment, otherwise). Therefore, there exists a unique minimizer, $d_k$, to the edge $k \in \mathcal{K}$. By extension, there exists a unique minimizer $d_j, j\neq k \in [\mathcal{K}]$ to each edge in  $\mathcal{K}$. Define the set $\mathcal{D} = \left\{d_j, j\neq k \in [\mathcal{K}]\right\}$. As the set $\mathcal{D}$ is finite, a minimizer will always exist. To show uniqueness, let us assume on the contrary that there exist $j, k \in [\mathcal{K}]$ with $j \neq k$ such that $d_j = d_k$ (this can, for instance, happen if when $P$ is regular and $x_c$ is the center of $P$, which is where agents have identical velocities). However, $d_j = d_k$ implies that $v^j_t = v^k_t$, which contradicts our assumption that $v^i_t \neq v^j_t$ for any robots $i$ and $j$ with $i \neq j$.

\end{proof}

\begin{theorem}
The deadlock-preventing control strategy for carrying out Equation~\ref{eq: perturbation} is minimally invasive.
\label{thm: min_invasive}
\end{theorem}

\begin{figure}[t]
    \centering
    \includegraphics[width=.8\columnwidth]{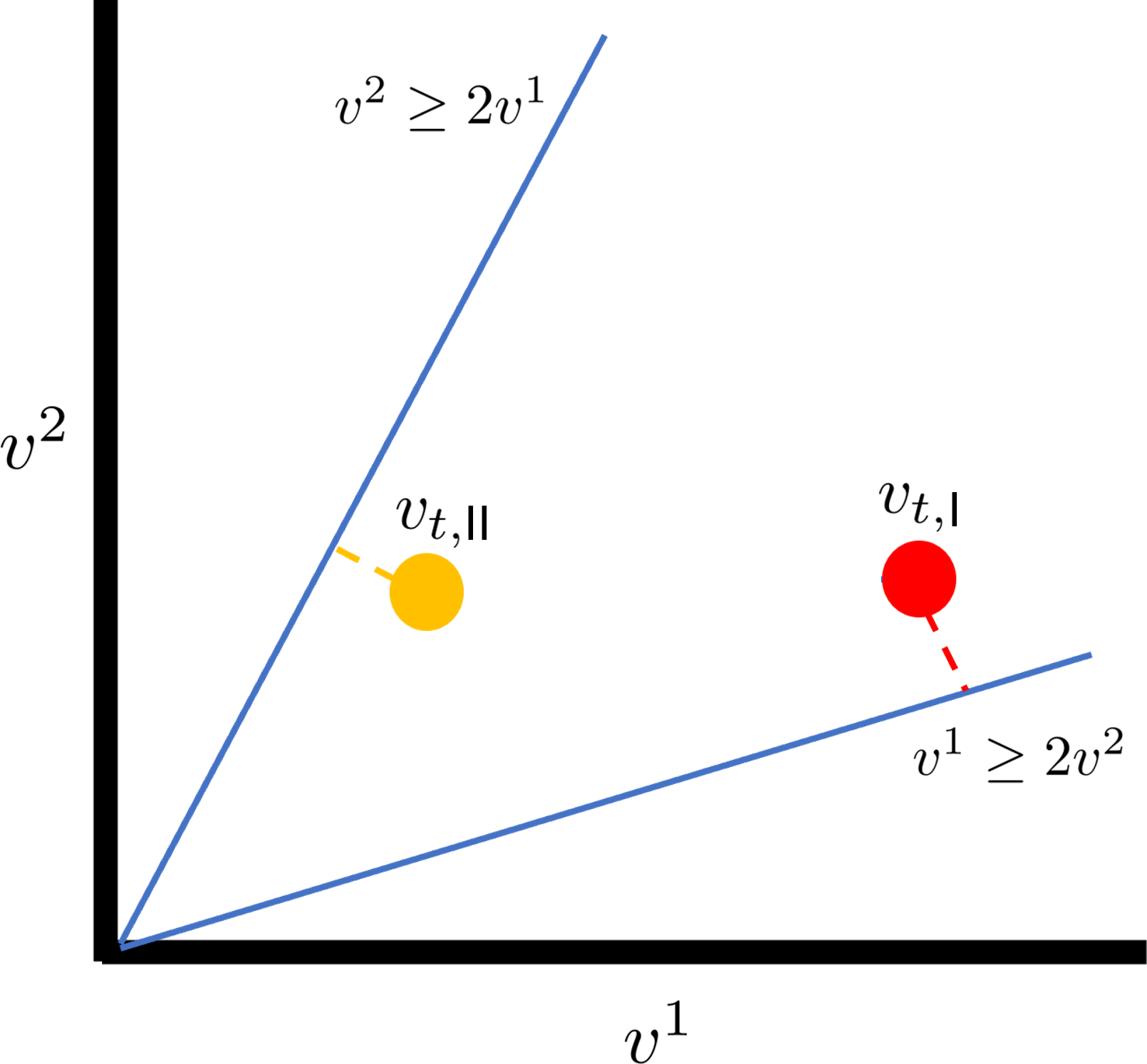}
    \caption{\textbf{Liveness set $\mathscr{C}_\ell(t)$ for the $2$ robot scenario:} Velocities for two deadlocked agents are shown by the red point $v_{t,I}$. To resolve the deadlock, Equation~\eqref{eq: perturbation} is used to project $v_{t,I}$ onto $v^1_t = 2 v^2$. Robot $1$ increases its velocity component $v^1$, while robot $2$ decreases its velocity component $v^2$ to align with the barrier. If robot $1$ deviates from $v_{t,I}$ and decreases its speed to the new yellow point $v_{t,II}$, the new optimal perturbation will be onto $v^2_t = 2 v^1$. Thus, robot $2$ adjusts its strategy by increasing its speed to align with a new perturbation. Assuming no speed deviations, there will be a unique projection to one of the safety barriers.}
    \label{fig: safety_certificates}
\end{figure}

\begin{proof}
The control strategy for carrying out the perturbation in Equation~\eqref{eq: perturbation} to modify the speed of a robot does not affect the rotation of the robot ($\Delta \theta^i_t = 0$), therefore ensuring that the robot tracks the current trajectory. This proves condition $1$ of Definition~\ref{def: min_invasive}.

To prove that the deadlock-preventing control, say $\Delta v_t$, is indeed $\delta_{opt}$, as defined by the second optimality criteria condition in Eq. 12 of Definition 4, we need to show that $\Delta v_t$ satisfies Equations 12a, 12b, 12c, and 12d.
% Let $v_{t+1} = \begin{bmatrix} v^1_t + \Delta v^1_t \\ v^2_t + \Delta v^2_t \end{bmatrix}$ where $\Delta v_{t+1} = \begin{bmatrix} \Delta v^1_t \\ \Delta v^2_t \end{bmatrix}$.}

\textbf{Equation 12a.}
We have to show that $\Delta v_t$ is the smallest change in $v_t$. This follows directly from the geometry of liveness sets in Figure\mbox{~\ref{fig: safety_certificates}}, where $\Delta v_t$ corresponds to the perpendicular projection from $v_t$ onto the boundary of the liveness set.

\textbf{Equation 12b.}
This is also straightforward from basic vector addition, considering the triangle formed with $v_t$ as the hypotenuse and $\Delta v_t$ as the base in Figure\mbox{~\ref{fig: safety_certificates}}, which gives $v_{t+1}$ as the perpendicular along the boundary of the liveness set.

\textbf{Equation 12c.}
As we consider single- and double-integrator dynamical systems with unconstrained control inputs, which are fully actuated systems, there always exist valid controls to achieve a desired configuration. This follows directly from properties of fully actuated systems (c.f. Chapter 1,\mbox{~\cite{underactuated}}).

\textbf{Equation 12d.}
The final constraint to satisfy states that the resulting state $x^i_{t+1}$ is deadlock-free, that is, $\left(x^i_{t+1},u^i_{t+1}\right)\notin\mathcal{D}^i(t+1)$. Suppose after applying $\Delta v_t$, $v_{t+1} \in \mathscr{C}_\ell(t+1)$. By Corollary~\mbox{\ref{cor: v_t}}, robots are deadlock-free since $v_{t+1} \in \mathscr{C}_\ell(t+1)$, $v^i_t \geq \zeta v^j_t$ (from Definition\mbox{~\ref{def: liveness_sets}}). According to Theorem\mbox{~\ref{thm: factor_of_2}}, we select $\zeta \geq 2$. Then, according to Theorem\mbox{~\ref{thm: smg->l<tau}}, $\zeta \geq 2 \implies \ell_j = \ell_i \geq \frac{\pi}{4} - \tan^{-1}\frac{1}{2}$. Finally, by Lemma\mbox{~\ref{lem: smg->deadlock}}, $\ell_j \geq \frac{\pi}{4} - \tan^{-1}\frac{1}{2}$ implies that robot $i$ is not in a deadlock with robot $j$, \textit{i.e.,} $\left \{  
x^i_{t+1}, u^i_{t+1} \not\in \mathcal{D}^i(t+1)\right \}$.

% Furthermore, using Corollary~\ref{cor: v_t}, the second condition in Definition~\ref{def: min_invasive} is also satisfied. 
\end{proof}

% Observe that a solution to~\eqref{eq: perturbation} achieves the first and third objective, that is, agents do not deviate from their preferred path and the new joint velocity $\widetilde v_t$ is such that agents deviate minimally from their current joint velocity $v_t$.
% We conjecture that scaling the maximum speeds of robots by some $\zeta$ fraction of the maximum speed of the lead robot is a sufficient condition to resolve a deadlock. In the case of $2$ robots, either $v^1$ or $v^2$ will be scaled by the other. 

\noindent\textit{Example $1$ (two agents):} While our approach generalizes to $k$ robots, we consider an example with $2$ robots for simplicity. In the $2$ robot scenario where each robot is equidistant from a doorway or intersection, the liveness set $\mathscr{C}^\ell_t$, shown in Figure~\ref{fig: safety_certificates}, is generated by scaling $v^2_t$ by $\frac{v^1_t}{\zeta}$, or vice-versa. We can then generate the following system of linear inequalities,
\begin{align*}
    v^1_t &\geq \zeta v^2_t \\
    v^2_t &\geq \zeta v^1_t
\end{align*}
This can be compactly represented as
\begin{equation}
    A_{2\times 2} v_t \geq 0
    \label{eq: av_geq_0}
\end{equation}
where $A_{2\times 2} = \begin{bmatrix}
1 & -\zeta \\
-\zeta & 1
\end{bmatrix}$ and $v_t= \left[v^1_t, v^2_t\right]^\top$.
Suppose the current value of $v_t$ is $p_1$ as shown in Figure~\ref{fig: safety_certificates}. The point $p_1$ indicates that $A_{2\times 2} p_1 < 0$ which lies outside $\mathscr{C}_\ell(t)$, implying that the two robots are in a deadlock according to Definition~\ref{def: liveness_sets} and Definition~\ref{def: deadlock}. Equation~\eqref{eq: perturbation} projects $p_1$ onto the nearest half-plane which is the $v^1= \zeta v^2$ barrier. Thus, robot $1$ will increase $v^1$ and robot $2$ will decrease $v^2$ by projecting $p_1$ down on $v^1= \zeta v^2$. This projection is the minimal deviation required on the part of both robots.  

% \textit{Remark:} \textbf{Invariance of $\mathscr{C}_\ell(t)$}. The set $\mathscr{C}_\ell(t)$ remains invariant in non-symmetrical deadlock configurations as well, since the liveness condition depends on relative position and relative velocity of the robots (See Figure~\ref{fig: geom3}).
% \input{algorithms/overall}

% We claim that the controller in Equation~\eqref{eq: SNUPI_safety_constrained_velocity_scaling} is a \textit{sufficient} controller for solving Problem~\ref{prob: optimal_social_nav}. That is, if a solution to Equation~\eqref{eq: SNUPI_safety_constrained_velocity_scaling} exists, then the robot will reach the goal without colliding. We contrast with the state-of-the-art safety-constrained controller~\cite{wang2017safety} where the safety barrier certificate is based on a safety distance threshold. Although this controller is a sufficient controller to avoid collisions, it is not a sufficient condition for avoiding deadlocks.

\noindent\textit{Example $2$ (three agents):} We can extend this to $3$ robots with speeds $v^1_t, v^2_t, v^3_t$. Assuming W.L.O.G that $v^{1} > v^{2} > v^{3}$, We found empirically that scaling $v^2$ by $\frac{v^1}{2}$, and $v^3$ by $\frac{v^1}{3}$ generates the $\mathscr{C}_\ell(t)$ for $3$ agents attempting to pass through a doorway from an equidistant. We can generate the following system of linear inequalities,
\begin{equation}
\begin{split}
    v^1_t&\geq 2v^2_t\\
    v^1_t&\geq 3v^3\\
    3v^2_t&\geq 2v^3\\
\end{split}    
\label{eq: av_3}
\end{equation}
This can be compactly represented as $A_{3\times 3} v_t \geq 0$ where $A_{3\times 3} = \begin{bmatrix}
1 & -2 & 0\\
1 & 0 & -3\\
0 & 3 & -2\\
\end{bmatrix}$ and $ v_t= \left[v^1_t, v^2_t, v^3_t\right]^\top$. Note that there will be $6$ possible permutations of $A$, resulting in $6$ tetrahedrons within a cube. Given a point $p = \left[v^1, v^2, v^3\right]^\top$ floating in this cube that does not lie within any of the $6$ tetrahedrons, then the optimal perturbation strategy is to project $p$ onto the face of the nearest tetrahedron.

\begin{subequations}
\begin{align}
\left( \Gamma^{i,*}, \Psi^{i,*}\right) =& \arg\min_{(\Gamma^i,  \Psi^{i})} \sum_{t={0}}^{T-1} \mathcal{J}^i\left(x^i_t, u^i_t\right) + \mathcal{J}^i_T\left(x^i_T \right) \\
\text{s.t}\;\;  x^i_{t+1}=& f\left(x^i_t\right) + g\left(x^i_t\right)u^i_t,\quad\forall t\in[1;T-1]\label{eq: 14b}\\
% \mathcal{C}^i\left( x^i_t \right) &\cap \mathcal{C}^j\left( x^j_t \right)= \emptyset \ \forall j \in \mathcal{N}\left(x^i_t \right) \ \forall t \label{eq: sub_coll}\\
&\Delta h^i\left(x^i_t, u^i_t\right) \geq -\gamma h^i\left(x^i_t\right)\label{eq: 14c}\\
% &h_v\left(x_t\right) > 0 \label{eq: 14d}\\
% &x^i_0\in \mathcal{X}_i\\ 
&x^i_T\in \mathcal{X}_g
\end{align}
\label{eq: SNUPI_safety_constrained_velocity_scaling_v}
\end{subequations}
% where $Q\succ 0$ and $R\succ 0$ are positive definite matrices and
% \begin{equation}
% \begin{split}
% \pi^{i,*} = \arg\min_{\pi^i}  &= \sum_{t={1}}^T \left\lVert 
% x^i_t - \widetilde x^i_t  \right\rVert^2\\
% \textrm{s.t} \quad x^i_t&= \mathcal{T}\left( x^i_{t-1},  \pi^i\left(o^i_{t-1}\right), u^{-i}\right)\\
%  \pi^i\left(\cdot\right) &\in \mathscr{U}^i_t = \left\{u^i_t \in U |       B\left(x_t^i\right)u_t^i\geq C\left(x_t^i\right)\right \} \\
% \end{split}    
% \label{eq: SNUPI_safety_constrained_velocity_scaling_v}
% \end{equation}
where constraints~\eqref{eq: 14b} and \eqref{eq: 14c} must be enforced at all times. Furthermore,
\begin{subequations}
\begin{align}
  & h^i\left(x_t^i\right)=\left[h^i_s\left(x_t^{i,1}\right),\ldots, h^i_s\left(x_t^{i,k-1}\right),  h_v\left(x_t\right)\right]^\top,\label{eqn:overall_h}\\
  % & h^i\left(x_t^i\right)=\left[h^i_s\left(x_t^{i,1}\right),\ldots, h^i_s\left(x_t^{i,k-1}\right),\;h_v\left(x_t^i\right)\right]^\top,\label{eqn:overall_h}\\
  & h^i_s\left(x^{i,j}_t\right)=\left\lVert p_t^i-p_t^j\right\rVert^2_2- r^2,\forall\quad j\in[1;k]\setminus i\label{eqn:CA_h}\\
  & h_v\left(x_t\right)=   \bar A_{k\times 4k} \left(x_t\right) \label{eqn:h_v}
    % & h_v\left(x_t\right)=A_{2\times 2}\left( Q u_t + \kappa \widehat x_t\right )\label{eqn:h_v}
  % & h_v\left(x_t\right)= \left( \bar A_{k\times k}Qu_t - \frac{\partial Q}{\partial x}u_t\bar A_{k\times k}x_t \right) + \kappa \bar A_{k\times k}x_t
  % \bar A_{k\times k}\left( Qu + \kappa x_t\right)
\end{align}    
  \label{eq: constraints}
\end{subequations}
% $B\left(x_t^i\right)$ and $C\left(x_t^i\right)$ are given by
% \begin{subequations}
% \begin{align}
%   & B\left(x_t^i\right)=\frac{\partial h\left(x_t^i\right)}{\partial x^t_i}g\left(x_t^i\right),\\
%   & C\left(x_t^i\right)=-\frac{\partial h\left(x_t^i\right)}{\partial x^t_i}f\left(x_t^i\right)-\kappa  h\left(x_t^i\right),\\
%   & h\left(x_t^i\right)=\left[h^i_s(x_t^{i,1}),\ldots, h^i_s(x_t^{i,N-1}),\;h_v\left(x_t^i\right)\right]^\top,\label{eqn:overall_h}\\
%   & h^i_s\left(x^{i,j}_t\right)=\left\lVert p_t^i-p_t^j\right\rVert^2_2- r^2,\forall\quad j\in[1;N]\setminus i\label{eqn:CA_h}\\
%   & h_v\left(x_t\right)=   \bar A_{k\times k} \left(x_t\right)\label{eqn:h_v}
%     % & h_v\left(x_t\right)=A_{2\times 2}\left( Q u_t + \kappa \widehat x_t\right )\label{eqn:h_v}
%   % & h_v\left(x_t\right)= \left( \bar A_{k\times k}Qu_t - \frac{\partial Q}{\partial x}u_t\bar A_{k\times k}x_t \right) + \kappa \bar A_{k\times k}x_t
%   % \bar A_{k\times k}\left( Qu + \kappa x_t\right)
% \end{align}    
%   \label{eq: constraints}
% \end{subequations}
where $h^i_s(x_t^{i,j})$ represents the CBF for the agent $i$ which ensures that agent $i$ does not collide with agent $j$ by maintaining a safety margin distance of at least $r$.

% The input constraints in system in (\mbox{\ref{eq: SNUPI_safety_constrained_velocity_scaling_v}}) do not affect feasibility as these constraints are implicitly and preemptively accounted for when determining the original action set $\mathcal{U}^i$ for any agent $i$.

The program\mbox{~\eqref{eq: SNUPI_safety_constrained_velocity_scaling_v}} may not be feasible for certain input-constrained systems, which we have discussed in Section\mbox{~\ref{subsec: input_constraints}}. For unconstrained systems, feasibility for the program in\mbox{~\eqref{eq: SNUPI_safety_constrained_velocity_scaling_v}} can be discussed using the definition of Control Barrier Functions (CBFs) in Equation\mbox{~\eqref{eq: discrete_barrier_func}}. In other words, Theorem 3 in\mbox{~\cite{ames2019control}} states that if there exists a valid CBF $h^i\left(x^i_t\right)$ that satisfies\mbox{~\eqref{eq: C_set}} and a $\mathcal{K}_\infty$ function $\kappa$, then\mbox{~\eqref{eq: discrete_barrier_func}} holds true. If\mbox{~\eqref{eq: discrete_barrier_func}} holds true, then that implies that there must exist a control input $u^i_t$ belonging to the compact set $U^i$ that satisfies\mbox{~\eqref{eq: discrete_barrier_func}}.

In this approach, we can construct a valid CBF apriori (typical of the kind of problem formulation under investigation\mbox{~\cite{wang2017safety}}), and following\mbox{~\eqref{eq: discrete_barrier_func}}, there always exists a control input $u^i_t$ that satisfies condition (20c). In general, however, we acknowledge that computing a valid CBF remains a challenging problem. While previous research has employed neural networks\mbox{~\cite{robey2020learning}} and Gaussian Processes\mbox{~\cite{jagtap2020control_gp}} to learn CBFs, guaranteeing the validity of learned or analytically derived functions as CBFs is still an open research question.

% Lastly, while the goal constrained condition (20d) can potentially render~\eqref{eq: discrete_barrier_func} infeasible, it is always possible to ensure the feasibility of the MPC by modifying the problem formulation. One effective approach is to add the goal constraint as a soft constraint in the objective function, which allows for a trade-off between achieving the goal and maintaining feasibility.

Further, $h_v\left( x_t\right) = \bar A_{k\times k} \left(x_t\right)$  is the expanded form of Equations~\ref{eq: av_geq_0} and~\ref{eq: av_3} where $x_t = \left[p^1_t,v^1_t,\theta^1_t, \omega^1_t,p^2_t, v^2_t,\theta^2_t, \omega^2_t  \right]^\top$ and controls, $u_t = \left[u^1_t,u^2_t\right]^\top$. Following the $2$ agent example, we can expand the matrix $A_{2 \times 2}$ as $\bar A_{2 \times 8} = \begin{bmatrix}
0 & 1 & 0 & 0 & 0& -\zeta&0&0\\
0 &  -\zeta &0 & 0 & 0 & 1&0&0\\
\end{bmatrix}$ to accommodate the aggregate of both the robots' states and controls. The weights or entries in the A matrix control the sensitivity to deadlocks. Specifically, tuning the matrix towards an identity matrix ($\zeta \longrightarrow 1$) decreases the sensitivity--the robots do not perturb their velocities and the robots would either end up in a deadlock or collide (if they were moving fast enough). On the other hand, increasing the relative weights ($\zeta >> 0$) makes the robots increasingly sensitive to deadlocks--that is, the robots would tend to react to false positive scenarios (scenarios that might be tending towards a deadlock, but are not strictly so).
% The inverse, however, may not hold true. That is, $v_t \notin \mathscr{C}_\ell(t)$ may not necessarily result in a social mini-game. For example, two agents equidistant from a doorway with equal speeds may prefer to travel parallel to each other. Therefore, perturbing agents' speeds via Equation~\eqref{eq: perturbation} only if $v_t \notin\mathscr{C}_\ell(t)$ may be an overly conservative approach. 
% In contrast to traditional CBFs where the CBF $h\left(x^i_t\right)$ is usually a function of spatial coordinates such as position only, our proposed CBF $h\left(x^i_t\right)$ in \eqref{eqn:overall_h} is a function of both position $h^i_s(x^{i,j}_t\right)$ (defined in \eqref{eqn:CA_h}) and velocity $h_v(x_t\right)$ (defined in \eqref{eqn:h_v}). Unifying these constraints in the formulation of CBF enables robots to simultaneously prevent collisions and deadlocks. Furthermore, f

For cases of single integrator dynamics where $v^i_t$ is a input, the constraint $Av_t \geq 0$ (where $v_t=[v^1_t,\;v^2_t]^\mathrm{T}$ for two agents) cannot be directly incorporated as a CBF, instead~\eqref{eqn:h_v} is invoked after the MPC selects an optimal control. However in cases of double or higher-order robotic systems (such as bipedal robots \cite{grizzle2014models_bipedal_robots}, Boston Dynamics Spot etc.), since $v^i_t$ is also a state, guaranteeing the invariance of the set $A_{k \times k}v_t\geq 0$ becomes trivial by modeling~\eqref{eqn:h_v} as a CBF constraint.

\textit{Remark:} Despite the inherent non-convexity of the MPC-CBF program (\mbox{\ref{eq: SNUPI_safety_constrained_velocity_scaling_v}}) due to the collision constraint (\mbox{\ref{eqn:CA_h}}), our decentralized strategy converges to a local solution within each planning window. We elevate a conventional local solution by endowing it with critical safety and liveness attributes through our CBF constraint (\mbox{\ref{eqn:h_v}}), thereby rendering the solution not only viable but preferable. This strategy represents a substantial improvement over methods that necessitate convergence to a static global solution--a process often marked by inefficiency and impracticality due to the complexities associated with computing global solutions.

\textit{Remark:} For $v^i_t \neq v^j_t$ for all $i,j$, there is always a unique perturbation. Consider in Example $1$ that robot $1$ decides to deviate from its current speed in $v_{t,1}$ and decides to decrease its speed, shown by the new point $v_{t,2}$. In that case, robot $2$'s optimal strategy will no longer be to decrease its speed as before. Now, the nearest safety barrier becomes $v^2_t= \zeta v^1$, and to project $v_{t,2}$ to this barrier, robot $2$ will instead increase its speed. Therefore, assuming a robot does not deviate from its current speed, there will be a unique projection to one of the safety barriers.

\subsection{Tie-breaking via Priority Orderings} 
\label{subsec: auctions}
If $v^i_t = v^j_t$ for some $i,j$, then there are multiple solutions and we implement the following tie breaking protocol. In social mini-games, we define a conflict zone as a region $\varphi$ in the global map that overlaps goals corresponding to multiple robots. A conflict, then, is defined by the tuple $\langle \mathcal{C}^t_{\varphi}, \varphi, t \rangle$, which denotes a conflict between robots belonging to the set $\mathcal{C}^t_{\varphi}$ at time $t$ in the conflict zone $\varphi$ in $\mathcal{G}$. Naturally, robots must either find an alternate non-conflicting path or must move through $\varphi$ according to a schedule informed by a priority order. A priority ordering is defined as follows,

\begin{definition}
\textbf{Priority Orderings ($\sigma$): } A priority ordering is a permutation $\sigma:\mathcal{C}^t_{\varphi} \rightarrow [1,k]$ over the set $\mathcal{C}^t_{\varphi}$. For any $i,j \in [1,k]$, $\sigma^i = j$ indicates that the $i^\textrm{th}$ robot will move on the $j^\textrm{th}$ turn with $\sigma^{-1}(j) = i$.
\label{def: turn_based_ordering}
\end{definition}

\noindent For a given conflict $\langle \mathcal{C}^t_{\varphi}, \varphi, t \rangle$, there are $\left\lvert \mathcal{C}^t_{\varphi} \right\rvert$ factorial different permutations.
% In social mini-games occurring in the real world, however, certain permutations are more optimal than others. From the human factors perspective, body language, gaits, and gaze play a major role in deciding the priority order. For instance, conservative individuals would be more likely to yield to assertive individuals. In traffic, social protocol dictates that emergency vehicles must be given the right of way.
There exists, however, an \textit{optimal} priority ordering, $\sigma_\textsc{opt}$.

\begin{definition}
\textbf{Optimal Priority Ordering ($\sigma_{\textsc{opt}}$): } A priority ordering, $\sigma$, over a given set $\mathcal{C}^t_{\varphi}$ is optimal if bidding $b^i = \zeta^i$ is a dominant strategy and maximizes $\sum_{\lvert \mathcal{C}^t_{\varphi} \rvert} \zeta^i \alpha^i$, where $\zeta^i$ is a private incentive or priority parameter known only to agent $i$, and $\alpha_q$ is a time-based reward for receiving an order position $\sigma^i = q$.
\label{def: optimal_turn_based_ordering}
\end{definition}

We run an auction, $(r^i, p^i)$, with an allocation rule $r^i(b^i) = \sigma^i = q$ and payment rule $p^i$ defined by $p^i(b^i) =  \sum_{j=q}^{\lvert \mathcal{C}^t_{\varphi} \rvert} \widehat b^{\sigma^{-1}(j+1)} \left( \alpha_j - \alpha_{j+1} \right)$. The payment rule is the ``social cost'' of reaching the goal ahead of the robots arriving on turns $q+1, q+2, \ldots, q+\mathcal{C}^t_{\varphi}$. The bids, $\widehat b^{\sigma^{-1}(q+1)}, \widehat b^{\sigma^{-1}(q+2)}, \ldots, \widehat b^{\sigma^{-1}\left(q+\lvert \mathcal{C}^t_{\varphi} \rvert\right)}$ represent proxy robot bids sampled from a uniform distribution, since robots do not have access to the bids of other robots. Using $(r^i, p^i)$ defined as above, each robot solves

\begin{equation}
b^{i,*} = \arg\min_{b^i} \left ( \alpha_{r^i(b^i)} - p^i(b^i) \right )
\label{eq: utility_social}
\end{equation}

It is known~\cite{roughgarden2016twenty, chandra2022gameplan} that the auction defined by $(r^i, p^i)$ yields $b^{i,*} = \zeta^i$ and maximizes $\sum_{\lvert \mathcal{C}^t_{\varphi} \rvert} \zeta^i \alpha^i$. To summarize the algorithm, the robot with the highest bid, that is, the highest incentive $\zeta^i$, is allocated the highest priority and is allowed to move first, followed by the second-highest bid, and so on. 

\textit{Remark:} The tie breaking protocol via auctions is not decentralized as there is an assumption of an auction program that receives bids and distributes the ordering among the agents. Since control is still decentralized, we consider this a distributed optimization problem instead.

\subsection{Application in Human Environments: Relaxing the Perturbation}

% \begin{figure}[t]
%     \centering
%     \begin{subfigure}[h]{0.445\columnwidth}
% \includegraphics[width = \linewidth]{images/robot_in_human.png}    
%     \caption{Velocity perturbation for robot in human environments. The point $P$ corresponds to the current joint velocity of the robot and the human. The length $\overrightarrow{PB}$ corresponds to the velocity difference, $\Delta \widetilde v^\textrm{robot}_t$, and $\overrightarrow{PA}$ corresponds to the solution obtained via Equation~\eqref{eq: perturbation} given by $\min_{\mu \in \mathscr{C}_\ell(t)} \left \lVert v_t - \mu\right \rVert_2$.}
%     \label{fig: app_in_human_env}
%     \end{subfigure}
%     %
%     \begin{subfigure}[h]{0.445\columnwidth}
% \includegraphics[width = \linewidth]{images/rcbf.png}    
%     \caption{Visualizing the concept of robust control barrier functions (RCBFs) used in single-integrator dynamics to ensure safety despite perturbation via Equation~\eqref{eq: perturbation}. The blue area represents the safety region $\mathcal{C}^i$ with no input uncertainty ($\rho = 0$). When input uncertainty $\rho \leq 0$ is applied, this safe region is bounded by a red dashed line to expand the safe set to $\mathcal{\widetilde C}^i$.}
%     \label{fig: rcbf}
%     \end{subfigure}
%     \caption{\textit{(left)} Interaction dynamics between a robot and a human \textit{(right)} Adaptation of the safety region in control systems with RCBFs to account for input uncertainties.}
%     \label{fig: analysis_figs}
%     % \vspace{-10pt}
% \end{figure}

\begin{figure}[t]
    \centering
\includegraphics[width = \columnwidth]{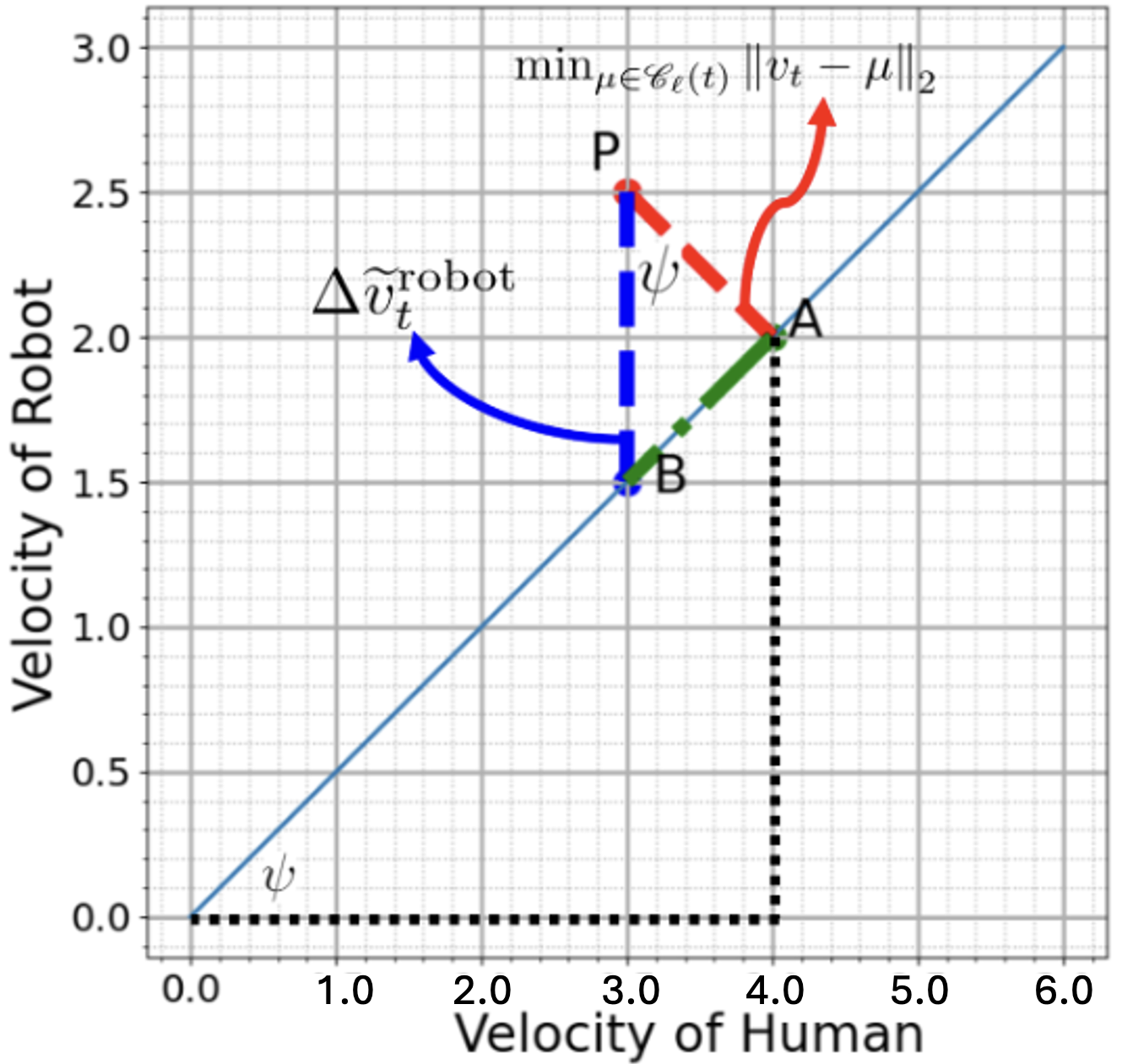}    
    \caption{Velocity perturbation for robot in human environments corresponding to the case when the robot slows down in response to the human (\textcolor{blue}{\textbf{blue line}}). The point $P$ corresponds to the current joint velocity of the robot and the human. The length $\overrightarrow{PB}$ corresponds to the velocity difference, $\Delta \widetilde v^\textrm{robot}_t$, and $\overrightarrow{PA}$ corresponds to the solution obtained via Equation~\eqref{eq: perturbation} given by $\min_{\mu \in \mathscr{C}_\ell(t)} \left \lVert v_t - \mu\right \rVert_2$.}
    \label{fig: app_in_human_env}
    \end{figure}

Humans are unpredictable and often make irrational decisions; it is impractical to assume that humans will precisely execute the perturbation strategy in Equation~\eqref{eq: perturbation}. By shifting the responsibility for ensuring $v_t \ \in \mathscr{C}_\ell(t)$ entirely to the robot, we can still guarantee deadlock prevention in human environments, albeit at the cost of trading away minimal invasiveness. More specifically, assume that the human does not perturb their velocity. Then, the robot faces two options.

% \begin{itemize}
    % \item 
If $v^{\textrm{robot}} < v^{\textrm{human}}$, then the robot will decrease its speed such that $\zeta v^{\textrm{robot}} \leq v^{\textrm{human}}$. Formally 
    
    \begin{equation}
        \Delta \widetilde v^\textrm{robot}_t = \frac{\min_{\mu \in \mathscr{C}_\ell(t)} \left \lVert v_t - \mu\right \rVert_2}{\cos\psi}
        \label{eq: robot_slowing}
    \end{equation} 
    
    where $\psi = \tan^{-1}\zeta $. We visualize this derivation geometrically in Figure\mbox{~\ref{fig: app_in_human_env}}. $\psi$ represents the slope of the boundary lines of the liveness set, which could be either $\zeta$ or $\frac{1}{\zeta}$, which is shown in Figure\mbox{~\ref{fig: app_in_human_env}} as the angle subtended by the black dotted lines. Since the liveness set is symmetric, we can select either slope without loss of generality. Thus, $\psi = \tan^{-1}(\zeta)$. When the robot slows down in response to the human (vertical \mbox{\textcolor{blue}{\textbf{blue}}} line), as shown in Figure\mbox{~\ref{fig: app_in_human_env}}, it is trivial to see that the angle between the \mbox{\textcolor{blue}{\textbf{blue}}} and the \mbox{\textcolor{red}{\textbf{red}}} lines is also $\psi$. From trigonometry, we obtain Equation\mbox{~\eqref{eq: robot_slowing}}.
    
    % \item 
    If $v^{\textrm{robot}} > v^{\textrm{human}}$, then the robot will increase its speed such that $ v^{\textrm{robot}}_{limit} \geq v^{\textrm{robot}} \geq \zeta  v^{\textrm{human}}$. Formally,
    
{\small    \begin{equation}
        \Delta \widetilde v^\textrm{robot}_t = \min \left\{ \frac{\min_{\mu \in \mathscr{C}_\ell(t)} \left \lVert v_t - \mu\right \rVert_2}{\sin\psi}, v^{\textrm{robot}}_{limit} -v^{\textrm{robot}} \right\}
        \label{eq: human slows}
    \end{equation}}
    where $\psi = \tan^{-1}\zeta$. We can similarly derive Equation\mbox{~\eqref{eq: human slows}} which represents the alternative scenario when the robot speeds up in response to the human. Here the difference is that the \mbox{\textcolor{blue}{\textbf{blue}}} line will now be a \textit{horizontal} line from P, intersecting the boundary of the liveness set hyperplane. Using basic euclidean geometry, it is easy to see that the $\cos\psi$ in the formula will change to $\sin\psi$. We apply the minimum operator because we have to take into consideration the physical upper limits of the robots speed.

% \end{itemize}

It is easy to observe that this is a relaxation since the perturbation is larger for the robot in this case. Let $\nu = \min_{\mu \in \mathscr{C}_\ell(t)} \left \lVert v_t - \mu\right \rVert_2$. Then,

\[\frac{\min_{\mu \in \mathscr{C}_\ell(t)} \left \lVert v_t - \mu\right \rVert_2}{\cos\psi} \geq \nu \]

as well as

\[\min \left\{ \frac{\min_{\mu \in \mathscr{C}_\ell(t)} \left \lVert v_t - \mu\right \rVert_2}{\sin\psi}, v^{\textrm{robot}}_{limit} -v^{\textrm{robot}} \right\} \geq \nu .\]

Here, the relaxation refers to the trading away of minimal invasiveness by shifting of the responsibility for ensuring $v_t \ \in \mathscr{C}_\ell(t)$ entirely to the robot.

\subsection{Safety Analysis for Perturbation via Equation~\eqref{eq: perturbation}}

Finally, we discuss the safety of the control strategy necessary to carry out the perturbation in Equation~\eqref{eq: perturbation}. In double-integrator (or higher order) dynamical systems, the constraints~\eqref{eq: av_geq_0} and~\eqref{eq: av_3} can be incorporated as CBFs. Formally, we generate the liveness set $\mathscr{C}_\ell(t)$ as,
\begin{equation}
\begin{split}
    \mathscr{C}_\ell(t) &= \left\{  v_t  \ \textnormal{s.t.} \ h_v\left(x_t\right) \geq 0 \right\} \\
    % \mathcal{V} &=  \left [ 0, \widehat v^1\right ] \times \ldots \times \left [ 0, \widehat v^k \right ]\\
    h_v( x_t) &= \bar A_{k\times k} \left(x_t \right)
\end{split}
\label{eq: game_safety_constraint}
\end{equation}

where $h_v\left( x_t\right)$ is the CBF associated with the safe set $\mathscr{C}_\ell(t)$ that can be used to ensure the forward invariance of $\mathscr{C}_\ell(t)$, $x_t = \left[p^1_t,v^1_t,\theta^1_t, \omega^1_t,p^2_t, v^2_t,\theta^2_t, \omega^2_t  \right]^\top$ and controls, $u_t = \left[u^1_t,u^2_t\right]^\top$. Following the $2$ agent example, we can expand the matrix $A_{2 \times 2}$ as $\bar A_{2 \times 8} = \begin{bmatrix}
0 & 1 & 0 & 0 & 0& -\zeta&0&0\\
0 &  -\zeta &0 & 0 & 0 & 1&0&0\\
\end{bmatrix}$ to accommodate the aggregate of both the robots' states and controls. The CBF $h_v(x_t)$ is then integrated as constraint in the optimization~\eqref{eq: SNUPI_safety_constrained_velocity_scaling_v} as follows,

    \begin{figure}
    \centering
\includegraphics[width = \columnwidth]{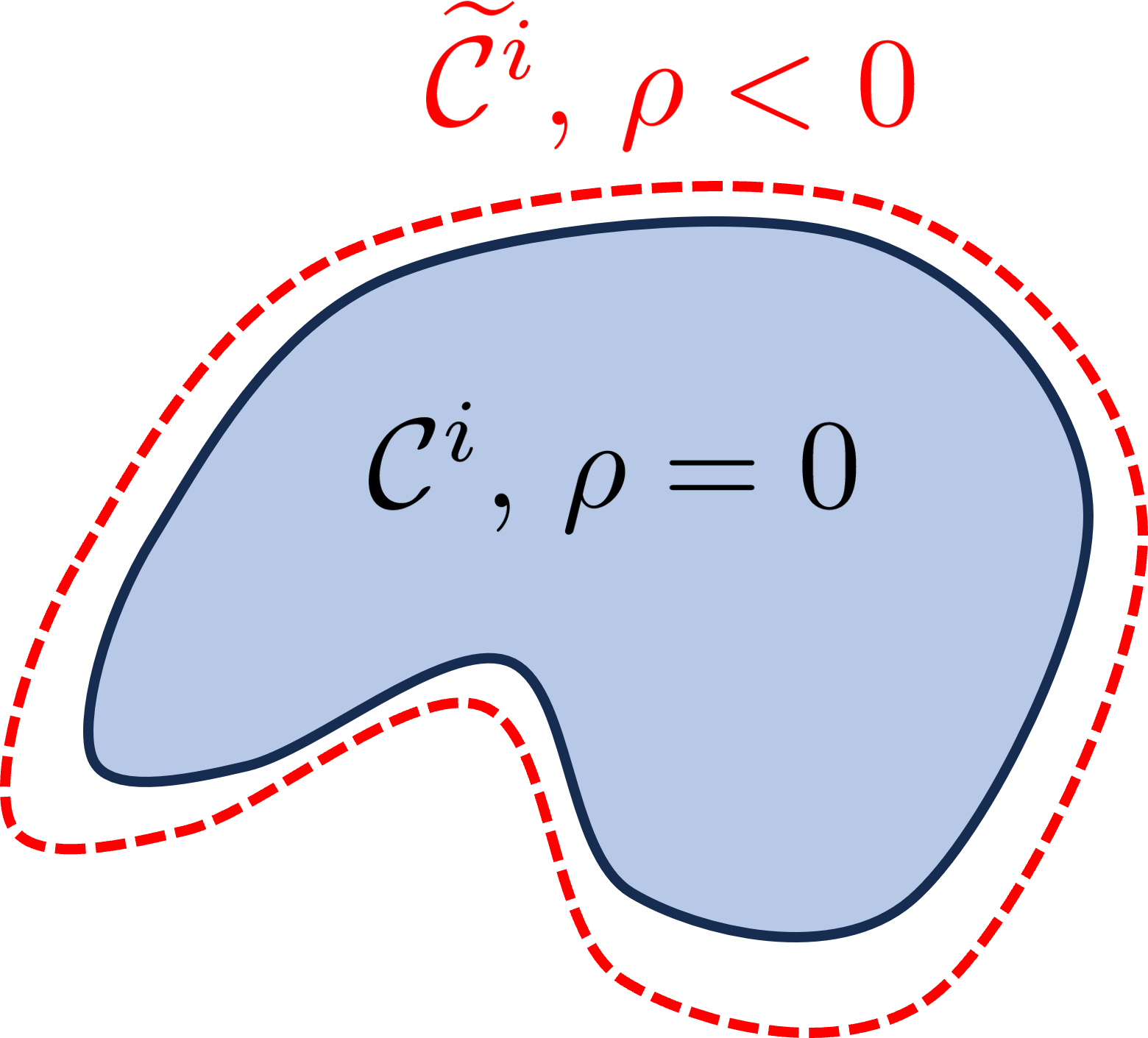}    
    \caption{Visualizing the concept of robust control barrier functions (RCBFs) used in single-integrator dynamics to ensure safety despite perturbation via Equation~\eqref{eq: perturbation}. The blue area represents the safety region $\mathcal{C}^i$ with no input uncertainty ($\rho = 0$). When input uncertainty $\rho \leq 0$ is applied, this safe region is bounded by a red dashed line to expand the safe set to $\mathcal{\widetilde C}^i$.}
    \label{fig: rcbf}
\end{figure}

\begin{equation}
   \Delta h^i\left(x^i_t, u^i_t\right) \geq -\gamma h^i\left(x^i_t\right)
  \label{eq: combined_cbf_vel_safety}
\end{equation}

\noindent where $h^i\left(x_t^i\right)=\left[h^i_s\left(x_t^{i,1}\right),\ldots, h^i_s\left(x_t^{i,k-1}\right),\;h_v\left(x_t\right)\right]^\top
$. This guarantees that the control strategy necessary to ensure that $v_t \ \in \mathscr{C}_\ell(t)$ does not escape the safe set. 

However, for single-integrator dynamics, we can use robust CBFs to guarantee safety for the perturbation control strategy. Assume that $\delta u^i_t$ represents the perturbation necessary to achieve the velocity governed by Equation~\ref{eq: perturbation}. The system dynamics with uncertainty in the input is described by:
\[
x^i_{t+1} = f\left(x^i_t\right) + g\left(x^i_t\right)\left(u^i_t + \delta u^i_t\right),
\]
where \(\delta u^i_t\) represents the input uncertainty bounded by \(\left\lvert\delta u^i_t\right\rvert \leq \lvert p \rvert\) for a known bound \(p\). Then, 
\begin{align}
   &\Delta h^i\left(x^i_t\right)\nonumber\\
   &=h\left(x_t^{i+1}\right)-h\left(x^i_t\right)\nonumber\\
   &=h\left(\left(x^i_t\right) + g\left(x^i_t\right)\left(u^i_t + \delta u^i_t\right)\right)-h\left(x^i_t\right)\nonumber\\
   &=h\left(\left(x^i_t\right) + g\left(x^i_t\right)\left(u^i_t + \delta u^i_t\right)\right)-\nonumber\\
   &h\left(\left(x^i_t\right) + g\left(x^i_t\right)\left(u^i_t \right)\right)h\left(x^i_t\right)+\nonumber\\
   &h\left(\left(x^i_t\right) + g\left(x^i_t\right)\left(u^i_t \right)\right)h\left(x^i_t\right)-h\left(x_t^i\right)\nonumber\\
   &=\Delta_{\text{original}} h^i\left(x^i_t\right)+\Delta_{\text{modification}} h^i\left(x^i_t,u^i_t\right)
   % \underbrace{ L_f h^i\left(x^i_t\right) + L_g h^i\left(x^i_t\right)u^i_t}_{\textrm{original}} 
    % + \underbrace{L_g h^i\left(x^i_t\right)\delta u^i_t}_{\textrm{modification}},
\end{align}
where 
% \clearpage

\begin{align*}
\Delta_{\text{modification}}h^i\left(x^i_t,u^i_t\right) 
   &= \left( h\left( \left( x^i_t \right) + g\left( x^i_t \right)\left( u^i_t + \delta u^i_t \right) \right) \right. \nonumber\\
   &\quad \left. - h\left( \left( x^i_t \right) + g\left( x^i_t \right) \left( u^i_t \right) \right) \right), \\
   \Delta_{\text{original}} h^i\left(x^i_t\right)
   &= \left( h\left( \left( x^i_t \right) + g\left( x^i_t \right) \left( u^i_t \right) \right) \right. \nonumber\\
   &\quad \left. - h\left(x_t^i\right) \right).
   \end{align*}

% $\Delta_{\text{modification}} h^i\left(x^i_t,u^i_t\right)=h\left(x^i_t\right) + g\left(x^i_t\right)\left(u^i_t + \delta u^i_t\right))-h\left(x^i_t\right) + g\left(x^i_t\right)\left(u^i_t \right))h\left(x^i_t\right)$ and $\Delta_{\text{original}} h^i\left(x^i_t\right)=h\left(x^i_t\right) + g\left(x^i_t\right)\left(u^i_t \right))h\left(x^i_t\right)-h\left(x_t^i\right)$.
The above no longer satisfies the condition~\eqref{eq: safety_barrier_certificate}. In the literature, this problem is often addressed by adding a compensation term to the barrier function $h^i\left(x^i_t\right)$ such that the superlevel set $\mathcal{C}^i$ expands to accommodate the perturbation. The resulting barrier function belongs to the class of Robust CBFs or RCBFs~\cite{rcbfs}. An RCBF implies the following holds for all $t \geq 0$:
\begin{equation}
 \Delta h^i\left(x^i_t\right) - \rho\left(x^i_t\right)  \geq -\gamma h^i\left(x^i_t\right).
\end{equation}

for some $\rho: \mathcal{X} \times \mathcal{U} \longrightarrow \mathbb{R}$. We can derive an expression for \(\rho\) in terms of \(p\) to ensure the robustness of the control barrier function. Given \(\left\lvert\delta u^i_t\right\rvert \leq \lvert p \rvert\), the worst-case impact occurs when \(\delta u\) is aligned with \(g\left(x^i_t\right)\left(u^i_t\right)\) in the direction that most negatively affects \(h^i\left(x^i_t\right)\). Thus, \(\rho\) must compensate for the maximum possible deviation in \(\Delta_{\text{modification}} h^i\left(x^i_t,u^i_t\right)\) due to \(\delta u^i_t\). Mathematically, this can be expressed as:
\begin{align}
\rho\left(x^i_t\right) = \max_{\left\lvert \delta u^i_t\right\rvert \leq \lvert p \rvert} \left\lvert \Delta_{\text{modification}} h^i\left(x^i_t,u^i_t\right)\right\rvert.
\end{align}

Given that \(\delta u^i_t\) is bounded by \(\lvert p \rvert\), the term \(\left\lvert \Delta_{\text{modification}} h^i\left(x^i_t,u^i_t\right) \delta u^i_t\right\rvert\) can be maximized by taking \(\delta u^i_t\) at its bound. The linear approximation of this impact, assuming the worst-case alignment, leads to:
\[
\rho\left(x^i_t\right)  \leq \lvert \Delta_{\text{modification}} h^i\left(x^i_t,u^i_t\right)\rvert \lvert p \rvert.
\]

Here, \(\left\lvert \Delta_{\text{modification}} h^i\left(x^i_t,u^i_t\right)\right\rvert\) represents the norm (or a suitable measure of magnitude) of the Lie derivative of \(h\) along \(g\), indicating how changes in the input directly influence the safety condition through \(g(x)\). The term \(|p|\) represents the maximum magnitude of the uncertainty.

\subsection{Dealing with Input Constraints}
\label{subsec: input_constraints}

Integral Control Barrier Functions (ICBFs)\mbox{~\cite{zinage2024disturbance}} ensure safety under limited actuation for multi-agent systems by treating the control input as an auxiliary state. With an ICBF, the concept of a safety set extends the traditional state-based safety to the following definition: $\mathscr{S}:=\{ \left ( x_t , u_t \right ) \in\mathcal{C}\times\mathcal{U}\}$, where $\mathcal{C}$ represents a state-based safety set and input constraints in $\mathcal{U}$. We can handle input constraints by using ICBFs in place of the traditional CBF.

Now, consider a continuously differentiable scalar-valued function $h:\mathbb{R}^n\times\mathbb{R}^m\rightarrow\mathbb{R}$ defined such that $h(x_t, u_t)> 0$ if $(x_t, u_t)\in\mathscr{S}$, $h(x_t,u_t)=0$ if $(x_t, u_t)\in\partial\mathscr{S}$ and $h(x_t,u_t)<0$ if $(x_t, u_t)\in(\mathbb{R}^n\times\mathbb{R}^m)\setminus\mathscr{S}$. To guarantee that $(x_t,u_t)\in\mathscr{S}$, $h(x_t,u_t)$ must satisfy the forward invariance condition i.e. $\dot{h}(x_t,u_t)+\alpha(h(x_t,u_t)) \geq 0$ where $\dot{h}(x_t,u_t)$ is computed along the system trajectories and $\alpha$ is a $\mathcal{K}_\infty$ function. Formally, ICBF is defined as follows:
\begin{definition}
\textbf{(ICBF)} \mbox{\cite{ames2020integral_ames}} An Integral Control Barrier Function (ICBF) is defined as a function $h:\mathbb{R}^n\times\mathbb{R}^m\rightarrow\mathbb{R}$ that characterizes a safe set $\mathscr{S}$. Then, $h$ is said to be an ICBF, for all $(x_t, u_t)\in\mathscr{S}$ if $p(x_t, u_t)=0$, implies that

\begin{align*}
q(x_t, u_t) &\leq 0, \quad \text{where} \\
q(x_t, u_t) &:=-\left( \nabla_{x_t} h(x_t,u_t) f(x_t, u_t) \right. \\
    &\quad + \nabla_{u_t} h(x_t,u_t) \phi(x_t, u_t) \\
    &\quad + \left. \alpha(h(x_t, u_t)) \right), \\
p(x_t, u_t) &:=\left( \nabla_{u_t} h(x_t,u_t) \right)^\top.
\end{align*}

\label{defn:icbf}
\end{definition}
The following theorem provides a method to synthesize safe controllers via ICBFs

\begin{theorem}
    \mbox{\cite{ames2020integral_ames}} If an integral feedback controller $\phi(x_t,u_t)$ and a safety set $\mathscr{S}$, defined by an ICBF $h(x_t,u_t)$, exist, then modifying the integral controller to $\dot{u_t}=\phi(x_t, u_t)+v^\star(x_t, u_t)$,
where $v^\star(x_t, u_t)$ is obtained by solving the following quadratic program (QP):
\begin{align*}
&v^\star(x_t, u_t) = \underset{v \in \mathbb{R}^m}{\operatorname{argmin}}\|v\|^2,\;\;\text {s.t.}\;\; p(x_t, u_t)^{\mathrm{T}} v \geq q(x_t, u_t)
\end{align*}
guarantees safety, i.e., forward invariance of the set $\mathscr{S}$.
\end{theorem}

\subsection{Complete Algorithm}
% \begin{algorithm}[t]
%     \SetKwInOut{Input}{Input}
%     \SetKwInOut{Output}{Output}
% \SetKwComment{Comment}{$\triangleright$\ }{}
% \SetAlgoLined
% \Input{Map $m$, initial pose, and goals}
% \Output{$v, \omega$}
% A$^*$ path $\gets$ \textsc{Get Global Path}\\
% $v, \omega \gets$ \textsc{Local Planning}(A$^*$ path, $o^i_t, \Xi$)\\

% \caption{Overview of the navigation system}
% \label{algo: overall}
% % \vspace{-10pt}
% \end{algorithm}

\begin{algorithm}
\SetAlgoLined
\KwResult{Run MPC for trajectory tracking or setpoint control with obstacle avoidance}
 Initialize $\gets \mathcal{X}_I, \mathcal{X}_g$ \;
 Define simulation parameters: $T_{\text{sim}} = 3, T_s = 0.1, T_{\text{horizon}} = 3$\;
 Define cost matrices and actuator limits:  $f = \text{diag}([11, 11, 0.005])$, $g = [1, 0.5]$, $v_{\text{limit}} = 0.3$, $\omega_{\text{limit}} = 3.8$\;
 
 \SetKwFunction{FMain}{Model}
 \SetKwProg{Fn}{Function}{:}{}
 \Fn{\FMain{}}{
  Define state and control variables\;
  Define system dynamics using $f, g$ in~\eqref{eq: control_affine_dynamics}\;
  Define stage-wise cost $\mathcal{J}^i$\;
  \KwRet model\;
 }
 
 \SetKwFunction{FMpc}{MPC}
 \SetKwProg{Fn}{Function}{:}{}
 \Fn{\FMpc{}}{
  Call \FMain{}\;
  Configure parameters using $T_{\text{sim}}, T_s, T_{\text{horizon}}$ \;
  Set objective function via Equation~\eqref{eq: best_response_example}\;
  Define state and input constraints, $v_{\text{limit}}, \omega_{\text{limit}}$\;
  Add safety constraints via Equation~\eqref{eq: 14c}\;
  \If{double-integrator dynamics}{
    Add liveness constraint via~\eqref{eqn:h_v}\;
 }
  \SetKwFunction{FStep}{Step}
 \SetKwProg{Fn}{Function}{:}{}
 \Fn{\FStep{$x_t$}}{
 \KwRet $u^i_t$\;
 }
  \KwRet mpc\;   
 }
 
 \SetKwFunction{FRun}{RunSimulation}
 \SetKwProg{Fn}{Function}{:}{}
 \Fn{\FRun{}}{
  \For{$t \leftarrow 0$ \KwTo $T_{\text{sim}}$}{
    \For{$i \leftarrow 0$ \KwTo $k$}{
    \FMpc{}\;
    $u^i_t\gets$\FMpc.\FStep{$x_{t}$}\;
   % Compute control input $u^i_t$\;
   \If{single-integrator dynamics \& $\ell_j\left( p^i_t, v^i_t\right) < \ell_\textnormal{thresh}$ \textnormal{\textcolor{red}{(deadlock)}}}{ \If{$v^i_t = v^j_t$}{
   break ties by computing $\sigma_{\textsc{opt}}$ (Section~\mbox{\ref{subsec: auctions}})}
   }
   Perturb $u^i_t$ via Equation~\eqref{eq: perturbation}
   }
    Update state $x^i_{t+1} \gets x^i_t$ via Equation~\eqref{eq: control_affine_dynamics}\;
    }

  }

 Call \FRun{}\;

 \caption{Safe and deadlock-free multi-agent navigation.}
 \label{algo: overall}
\end{algorithm}

The complete algorithm is presented as pseudocode in Algorithm~\ref{algo: overall}. The first step involves initializing the starting and goal positions (\(\mathcal{X}_I\) and \(\mathcal{X}_g\), Line 1). The algorithm then sets key simulation parameters: total simulation time (\(T_{sim}\)), sampling time (\(T_s\)), and prediction horizon (\(T_{horizon} = 3\)) [Line 2]. Following this setup, it defines the cost matrices and actuator limits, including the velocity limit (\(v_{limit} \)) and angular speed limit (\(\omega_{limit}\)) [Line 3]. The model defines the state and control variables, system dynamics based on the cost matrices (\(f, g\)), and stage-wise cost \(\mathcal{J}^i\) [Lines 4-9]. Next, the MPC function is configured, importing the model defined previously. This function sets up the parameters using the simulation settings, defines the objective function, and establishes state and input constraints, including safety constraints [Lines 10-19]. If the system has double-integrator dynamics, a liveness constraint is also added.

Finally, the algorithm executes the main simulation loop [Lines 21-31]. During each time step, the MPC controller computes the safe control input $u^i_t$ for each agent considering the joint state $x_t$, using the MPC controller [Line 25]. If the system operates with single-integrator dynamics and detects a deadlock (Theorem~\ref{thm: smg->l<tau}), the control input (\(u^i_t\)) is perturbed to resolve the deadlock breaking ties if necessary by running the auction described in Section~\mbox{\ref{subsec: auctions}}). The state is then updated for the next time step using the system dynamics. The MPC controller in Line $25$ solves an optimal control problem modified from~\eqref{eq: best_response_example}. The best response, $\left( \Gamma^{i,*}, \Psi^{i,*}\right)$ can be solved by adding Equations~\eqref{eq: discrete_barrier_func} and~\eqref{eq: av_geq_0} as constraints.

% \subsection{Analysis}
% \label{subsec: deadlock_analysis}
% \begin{definition}
%     \textbf{Minimally Invasive Deadlock Prevention (or Resolution)}-- A deadlock preventive (or resolving) control strategy $u^i_t = [v^i_t, \omega^i_t]^\top$ prescribed for robot $i$ at time $t$ is said to be minimally invasive if:
%     \begin{enumerate}
%         \item $\Delta \theta^i_t = 0$ (does not deviate from the preferred trajectory).
%         \item $v_{t+1}^i = v_t^i +\delta$ where $\delta = \arg\min\left\Vert v_t^i + \delta \right\rVert, \delta \in \mathbb{R}$ such that $\ell_j(p_t^i,v_t^i +\delta) \geq \ell_\textnormal{thresh}$ (smallest change in speed that prevents a deadlock).
%     \end{enumerate}
    
%     \label{def: min_invasive}
% \end{definition}

% \begin{theorem}
% The control strategy for Equation~\ref{eq: perturbation} is collision-free if the CBF is buffered.
% \end{theorem}
% \begin{proof}
% When robot slows down, then safe. This follows from the fact that if a path is collision-free, then all sub-paths are also collision-free. When it speeds up, the speed up is $\Delta v^i_t = d = \frac{\mu}{2}\cos\psi$ where $\psi = \tan^{-1}\zeta$. Therefore, by placing a buffer of width $d$ around the obstacles, we can ensure that the speed-up would not escape the safe set.
% \end{proof}

\section{Evaluation and Discussion}
\label{sec: experiments}
% \begin{figure}[t]
%     \centering
%     \begin{subfigure}[h]{0.49\columnwidth}
% \includegraphics[width = \columnwidth]{images/webviz1.png}   
%     \caption{Simple doorway environment.}
%     \label{fig: long_traj}
%     \end{subfigure}
%     %
%     \begin{subfigure}[h]{0.49\columnwidth}
%     \includegraphics[width=\columnwidth]{images/webviz2.png}
%     \caption{Static obstacles in robots path.}
%     \label{fig: clipped_traj}
%     \end{subfigure}
%     \caption{\textbf{Visualizing the local planner:} The orange clusters represent lidar scans of the obstacles, the red cross is the local waypoint, the blue curves represent the 2D trajectory search space, and the red outer curves represent the minimum and maximum curvatures. In Figure~\ref{fig: long_traj}, we highlight that the optimal trajectory is one of the two longer blue curves that extend closer to the goal. In Figure~\ref{fig: clipped_traj}, we show that the search space gets clipped due to an obstacle in front.}
%     \label{fig: mini-games}
%     \vspace{-10pt}
% \end{figure}

We deployed our approach in simulation as well as on physical robots in social mini-games occurring at doorways, hallways, and intersections, and analyze--$(i)$ safety and efficiency compared to local planners based on reactive planning, receding horizon planning, multi-agent reinforcement learning, and auction-based scheduling and $(ii)$ the smoothness of our deadlock resolution approach compared to alternative perturbation strategies.

\subsection{Experiment Setup}
\label{subsec: exp_setup}
We numerically validate our approach on differential drive robots in social mini-games that occur at doorways, hallways, intersections, and roundabouts, and analyze its properties. We use the IPOPT solver~\cite{wachter2006implementation} for solving the MPC optimization. We consider the following differential drive robot:
% \begin{align}
% \left[\begin{array}{c}
% \dot{p}^{i,1} \\
% \dot{p}^{i,2} \\
% \dot{\phi}^i\\
% \dot{v}^i\\
% \dot{\omega}^i
% \end{array}\right]=\left[\begin{array}{cccc}
% \cos \phi^i & 0 &0 & 0 \\
% \sin \phi^i & 0 & 0 & 0 \\
% 0 & 1 & 0 & 0\\
% 0 & 0 & 1/m & 0\\
% 0 & 0 & 0 & I^{-1}
% \end{array}\right]\left[\begin{array}{c}
% v^i \\
% \omega^i\\
% u_1\\
% u_2
% \end{array}\right],
% \label{eqn:conti_drive_robot}
% \end{align}

\begin{align}
\left[\begin{array}{c}
\dot{p}^{i,1} \\
\dot{p}^{i,2} \\
\dot{\phi}^i\\
\dot{v}^i\\
\dot{\omega}^i
\end{array}\right]=\left[\begin{array}{cccc}
\cos \phi^i & 0 &0 & 0 \\
\sin \phi^i & 0 & 0 & 0 \\
0 & 1 & 0 & 0\\
0 & 0 & 1/m & 0\\
0 & 0 & 0 & I^{-1}
\end{array}\right]\left[\begin{array}{c}
v^i \\
\omega^i\\
u_1\\
u_2
\end{array}\right],
\label{eqn:conti_drive_robot}
\end{align}

where subscript $i$ denotes the $i^{\text{th}}$ agent, $m$ and $I$ are the mass and moment of inertia about the center of mass respectively, $\left[p^{i,1}, p^{i,2} \right]^\top \in \mathbb{R}^2$ represent the position vector of the robot, $\phi^i \in \mathbb{S}^1$ represents its orientation, $v^i, \omega^i$ are the linear and angular velocities of the robot, respectively, and $u^i=[u_1,\;u_2]^\mathrm{T} \in \mathcal{U}^i$ is the control input. The discrete-time dynamics of \eqref{eqn:conti_drive_robot} can be described by:
\begin{align}
x^i_{t+1}=f\left(x^i_t\right)+g\left(x^i_t\right)u^i_t
\end{align}
where the sampling time period $\Delta T=0.1s$, $x^i_t=[p^{i,1},\;p^{i,2},\;\phi^i,\;v^i,\;\omega^i]^\mathrm{T}$ is the state and $u^i_t=[u_1,\;u_2]^\mathrm{T}$ is the control input.
The objective is to compute control inputs that solve the following non-linear optimal control problem minimize the following cost function,
\begin{equation}
    \begin{split}
        &\underset{u_{1:T-1}}{\min}\;\;\sum_{t=0}^{T-1}{x^i_t}^\top Q{x^i_t}+{u^i_t}^\top Ru^i_t\\
    \text{s.t}\quad &x^i_{t+1}=f\left(x^i_t\right)+g\left(x^i_t\right)u^i_t,\quad\forall\;t\in[1;T-1]\\
&x_t\in\mathcal{X},\;u_t\in\mathcal{U}^i\quad\forall\;t\in\{1,\dots,T\}
    \end{split}
\end{equation}
The safety for the differential drive robot $i$ is guaranteed by the satisfaction of CBF constraint \eqref{eq: game_safety_constraint} as we consider control inputs that belong to $\mathscr{U}^i$. For each agent, the CBF for the obstacles is characterized by $h_s\left(x^{i,m}_t\right)$ given by
\begin{align}
    h_s\left(x^{i,m}_t\right)=\left(p^{i,1}-c^{1,m}\right)^2+\left(p^{i,1}-c^{2,m}\right)^2-r^2\nonumber
\end{align}
where $r>0$, $x^i_t$ is specified by $\left[p^i_t, \theta^i_t,v^i_t, \omega^i_t \right]\in \mathbb{R}^2 \times \mathbb{S}^1\times \mathbb{R}^2\times \mathbb{R}$ representing the current position, heading, linear and angular rates of the $i^\textrm{th}$ robot and $(c_{1,m},c_{2,m})$ is the center of a circle with radius $r>0$ and $m\in\{1,\dots, M\}$ where $M$ is sufficiently large to cover all the obstacles. Therefore, the safe region are characterized by the set \begin{align}
    \mathcal{X}=\left\{x^i_t:\;h_s\left(x^{i,m}_t\right)>0,\;\forall\; m\in\left\{1,\dots, M\right\}\right\}\nonumber
\end{align}
Further, to avoid collisions with another agent, each agent $i$ treats the other agent $j$ ($i\neq j$) as an obstacle. Consequently, the CBF for agent $i$ is given by Equation~\eqref{eqn:CA_h}. 
% \begin{align}
%     h^a_i(x,y)=(x-p^{j,1})^2+(y-p^{j,2})^2-r^2,\quad i\neq j
% \end{align}
The CBF $h_v\left(x_t\right)$ for an agent $i$ is given by
\begin{align}
    h_v(x^i_t)=Av_t,\quad i\neq j
\end{align}
where $v_t=[v^i_t,\;v^j_t]^\top$, $A\in\mathbb{R}^{2\times 2}$ for all the experiments with two robots, but $A\in\mathbb{R}^{3 \times 3}$ matrix for the experiments with $3$ robots. For our doorway and intersection simulations, we choose $A=\begin{bsmallmatrix}
    1&-2\\-2& 1
\end{bsmallmatrix}$.

\subsection{Environments, Robots, Baselines, and Metrics}

% \noindent\textbf{Simulation}--We use the SocialGym 2.0 multi-agent social navigation simulator~\cite{socialgym}, which is available on Github\footnote{\href{https://github.com/ut-amrl/social\_gym}{https://github.com/ut-amrl/social\_gym}.}

\noindent\textbf{Environment}—Our approach has been rigorously designed and subjected to tests within two socially relevant mini-games: a doorway passage and a corridor intersection, both confined within a spatial area measuring $3$ meters by $3$ meters. In the doorway setting, we calibrated the gap size to an approximate width of $0.5$ meters. In the corridor intersection scenario, the arms branching from the central conflict zone have been strategically set to range between $1.5$ and $2$ meters in width, while the conflict zone itself occupies an area between $2.5$ to $4$ square meters. For the doorway experiment, robots initiate their navigation from one side uniformly spaced at an approximate distance of $1.8$ meters from the gap aiming to safely transit to the opposing side. Conversely, the corridor intersection tests focus on autonomous navigation through the intersection with a rotational presence of one robot on each arm. Robots have the freedom to navigate toward any of the remaining three arms, the overarching goal being secure passage through this complex spatial layout. 
% For simulation methods, we used a CPU

\noindent\textbf{Robots}—Our empirical evaluations involved multi-robot trials conducted on an eclectic mix of robotic platforms: UT Automata F$1/10$ car platforms, Clearpath Jackal, and Boston Dynamics Spot. These platforms were also engaged in human-robot interaction tests, specifically using the Jackal. The robots were judiciously selected to capture a diverse range of shapes, sizes, and kinodynamic behaviors. The Spot distinguishes itself as a legged robot, while both the Jackal and F$1/10$ cars are wheeled, though the Jackal has the unique capability of executing point turns, which the cars lack. Speed capabilities also vary: both the Spot and the Jackal have a maximum speed limit of $1.5$ meters per second, whereas the UT Automata F$1/10$ cars can accelerate up to $9$ meters per second. Configurations were experimented with broadly, the sole exception being the co-deployment of Spot and Jackal with the F$1/10$ cars, as the latter are not detectable by the larger robots due to their smaller size.
\begin{figure*}[t]
    \centering
    \begin{subfigure}[h]{0.325\linewidth}
\includegraphics[width = \linewidth]{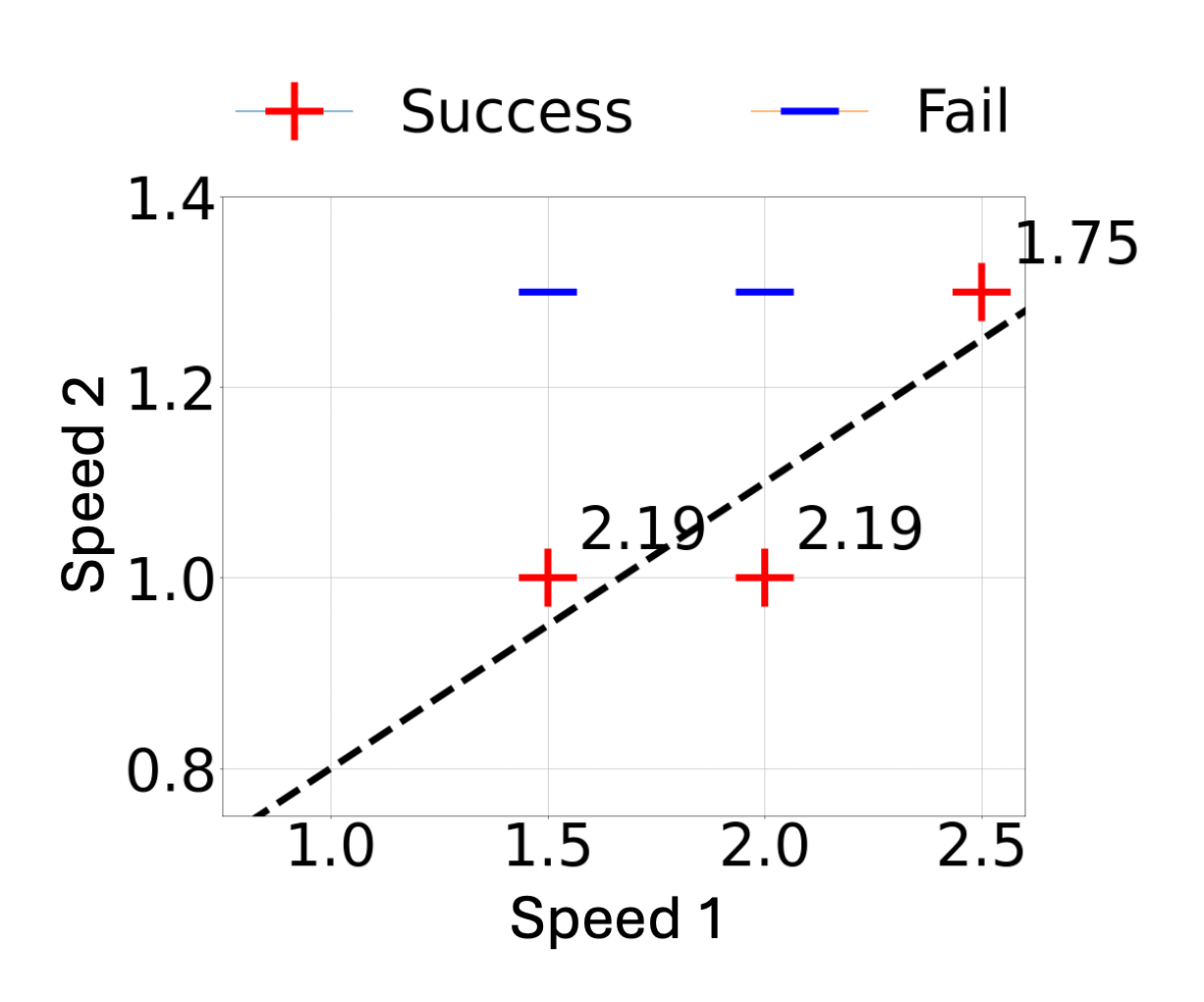}    
    \caption{F$1/10$, door.}
    \label{fig: doorway_result}
    \end{subfigure}
    \begin{subfigure}[h]{0.325\linewidth}
\includegraphics[width = \linewidth]{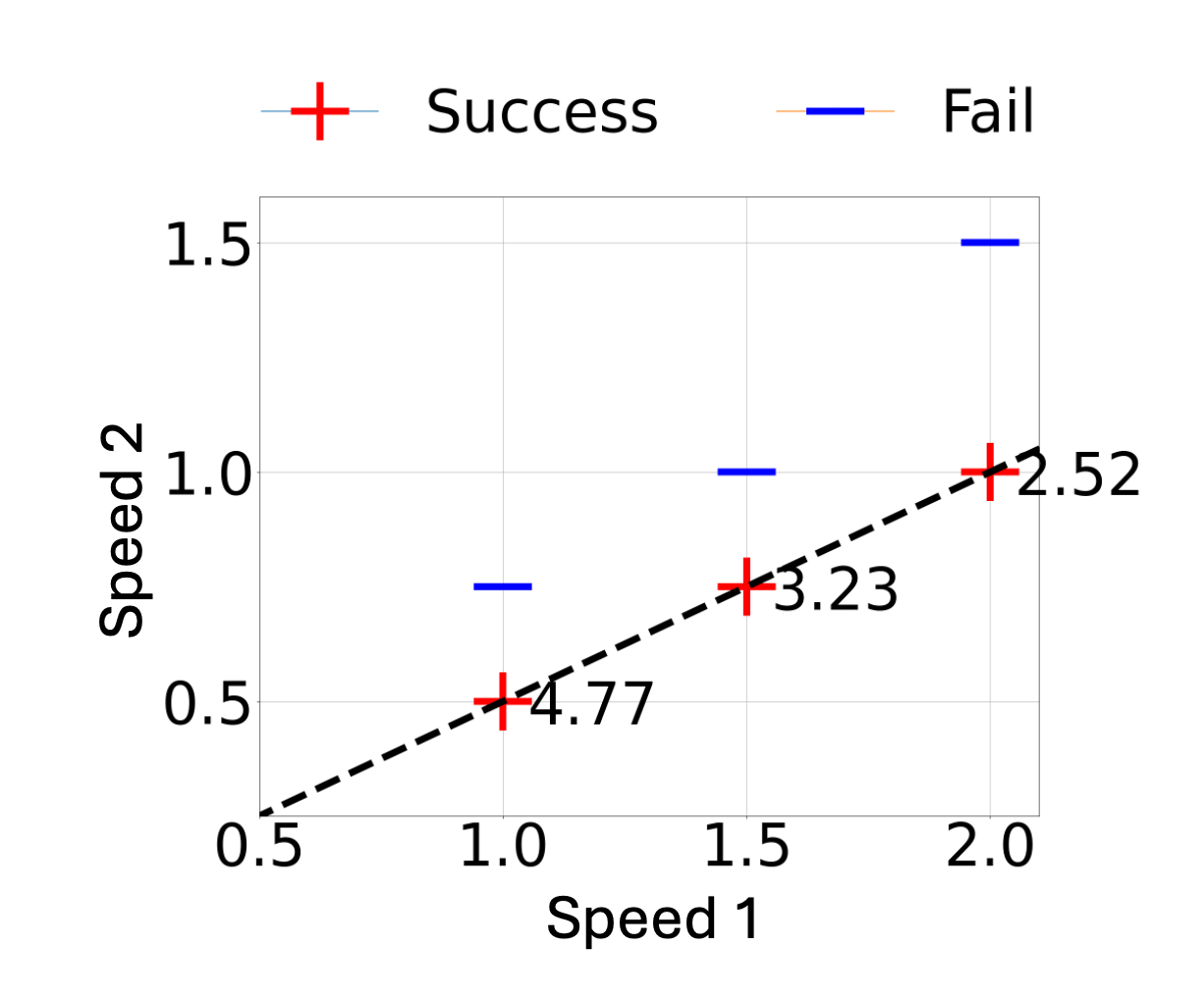}    
    \caption{F$1/10$, intersection.}
    \label{fig: intersection_result}
    \end{subfigure}
    \begin{subfigure}[h]{0.325\linewidth}
\includegraphics[width = \linewidth]{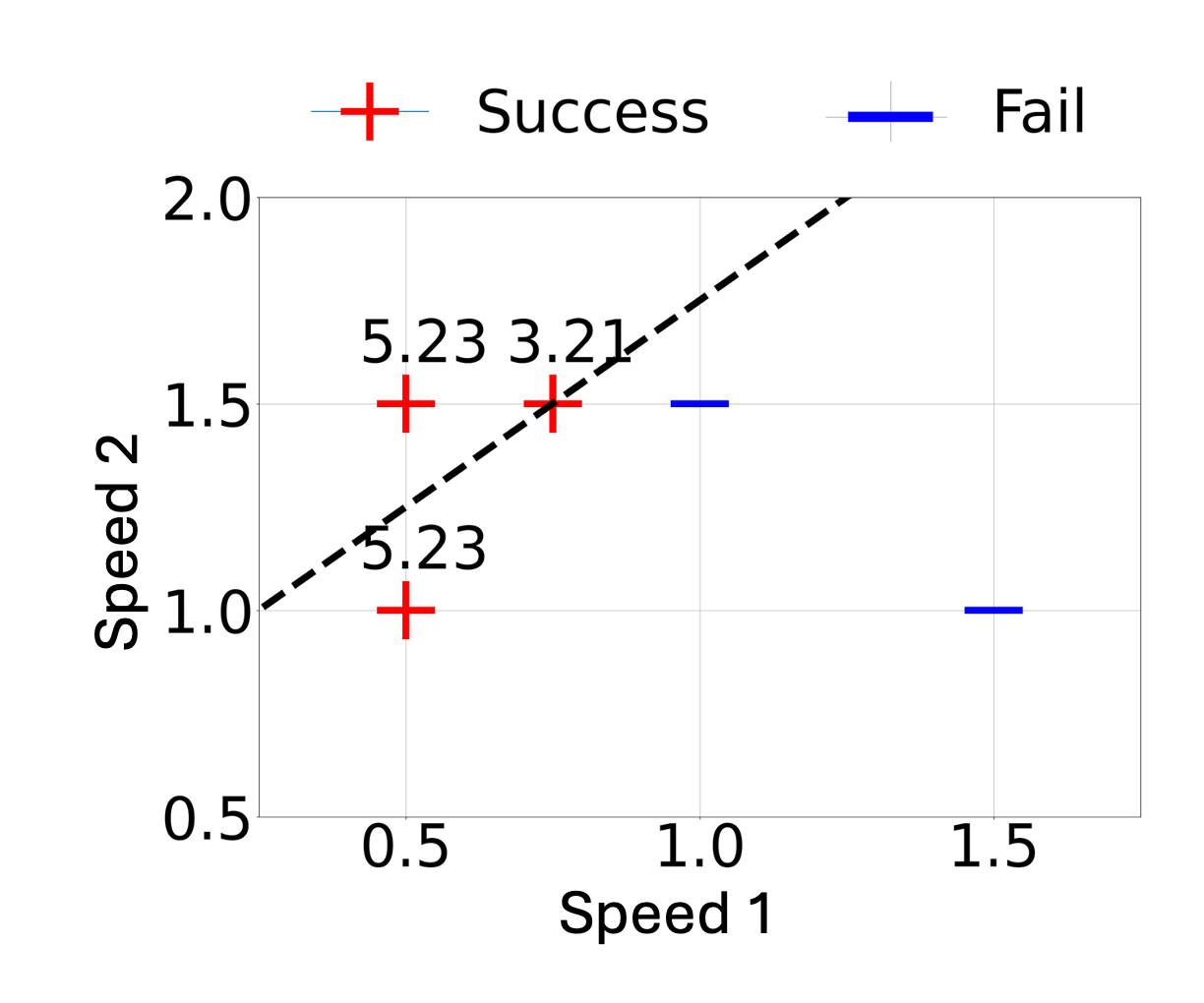}   
    \caption{Spot/Jackal, door.}
    \label{fig: jackal_result}
    \end{subfigure}
    \caption{\textbf{Generating liveness sets using $2$ F$1/10$ car platforms, Jackal, and the Spot for the doorway social mini-game:} We record successful and failed trials along with the makespan in the case of successful trials. Figures~\ref{fig: doorway_result} and~\ref{fig: intersection_result} use F$1/10$ cars while Figure~\ref{fig: jackal_result} shows results for the Jackal versus the Spot. We notice that robots succeed when the velocities  (``velocity 1'' and ``velocity 2'') satisfy Equation~\ref{eq: av_geq_0}.}
    \label{fig: main_exps}
    % \vspace{-10pt}
\end{figure*}

\noindent\textbf{Evaluation Baselines}--In evaluating our novel local navigation algorithm designed for physical robots operating in real-world environments, we chose to compare it against the dynamic window approach (DWA), a standard local navigation algorithm provided by the \texttt{move\_base} package in the Robot Operating System (ROS)\footnote{\href{http://wiki.ros.org/move\_base}{http://wiki.ros.org/move\_base.}}. This decision was motivated by the fact that \texttt{move\_base} serves as a well-regarded baseline that is both widely used and accepted in the robotics community, thereby allowing us to gauge the relative merits of our approach in a context that is immediately understandable and relevant to researchers and practitioners alike. 

In our simulation-based experiments, we compare our proposed algorithm against both centralized and decentralized local planning algorithms to foster a multifaceted comparison. On the centralized front, we compare with ORCA-MAPF~\cite{orcamapf} that integrates decentralized collision avoidance facilitated by ORCA with centralized Multi-Agent Path Finding (MAPF) to robustly resolve deadlock. Decentralized planners, on the other hand, can be sub-categorized into reactive and receding-horizon approaches. Within the reactive domain, we examine a diverse suite of methods that encompass quadratic programming-based controllers~\cite{wang2017safety}, NH-TTC~\cite{davis2019nh}, and NH-ORCA~\cite{nh-orca}. We also compare with learning-based methods, such as CADRL~\cite{cadrl, sacadrl}. 
% Additionally, we include a recent approach, SNUPI~\cite{snupi}, a paradigm that leverages auction mechanisms to determine a navigational priority sequence for agents within social mini-games.
For receding-horizon planners, our selection incorporates DWA, courtesy of the \texttt{move\_base} package in ROS, MPC-CBF~\cite{zeng2021safety_cbf_mpc}, and IMPC~\cite{impc}, for its Model Predictive Control (MPC) framework coupled with deadlock resolution capabilities.

\noindent\textbf{Evaluation Metrics}—To comprehensively assess navigation algorithms, we employ a carefully curated set of metrics, each averaged across multiple trials, that analyze three essential dimensions: safety, smoothness, and efficiency. We incorporate metrics such as the success rate, collision rate, Avg. $\Delta V$, path deviation, makespan ratio, and specific flow rate. Avg. $\Delta V$ represents the average change in velocity from the starting point to the destination, measuring how fluidly a robot can adjust its velocity while still reaching its destination, without the risk of frequent stops or abrupt changes. Path deviation measures the deviation from a robot's preferred trajectory. Makespan ratio encapsulates the time-to-goal disparity between the first and last agent, and specific flow rate offers a macroscopic view, quantifying the overall `volume of agents' successfully navigating through a constrained space, such as a doorway. 

While metrics like Avg. $\Delta V$ and path deviation evaluate an individual robot's behavior, specific flow rate captures the collective efficiency of multiple agents navigating through constrained spaces, such as doorways. This distinction is significant because achieving a flow rate similar to that of human crowds implies that the navigation algorithm under examination is not merely efficient at an individual level, but also highly effective in optimizing multi-agent movement.

% In addition, we also simulate a classical planner using the dynamic window approach as well as an auction-based approach~\cite{snupi}. CADRL~\cite{cadrl} and its LSTM-based variant, which we denote as CADRL(L), are state-of-the-art multi-agent social navigation methods. CADRL and CADRL(L) use a reward function where a robot is rewarded upon reaching the goal and penalized for getting too close or colliding with other robots as well as taking too long to reach the goal. Enforced Order is a MARL baseline that encourages robots to engage in social behaviors such as queue formations. We train these baselines using PettingZoo and Stable Baselines3~\cite{sb3} and report results across a range of social navigation metrics. The metrics, each averaged across the number of episodes include success rate, time still, and average $\Delta$ velocity. The time still measures the number of times the robots had to stop and the average $\Delta$ velocity measures the numbers of times the robots had to deviate from their previous velocity.

\subsection{Liveness Sets}

We empirically generate the liveness sets for the $2$ agent and $3$ agent social mini-games occurring at doorways and intersections. We deploy different combinations of robots with varying velocities in the social mini-games and the distribution of velocities that results in safe navigation is the empirical liveness set for that social mini-game. In Figure~\ref{fig: doorway_result}, we report the outcomes of trials conducted in the doorway mini-game using $2$ F$1/10$ car platforms. We conducted $5$ trial runs, averaged across $3$ iterations, with varying velocities of which $3$ succeeded and $2$ failed. We observed that the trials succeeded when the cars' velocities satisfied Equation~\eqref{eq: av_geq_0}, whereas the trials failed when the velocities did not satisfy Equation~\eqref{eq: av_geq_0}. We repeated this experiment in the intersection scenario as shown in Figure~\ref{fig: intersection_result} and obtained an identical liveness set. With $3$ F$1/10$ robots we obtained a liveness set similar to Equation~\eqref{eq: av_3}.

\begin{figure*}
    \centering
    \includegraphics[width=\linewidth]{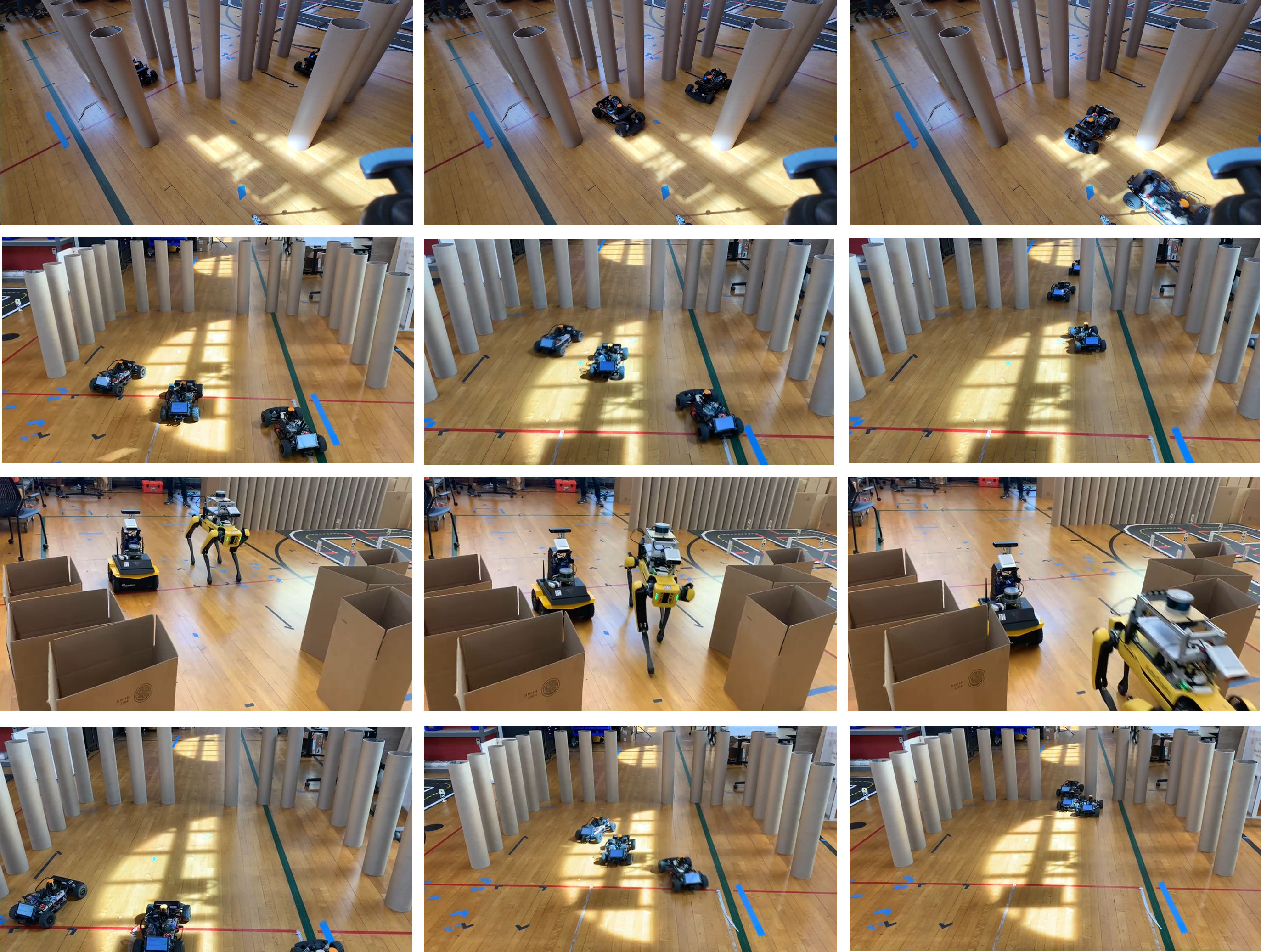}
    \caption{\textbf{Deployment in multi-robot scenarios:} \textit{(rows $1,2,3$)} We demonstrate our approach at a corridor intersection and doorway using the F$1/10$ cars, Spot, and the Jackal robots. \textit{(row $4$)} DWA from the ROS \texttt{move\_base} package results in a collision. 
    }
    \label{fig: qualitative}
    % \vspace{-10pt}
\end{figure*}

Finally, we show that the liveness set given by Equation~\eqref{eq: av_geq_0} for the social mini-game at doorways holds for cross robot platforms. In Figure~\ref{fig: jackal_result}, we demonstrate the Spot and the Jackal robots in the doorway scenario using an identical setup as with the F$1/10$ platforms and report the results. To gather more evidence for the liveness sets, we scaled the velocities by $1/\zeta^i$ in SocialGym 2.0\footnote{\href{https://github.com/ut-amrl/social\_gym}{https://github.com/ut-amrl/social\_gym}.} and observed similar results. That is, using a $1/\zeta^i$ velocity scaling yields a $76\%$ and $96\%$ success rate in the doorway and intersection scenario, respectively. The variability and reduction in the success rates is due to the stochastic nature of the reinforcement learning control policies employed in SocialGym 2.0, a reinforcement learning simulator, that are unable to provide hard guarantees on safety. We clarify that the purpose of the experiments in SocialGym 2.0 is only to validate the concept of liveness sets. That is, in those instances where the agents did not collide due to stochasticity, agents respected the liveness sets.

% We present qualitative results of these experiments in Figure~\ref{fig: qualitative}.
\subsection{Real World Experiments}

\begin{figure*}
\centering
    \begin{subfigure}[h]{0.32\linewidth}
    \includegraphics[width=\textwidth]{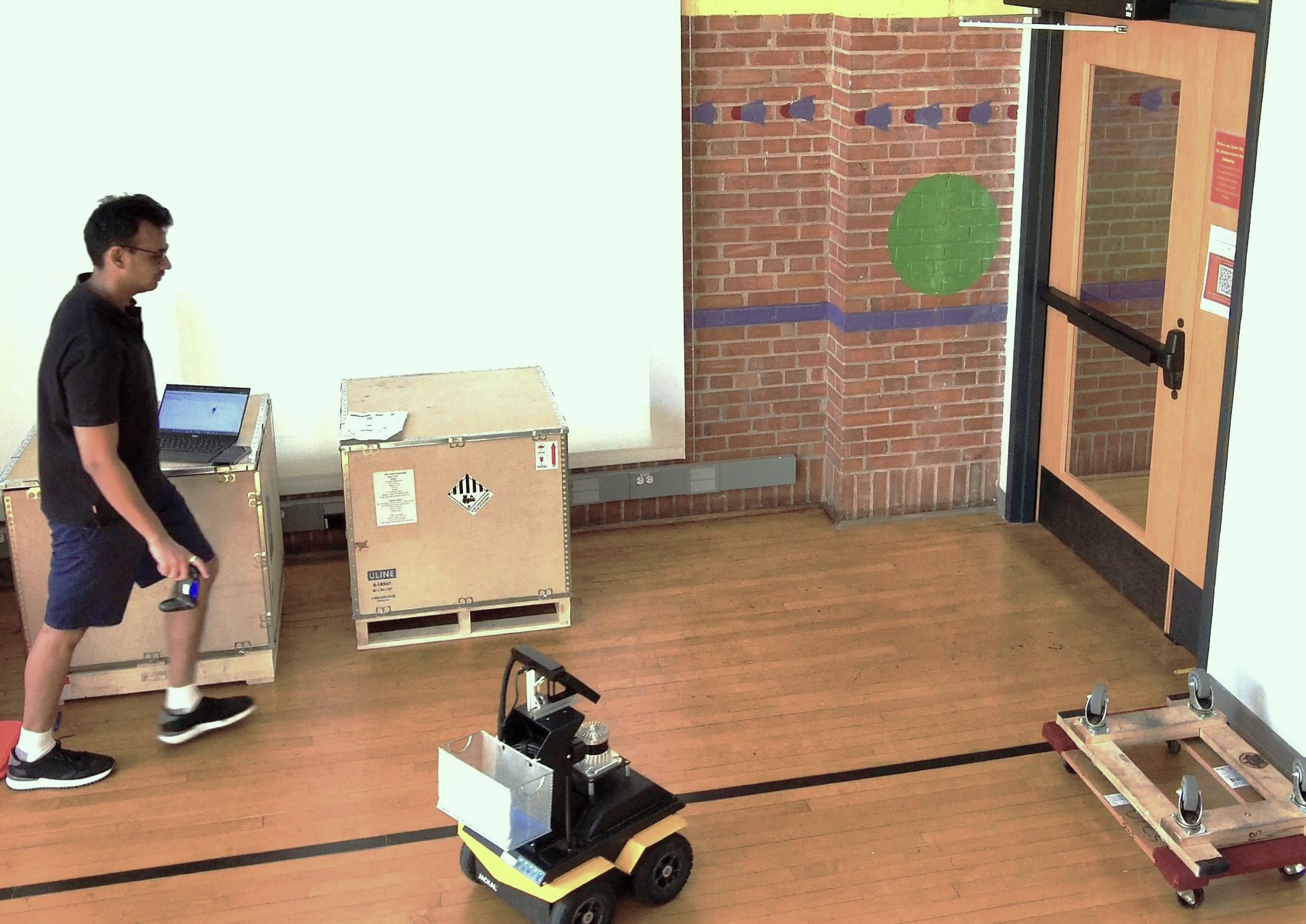}
    \caption{\textit{Our approach:} Initial time step.}
    \label{fig: pass0}
  \end{subfigure}
 \begin{subfigure}[h]{0.32\linewidth}
    \includegraphics[width=\textwidth]{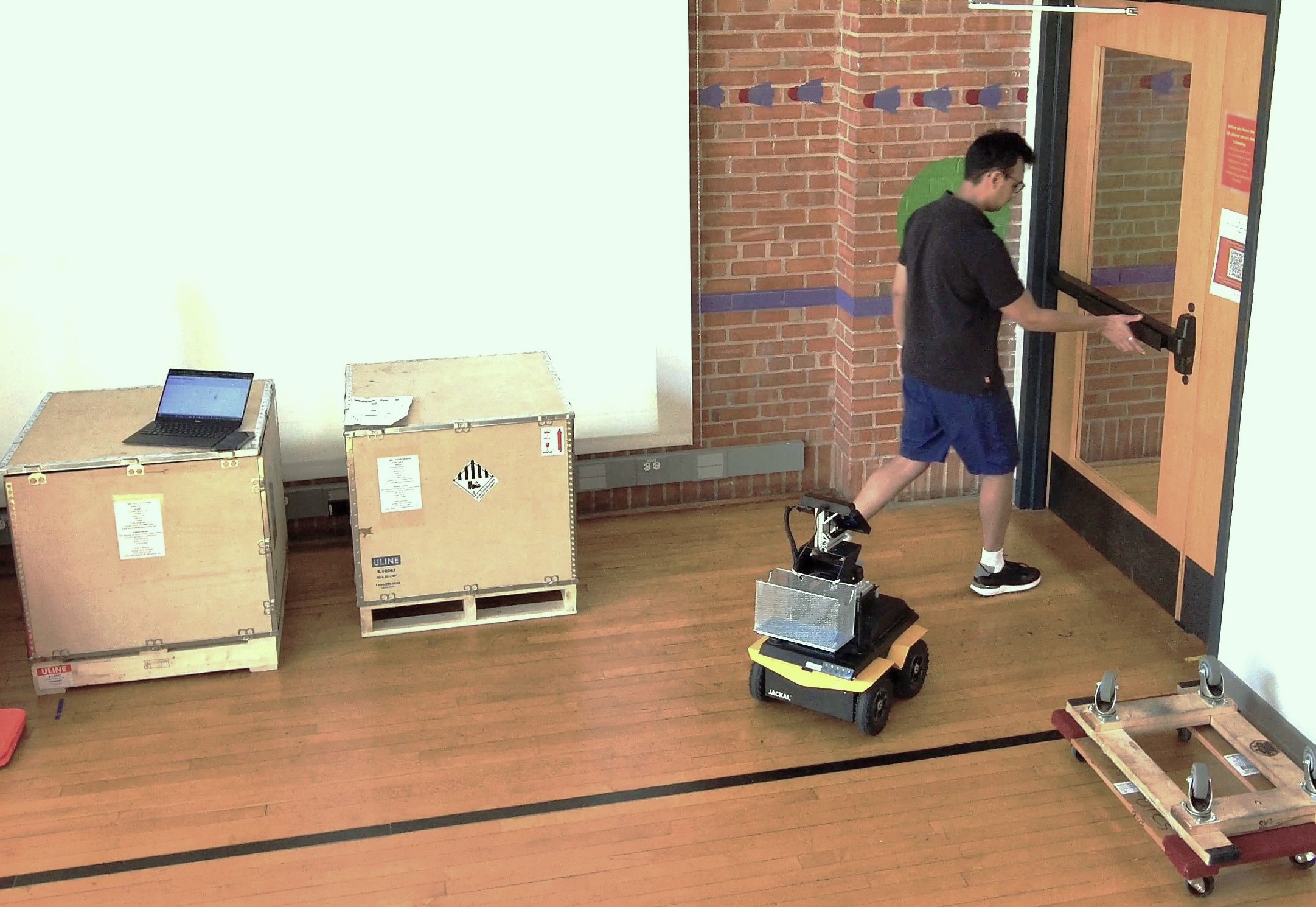}
    \caption{\textit{Our approach:} Robot yields to human.}
    \label{fig: pass1}
  \end{subfigure}
\begin{subfigure}[h]{0.32\linewidth}
    \includegraphics[width=\textwidth]{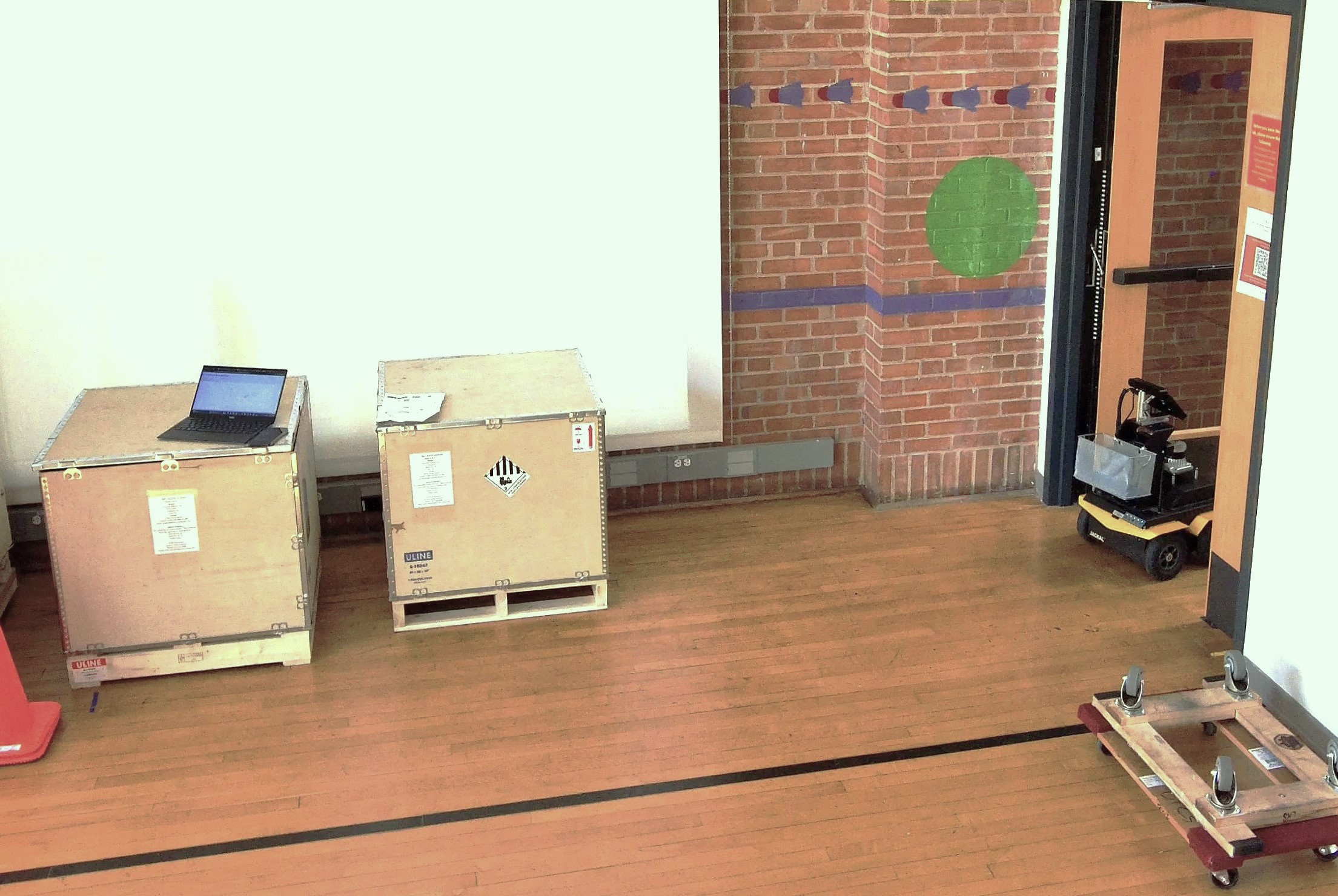}
    \caption{\textit{Our approach:} Robot follows through.}
    \label{fig: pass2}
  \end{subfigure}
   \begin{subfigure}[h]{0.32\linewidth}
    \includegraphics[width=\textwidth]{images/realworld/pass-1_bright.jpg}
    \caption{\textit{DWA:} Initial time step.}
    \label{fig: fail0}
  \end{subfigure}
     \begin{subfigure}[h]{0.32\linewidth}
    \includegraphics[width=\textwidth]{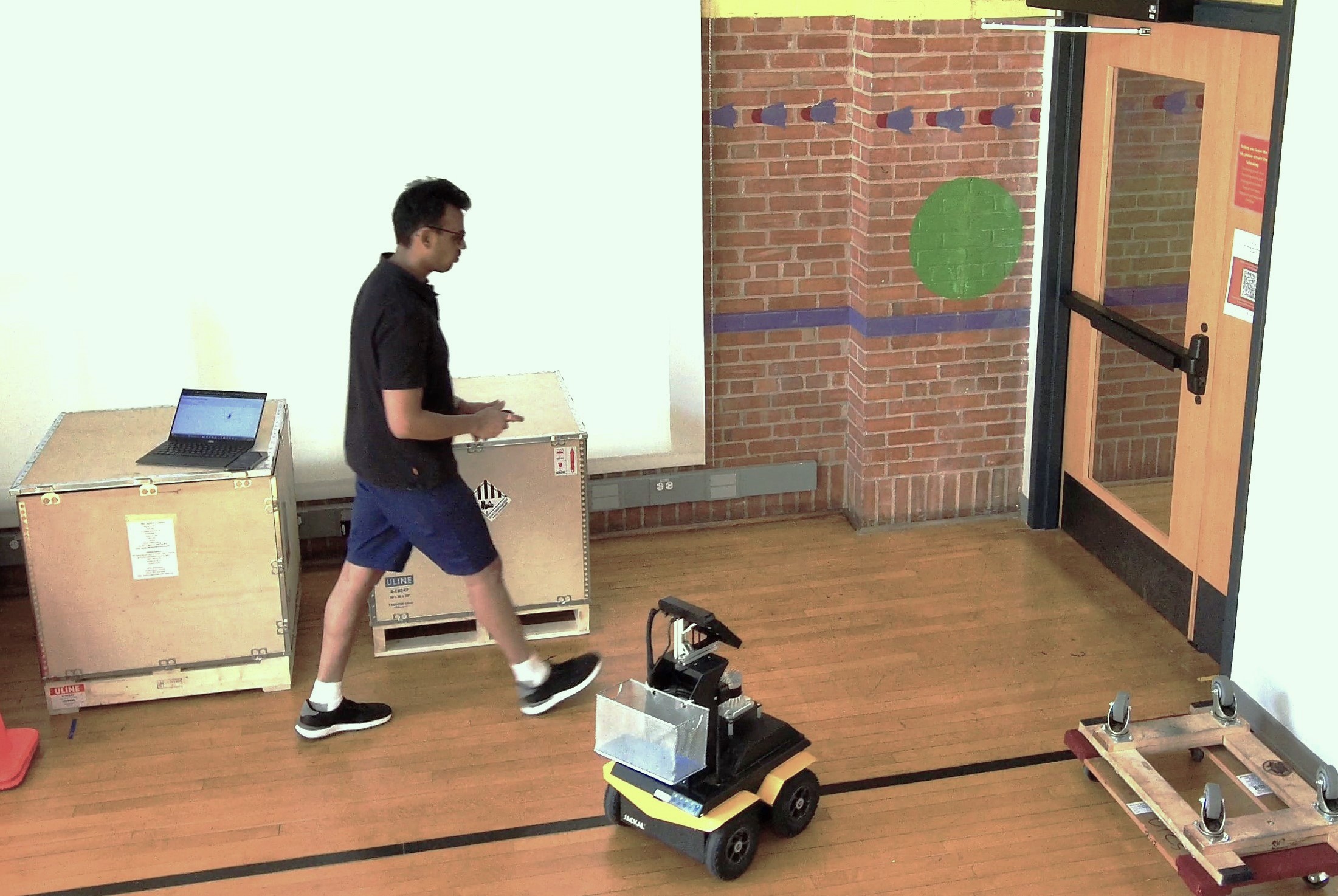}
    \caption{\textit{DWA:} Robot does not yield to human.}
    \label{fig: fail1}
  \end{subfigure}
 \begin{subfigure}[h]{0.32\linewidth}
    \includegraphics[width=\textwidth]{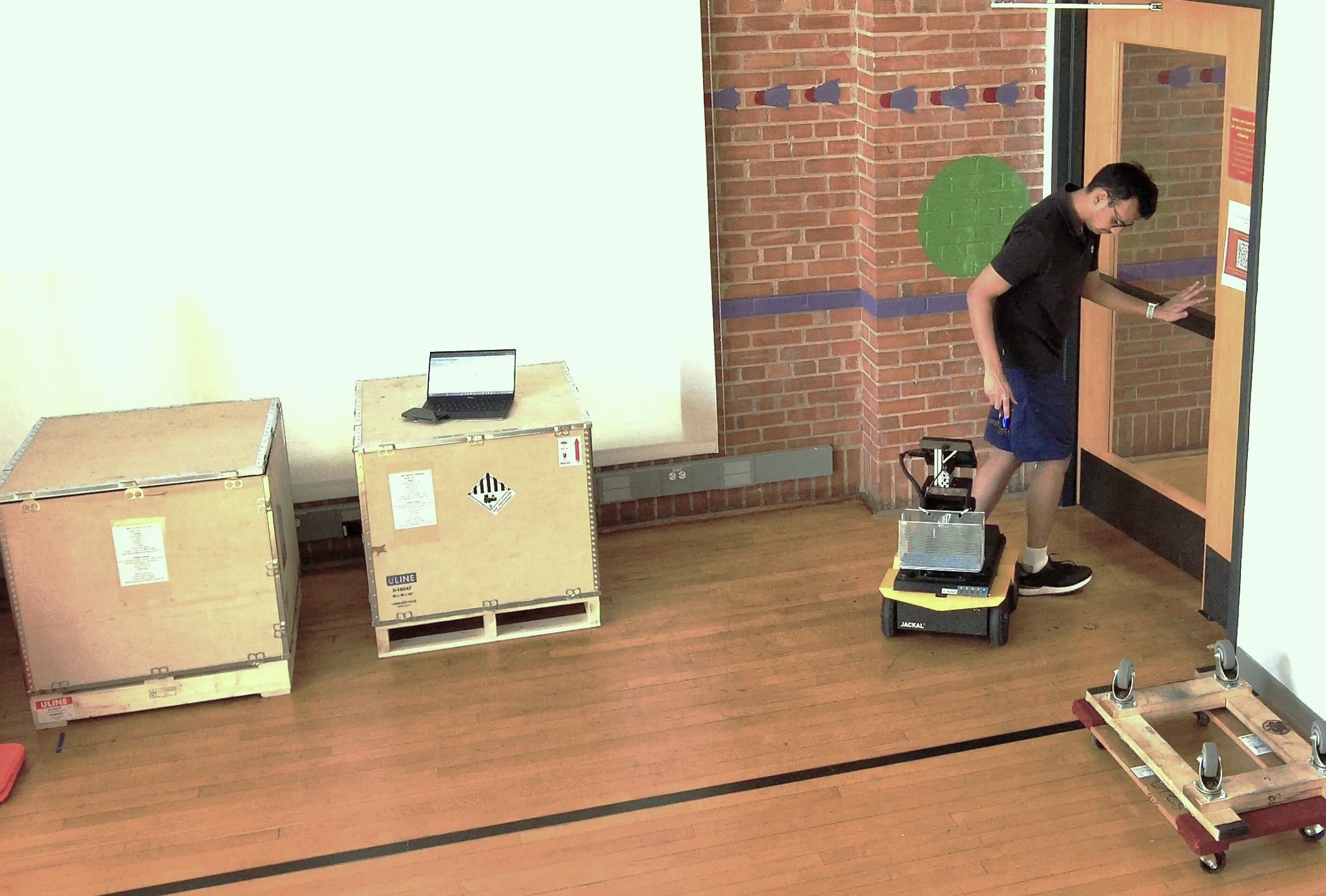}
    \caption{\textit{DWA:} Robot collides with human.}
    \label{fig: fail2}
  \end{subfigure}
   \begin{subfigure}[h]{0.32\linewidth}
    \includegraphics[width=\textwidth]{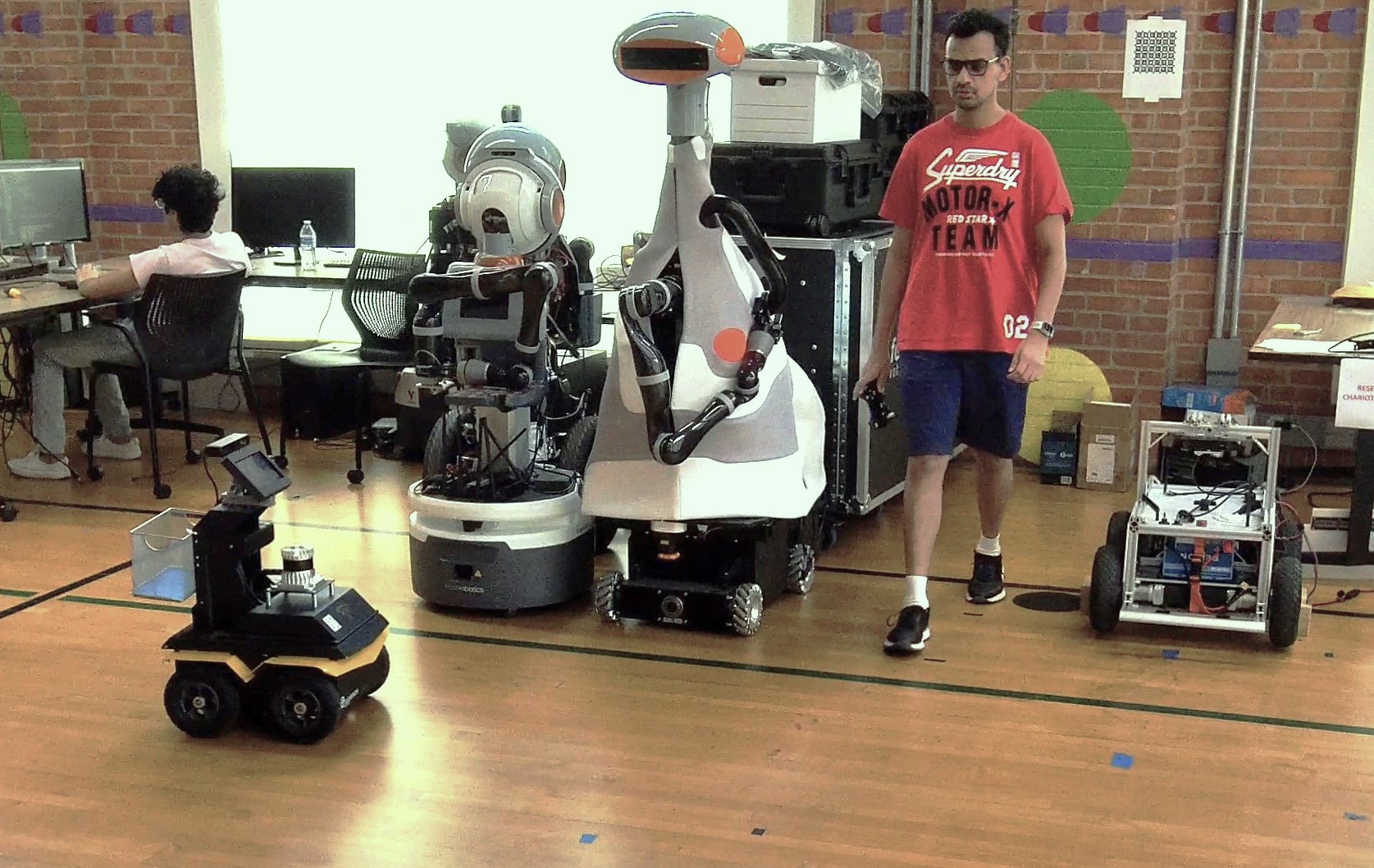}
    \caption{\textit{Our approach:} Initial time step.}
    \label{fig: fail1i}
  \end{subfigure}
     \begin{subfigure}[h]{0.32\linewidth}
    \includegraphics[width=\textwidth]{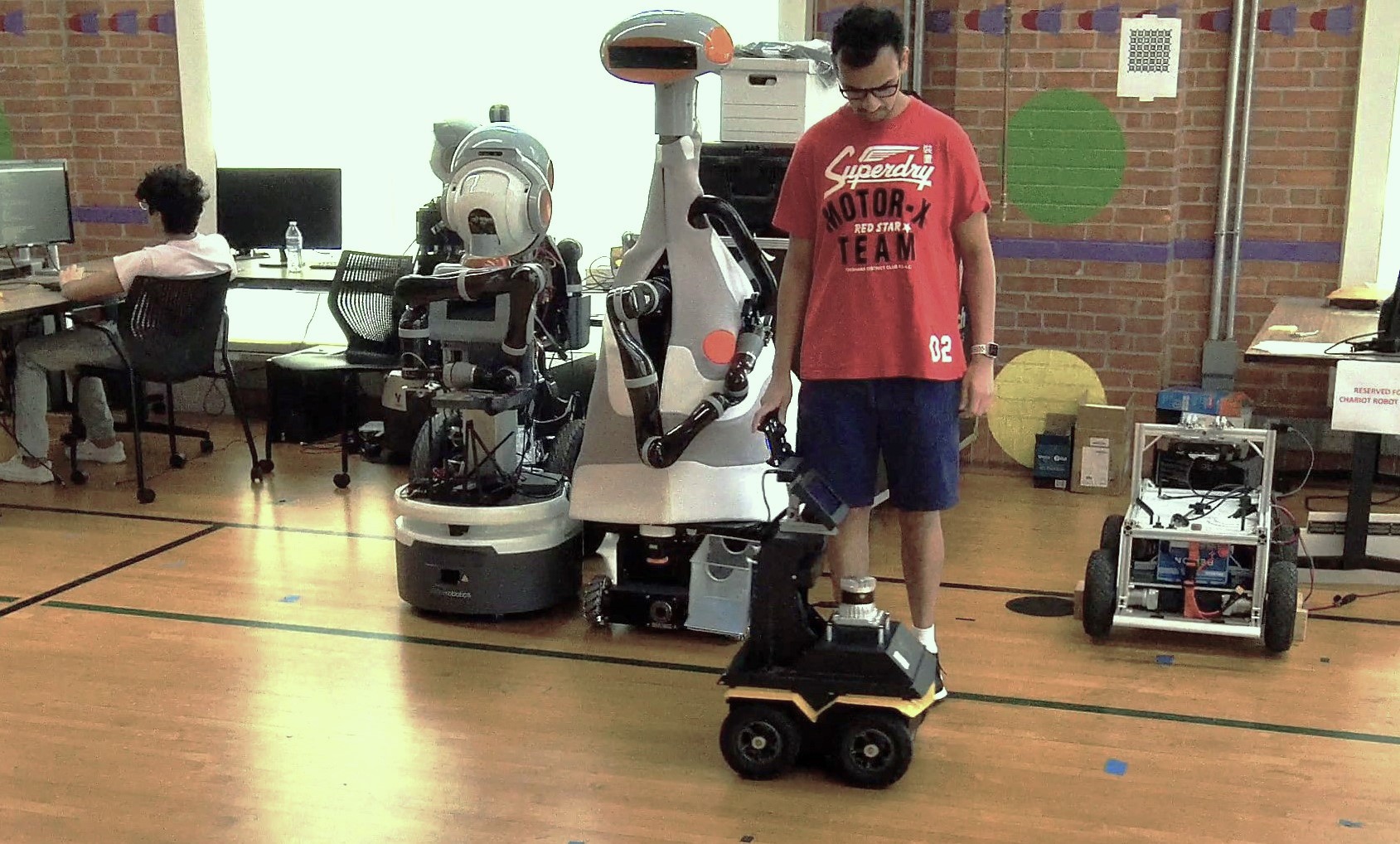}
    \caption{\textit{Our approach:} Human slows down.}
    \label{fig: fail2i}
  \end{subfigure}
 \begin{subfigure}[h]{0.32\linewidth}
    \includegraphics[width=\textwidth]{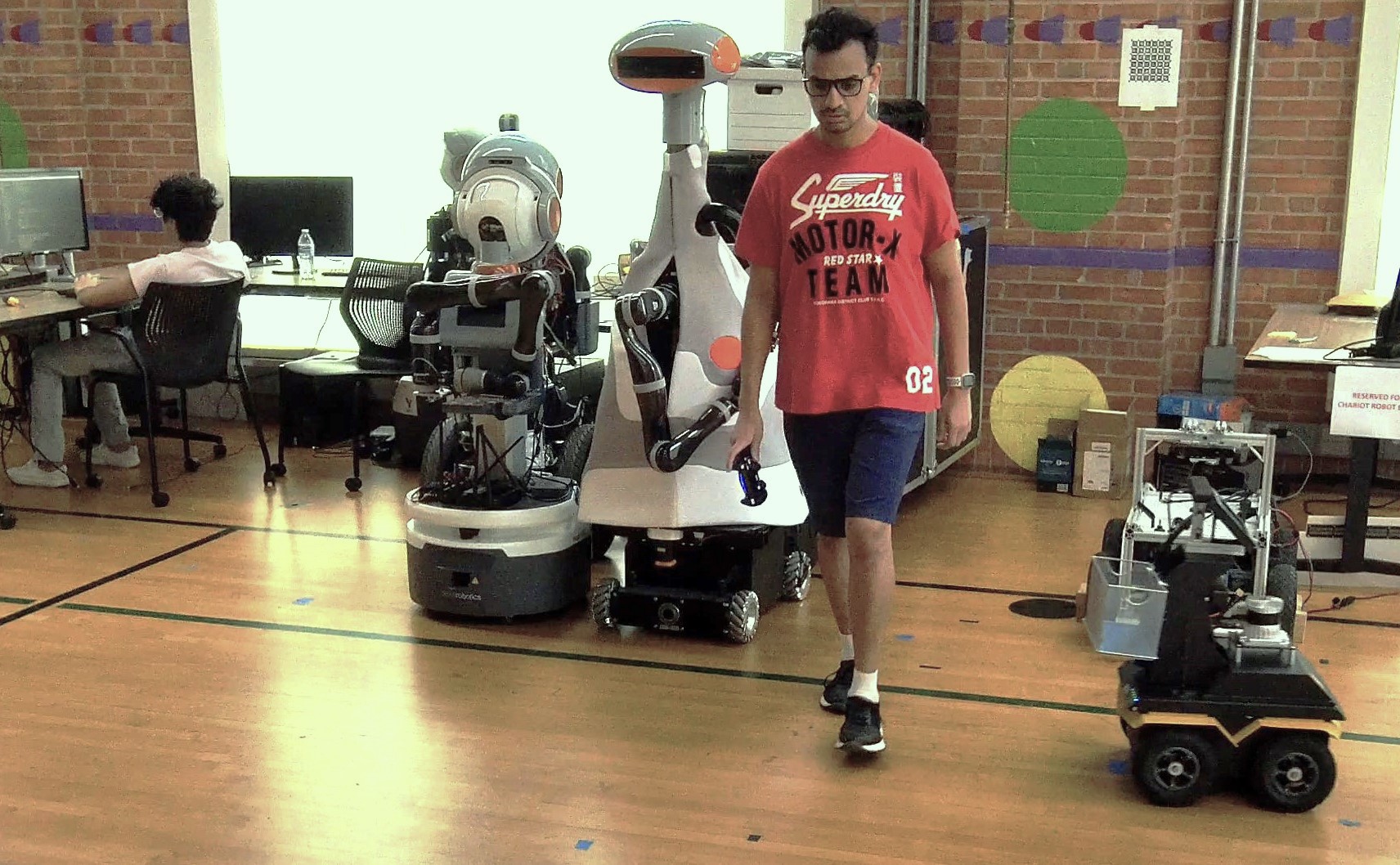}
    \caption{\textit{Our approach:} Robot speeds up.}
    \label{fig: fail3i}
  \end{subfigure}
    \begin{subfigure}[h]{0.32\linewidth}
    \includegraphics[width=\textwidth]{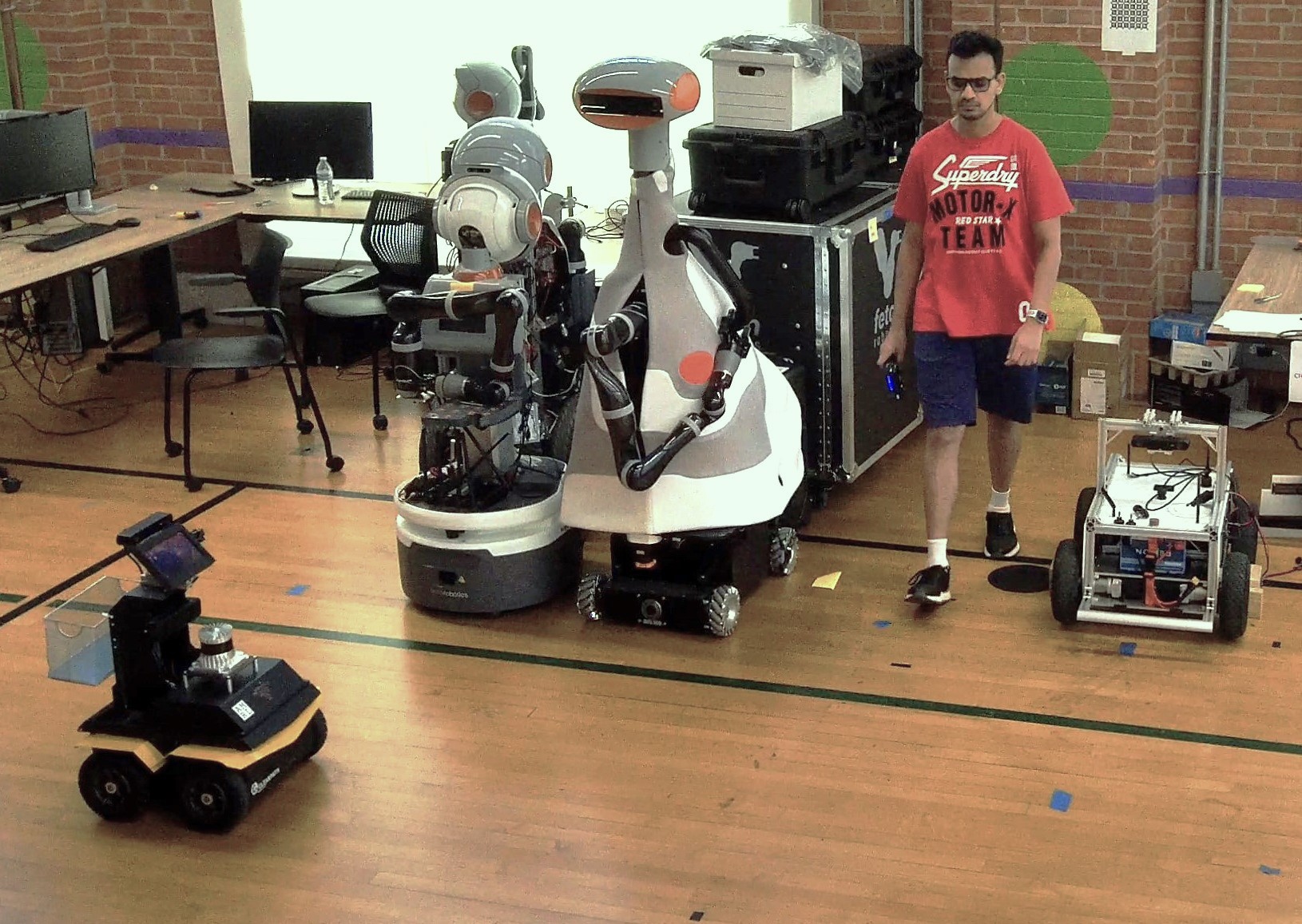}
    \caption{\textit{DWA:} Initial time step.}
    \label{fig: pass1i}
  \end{subfigure}
 \begin{subfigure}[h]{0.32\linewidth}
    \includegraphics[width=\textwidth]{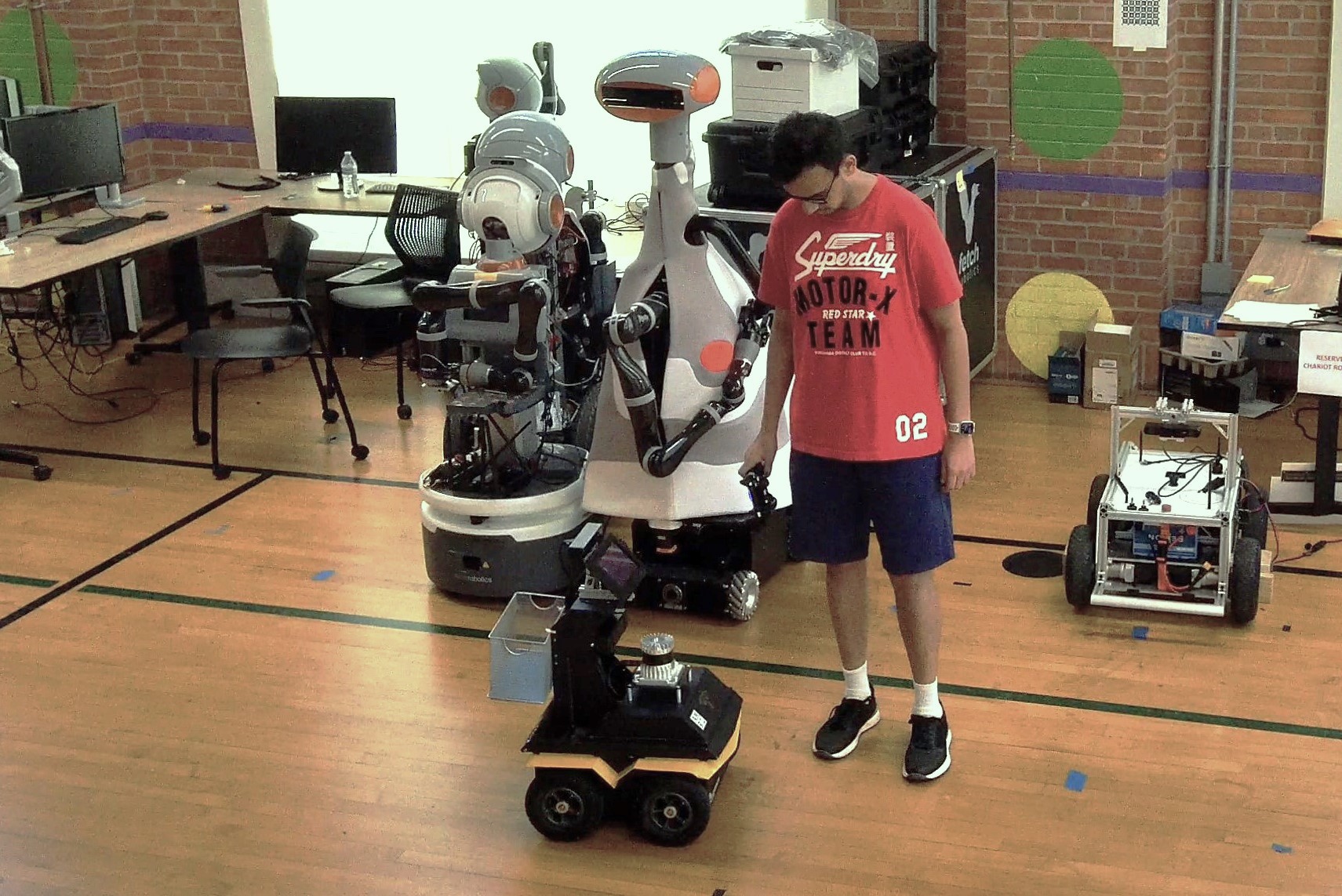}
    \caption{\textit{DWA:} Robot stops when human slows.}
    \label{fig: pass2i}
  \end{subfigure}
\begin{subfigure}[h]{0.32\linewidth}
    \includegraphics[width=\textwidth]{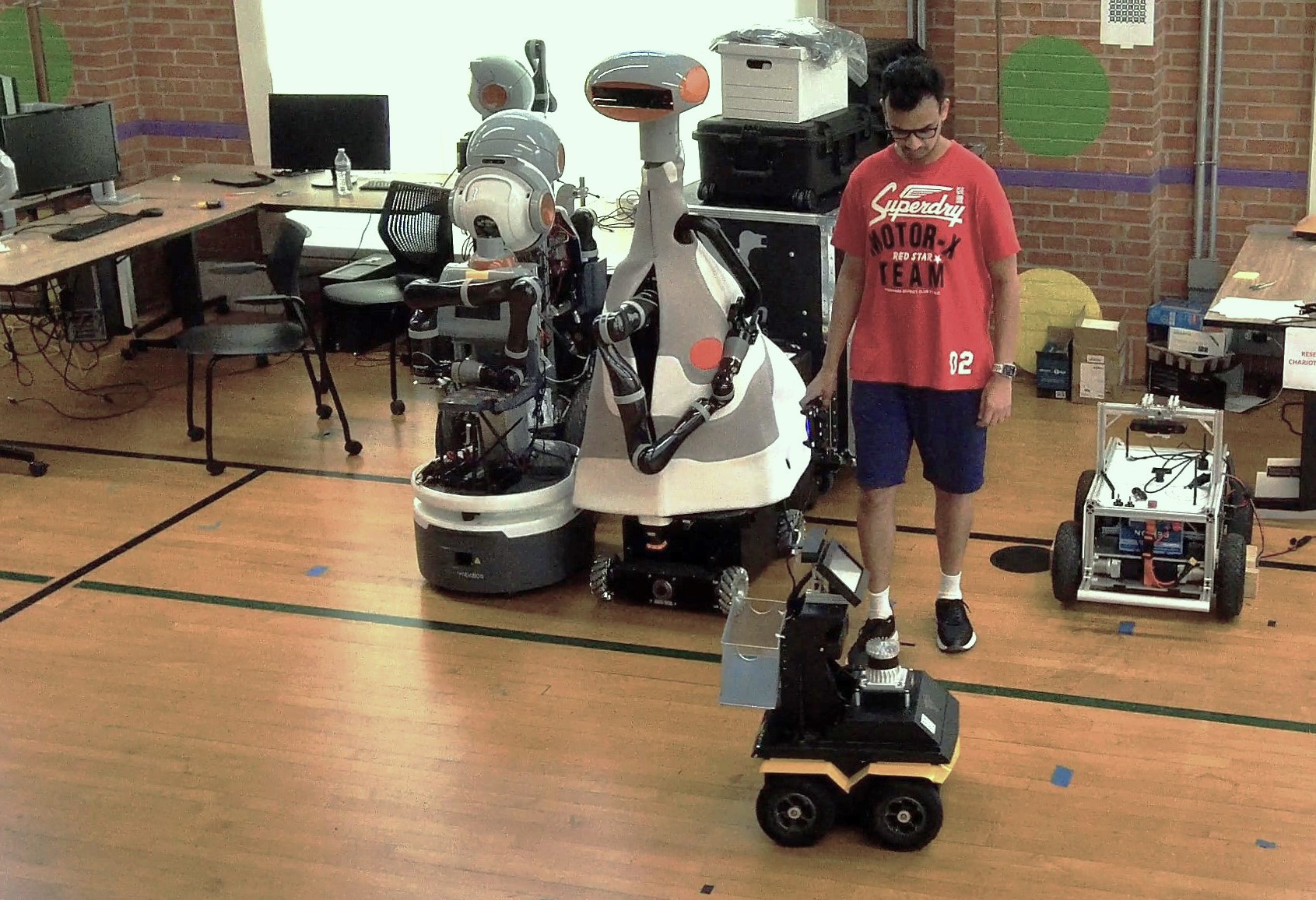}
    \caption{\textit{DWA:} Robot moves once human retreats.}
    \label{fig: pass3i}
  \end{subfigure}
\caption{\textbf{Deployment in human-robot scenarios:} \textit{(Doorway)} Figures~\ref{fig: pass0}, ~\ref{fig: pass1}, and~\ref{fig: pass2} demonstrate our deadlock avoiding strategy enables the robot to yield to the human by slowing down, rather than stopping, and smoothly follows the human through the door. Figures~\ref{fig: fail0}, ~\ref{fig: fail1}, and~\ref{fig: fail2} demonstrates a baseline DWA planner that results in the robot colliding with the human. \textit{(Intersection)} Figures~\ref{fig: pass1i}, ~\ref{fig: pass2i}, and~\ref{fig: pass3i} demonstrate our game-theoretic deadlock avoiding strategy enables the robot to proactively speed up when it notices the human slow down. Figures~\ref{fig: fail0}, ~\ref{fig: fail1}, and~\ref{fig: fail2} demonstrates a baseline DWA planner that results in the robot stopping abruptly in front of the human, only moving forward once the human steps back.} 
  \label{fig: realworld}
  % \vspace{-15pt}
\end{figure*}  

\noindent\textbf{Multi-Robot Setting}—In our evaluation against the \texttt{move\_base} DWA algorithm, a trial is deemed a success if robots navigate through the doorway or intersection without encountering collisions or deadlocks, while any instance of these events marks a failure. As illustrated in Figure~\ref{fig: qualitative}, the initial three rows capture the successful performance of our algorithm across intersection and doorway scenarios, employing the F$1/10$ car platforms, the Jackal, and the Spot. Conversely, the last row reveals a failure case where DWA experiences a collision in the doorway scenario, thereby serving as a contrasting backdrop to the successful outcomes exhibited in the second row using our algorithm in identical configurations.

Crucially, our experimental findings reveal that our proposed navigation framework exhibits a level of efficiency strikingly similar to human navigation. Grounded in research that has studied human navigational speeds and makespan durations in bottleneck scenarios and narrow doorways~\cite{garcimartin2016flow,kretz2006experimental}, the average human specific flow rate\footnote{specific flow rate is measured in $\frac{N}{zT}$, where $N$ is the number of robots, $T$ is the makespan in seconds, and $z$ is the gap width in meters.} has been observed to range between $2.0-2.1$ (m$\cdot$s)$^{-1}$ in environments analogous to our test setups. In comparative terms, our navigation algorithm enabled physical robots to traverse doorways with an average specific flow rate of $2.0$ (m$\cdot$s)$^{-1}$, essentially mirroring the navigational efficiency commonly exhibited by humans. This result amplifies the significance of the specific flow rate metric, highlighting its utility in benchmarking the collective efficiency of robotic navigation against human standards.
\begin{table*}[t]
    \centering
    \resizebox{\linewidth}{!}{
    \begin{tabular}{crcccccc}
         \toprule
          &Baseline& Label & CR(\%) & DR(\%) & Avg. $\Delta$V &Path Deviation&Makespan Ratio  \\
         \midrule
         \multirow{11}{*}{\rotatebox{90}{DOORWAY}}&Wang et al.~\cite{wang2017safety}&D/P&0&0&$0.380 \pm 0.12$&$1.874\pm 0.28$&$2.680 \pm 1.00$ \\
         &ORCA-MAPF~\cite{orcamapf}&C/F&0&0&$0.100 \pm 0.00$&$0.000\pm 0.00$&$3.040 \pm 0.00$ \\
        &Auction-based~\cite{wang2017safety}&D/P&0&0&$0.220 \pm 0.01$& $0.145\pm 0.00$&$3.370 \pm 0.01$ \\
        &IMPC-DR~\cite{impc}&D/F&0&0&$0.080 \pm 0.00$& $0.160\pm 0.00$&$1.530 \pm 0.00$ \\
        &CADRL~\cite{sacadrl}&CTDE/P&50&0&$0.036\pm 0.00$& $1.000\pm 0.00$&$3.500 \pm 0.00$ \\
      &GBPPlanner~\cite{gbp}&D/F&0&0&$0.015\pm 0.00$& $2.670\pm 0.00$&$1.250\pm 0.00$ \\
    &MPC + liveness ($\gamma = 1$)&D/P&$0$&$0$&$0.270\pm0.01$&$0.912\pm0.00$&$1.274\pm0.01$ \\
        &NH-ORCA~\cite{nh-orca}&D/P&0&100&-& -&- \\
      &NH-TTC~\cite{davis2019nh}&D/F&0&100&-& -&- \\
      &MPC-CBF~\cite{zeng2021safety_cbf_mpc}&D/P&0&100&-&-&- \\    \rowcolor{gray!25}&\textbf{Ours}&\textbf{D/P}&0&0&$\mathbf{0.001 \pm 0.00}$& $\mathbf{0.089\pm 0.02}$&$\mathbf{1.005 \pm 0.00}$ \\
        % &No Local&&&& \\
        \cmidrule{2-8}
       \multirow{11}{*}{\rotatebox{90}{INTERSECTION}}&Wang et al.~\cite{wang2017safety}&D/P&0&0&$0.300 \pm 0.10$&$0.400\pm 0.14$&$1.900 \pm 0.04$ \\
       &ORCA-MAPF~\cite{orcamapf}&C/F&0&0&$0.250 \pm 0.00$&$0.000\pm 0.00$&$2.220 \pm 0.00$ \\
        &Auction-based~\cite{wang2017safety}&D/P&0&0&$0.290 \pm 0.05$&$0.111 \pm 0.04$&$2.240 \pm 0.04$ \\ 
    &IMPC-DR~\cite{impc}&D/F&0&0&$0.080 \pm 0.00$&$0.151\pm 0.00$&$1.130 \pm 0.00$ \\        
   &CADRL~\cite{sacadrl}&CTDE/P&0&0&$0.031\pm 0.00$& $1.220\pm 0.00$&$2.000\pm 0.00$ \\
      &GBPPlanner~\cite{gbp}&D/F&0&0&$0.017\pm 0.00$& $7.520\pm 0.00$&$0.710\pm 0.00$ \\
    &MPC + liveness ($\gamma = 1$)&D/P&$0$&$0$&$0.340\pm0.01$&$0.922\pm0.00$&$1.298\pm0.01$ \\
        &NH-ORCA~\cite{nh-orca}&D/P&0&100&-& -&- \\
      &NH-TTC~\cite{davis2019nh}&D/F&0&100&-& -&- \\
      &MPC-CBF~\cite{zeng2021safety_cbf_mpc}&D/P&0&100&-&-&- \\
\rowcolor{gray!25}&\textbf{Ours}&\textbf{D/P}&0&0&$\mathbf{0.002 \pm 0.00}$&$\mathbf{0.066 \pm   0.01}$&$\mathbf{1.005 \pm 0.01}$ \\
        \cmidrule{2-8}
       \multirow{11}{*}{\rotatebox{90}{HALLWAY}}&Wang et al.~\cite{wang2017safety}&D/P&0&0&$0.055 \pm 0.01$&$0.327\pm 0.32$&$1.160\pm 0.04$ \\
       &ORCA-MAPF~\cite{orcamapf}&C/F&0&0&$0.110 \pm 0.00$&$1.990\pm 0.00$&$1.030 \pm 0.00$ \\
        &Auction-based~\cite{wang2017safety}&D/P&0&0&0$.008 \pm 0.000$&$0.190\pm 0.00$&$1.440 \pm 0.04$ \\
        &IMPC-DR~\cite{impc}&D/F&0&0&$0.135 \pm 0.00$&$0.194\pm 0.00$&$2.080 \pm 0.00$ \\
           &CADRL~\cite{sacadrl}&CTDE/P&0&0&$0.047\pm 0.00$& $1.000\pm 0.00$&$1.000\pm 0.00$ \\
          &GBPPlanner~\cite{gbp}&D/F&0&0&$0.018	\pm 0.00$& $1.590\pm 0.00$&$1.000\pm 0.00$ \\
      &MPC + liveness ($\gamma = 1$)&D/P&$0$&$0$&$0.290\pm0.01$&$0.824\pm0.00$&$1.315\pm0.01$\\
        &NH-ORCA~\cite{nh-orca}&D/P&0&100&-& -&- \\
      &NH-TTC~\cite{davis2019nh}&D/F&0&100&-& -&- \\
      &MPC-CBF~\cite{zeng2021safety_cbf_mpc}&D/P&0&100&-&-&- \\
        \rowcolor{gray!25}&\textbf{Ours}&\textbf{D/P}&0&0&$\mathbf{0.001 \pm 0.00}$&$\mathbf{0.047\pm 0.00}$&$\mathbf{1.005 \pm 0.00}$ \\
        %\cmidrule{2-6}
       %\multirow{2}{*}{\rotatebox{0}{ROUND..}}&Random CBF~\cite{wang2017safety}&$0.69 \pm 0.13$&$1.597$&$1.69 \pm 0.12$& \\
        %&Game-theoretic&$1.24 \pm 0.01$&$1.610 \pm 0.00$&$1.12 %\pm 0.01$&\\        
         \bottomrule
    \end{tabular}
    }
    \caption{We observe that alternate perturbation strategies are less smooth due to larger average changes in velocity and path deviation compared to game-theoretic perturbation. The \textit{Label} column describes the amount of centralization and information required--Decentralized (D), Centralized (C), Full Information (F) or Partial Information (P).}
    \label{tab: perturb_comparison}
    % \vspace{-15pt}
\end{table*}

% \input{images/CBF+Perturb_VS_CBF}
% \subsection{Comparing with Naive Multi-agent Planning on Real Robots}
% \begin{figure}[t]
% \centering
%    \begin{subfigure}[h]{0.485\columnwidth}
%     \includegraphics[width=\textwidth]{images/PySim/nhttc_1.jpg}
%     \caption{Initial time step}
%     \label{fig: nhttc1}
%   \end{subfigure}
%   %
%  \begin{subfigure}[h]{0.485\columnwidth}
%     \includegraphics[width=\textwidth]{images/PySim/nhttc_2.jpg}
%     \caption{Both agents enter a deadlock.}
%     \label{fig: nhttc2}
%   \end{subfigure}
% %
%  \begin{subfigure}[h]{0.485\columnwidth}
%     \includegraphics[width=\textwidth]{images/RVO/ORCA_initial.png}
%     \caption{Initial time step}
%     \label{fig: orca1}
%   \end{subfigure}
%   %
%  \begin{subfigure}[h]{0.485\columnwidth}
%     \includegraphics[width=\textwidth]{images/RVO/ORCA.png}
%     \caption{Both agents enter a deadlock.}
%     \label{fig: orca2}
%   \end{subfigure}
% \caption{\textbf{Demonstrating baselines such as NH-TTC~\cite{davis2019nh} and NH-ORCA~\cite{nh-orca}} in a doorway scenario: In both cases, agents result in a deadlock due to infeasible control sets.} 
%   \label{fig: nhorca_nhttc}
%   % \vspace{-15pt}
% \end{figure} 
\begin{figure*}[t]
    \centering
    \begin{subfigure}[h]{\columnwidth}
\includegraphics[width = \linewidth]{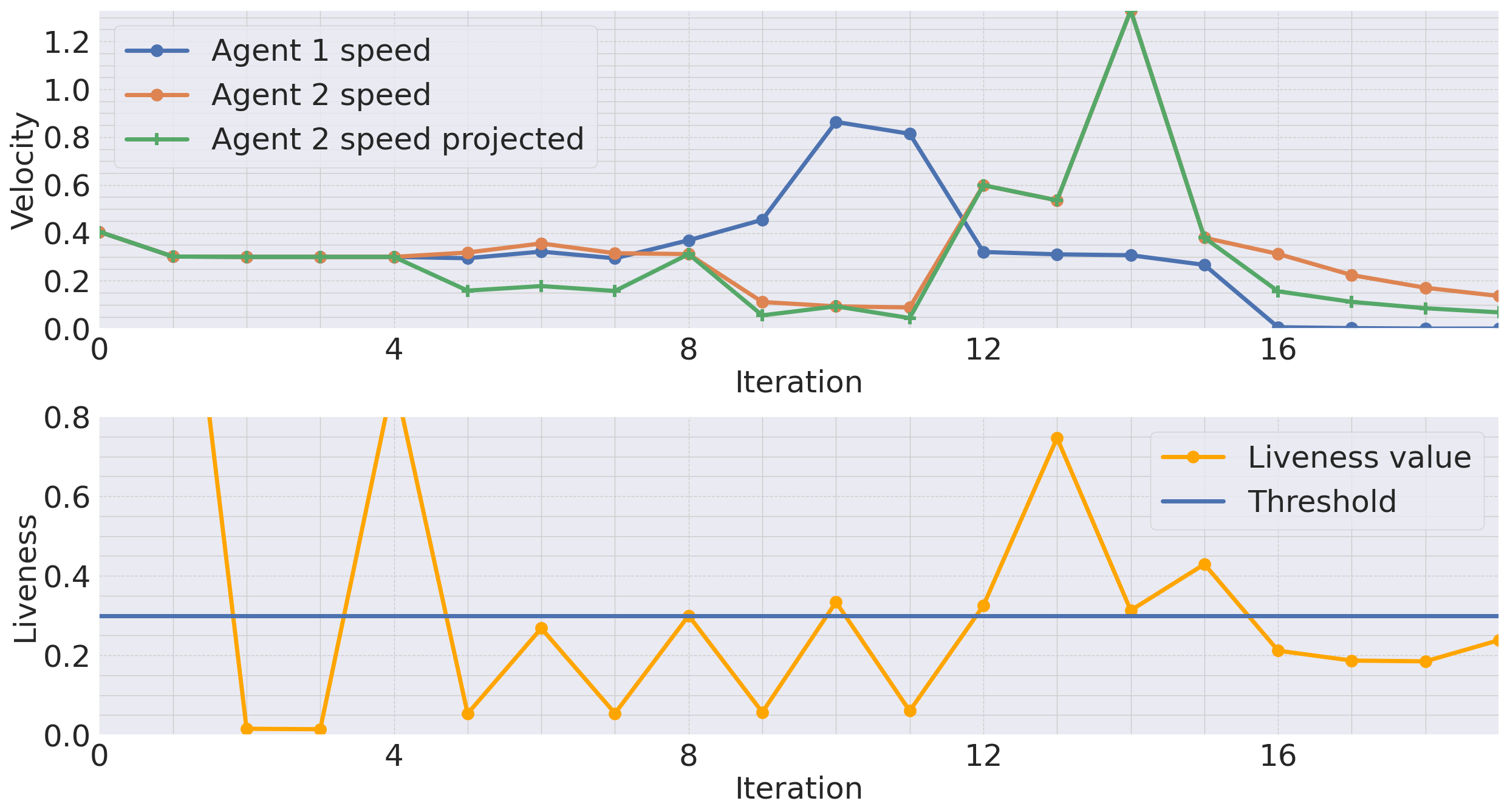}    
    \caption{\textit{(deadlock prevented)} $d = 1.11m,  \Delta x_g = 0$}
    \label{fig: vel1}
    \end{subfigure}
    \begin{subfigure}[h]{\columnwidth}
\includegraphics[width = \linewidth]{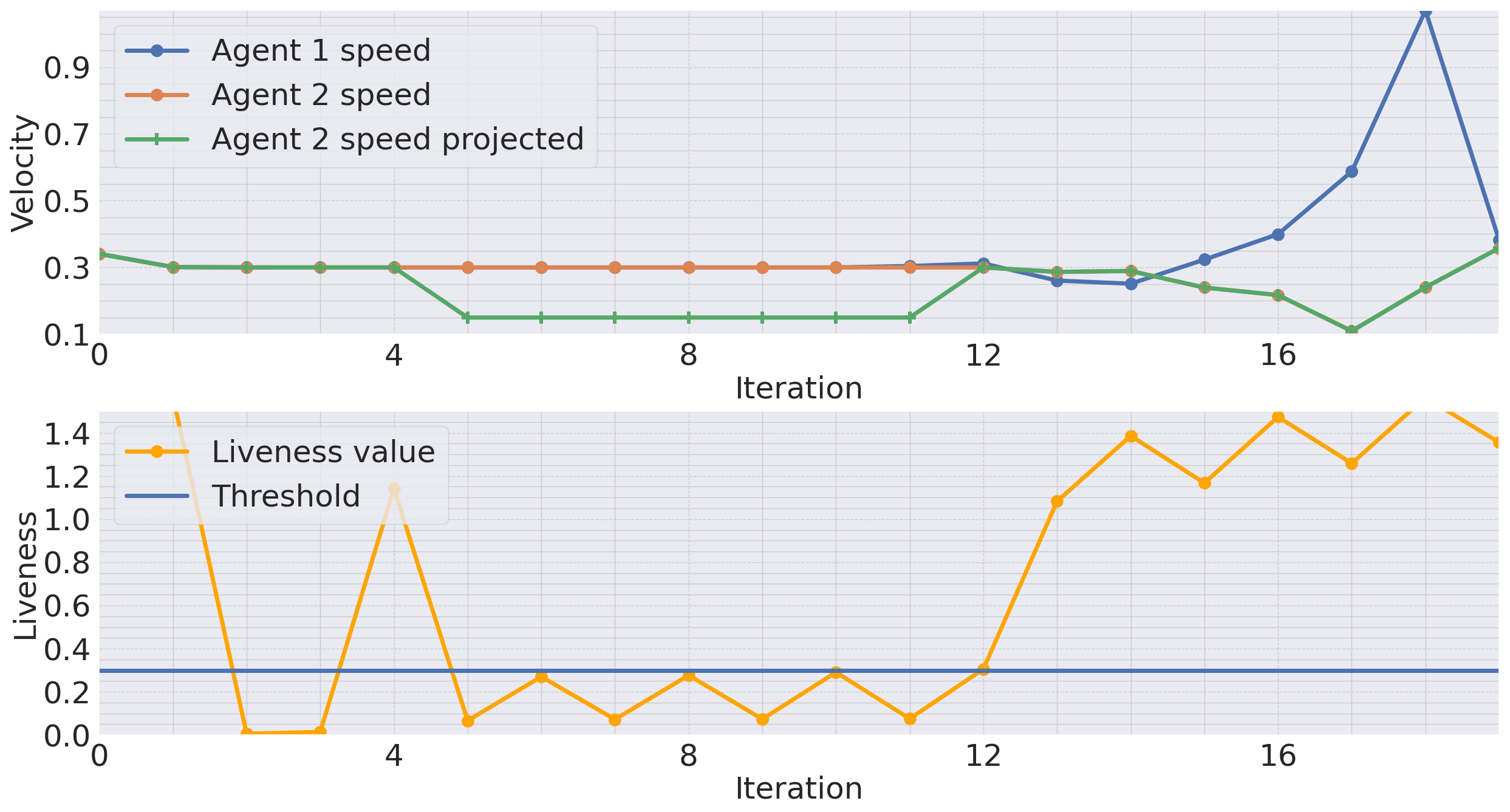}    
    \caption{\textit{(deadlock prevented)} $d = 2.55m, \Delta x_g = 0.2m$}
    \label{fig: vel2}
    \end{subfigure}
    \begin{subfigure}[h]{\columnwidth}
\includegraphics[width = \linewidth]{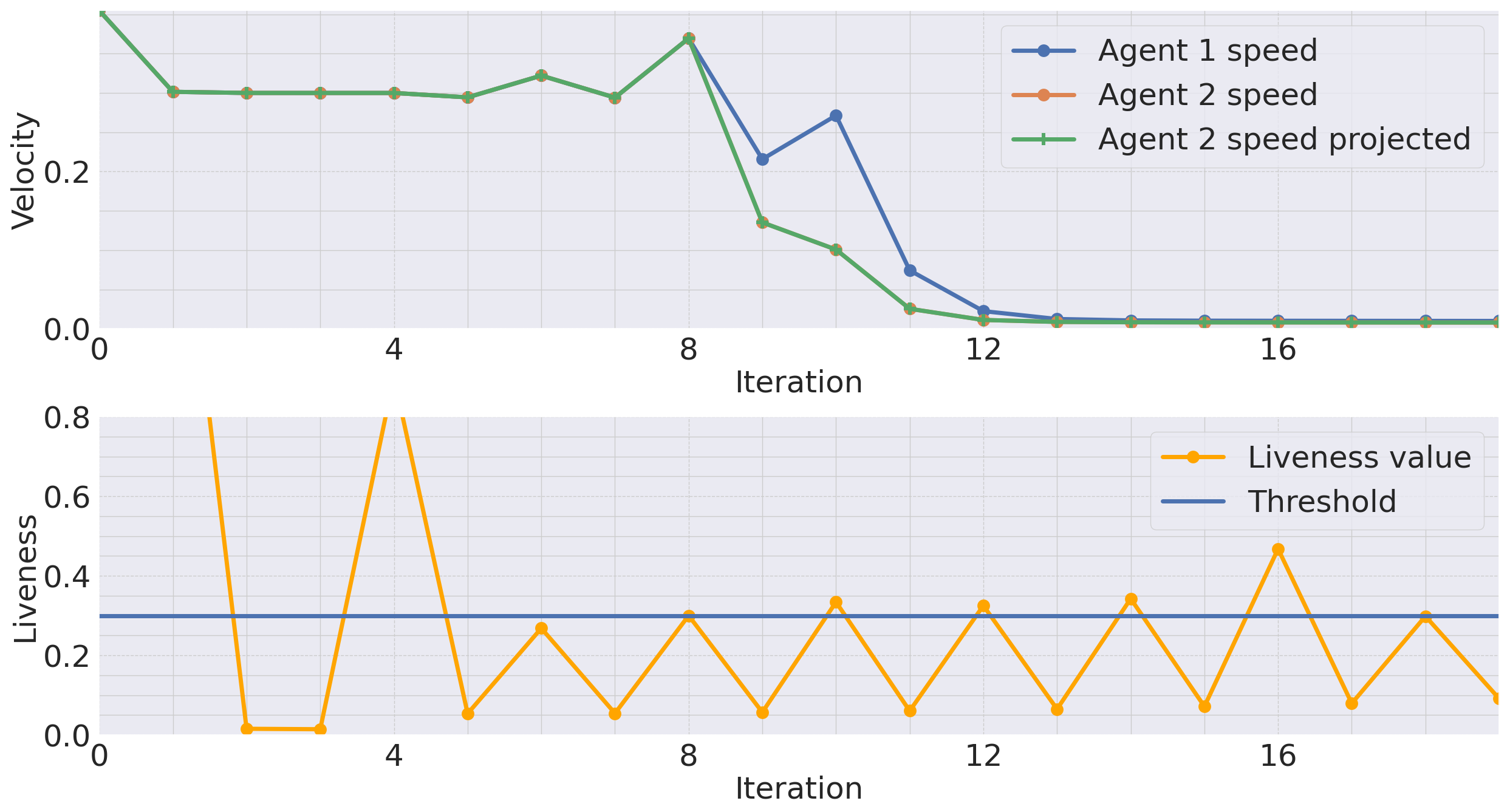}    
    \caption{\textit{(deadlocked)} $d = 1.11m,  \Delta x_g = 0$}
    \label{fig: vel3}
    \end{subfigure}
    \begin{subfigure}[h]{\columnwidth}
\includegraphics[width = \linewidth]{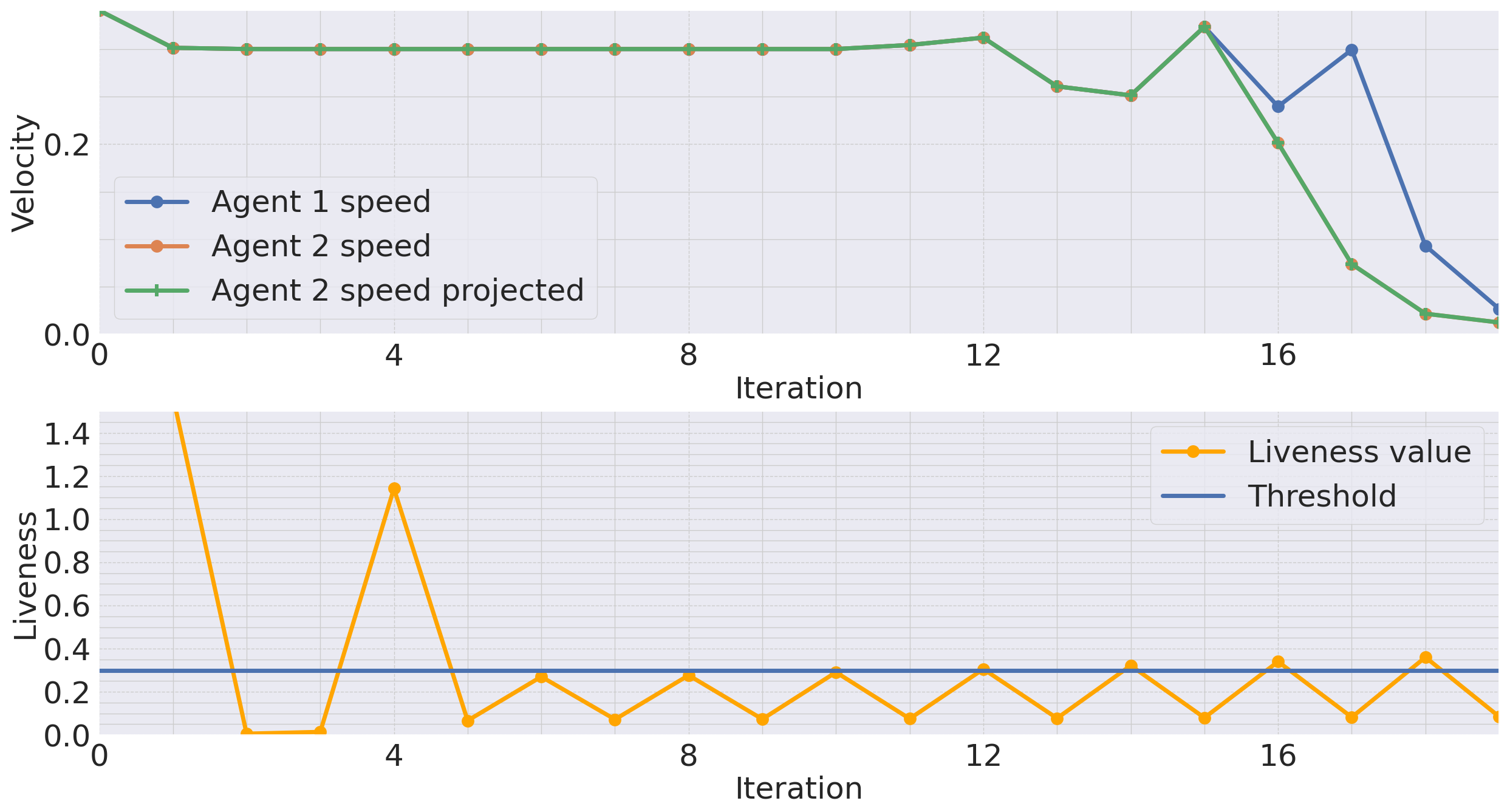}    
    \caption{\textit{(deadlocked)} $d = 2.55m, \Delta x_g = 0.2m$}
    \label{fig: vel4}
    \end{subfigure}
    \caption{Deadlock is detected around $t=4$. Agent $2$ projects its \textcolor{orange}{original velocity} to a \textcolor{ForestGreen}{scaled velocity} such that differs from agent $1$'s \textcolor{blue}{velocity} by at least a factor of $\zeta$. Deadlock is prevented between $t=5$ and $t=10$ as confirmed from $\ell_j\left( p^i_t, v^i_t\right) \geq \ell_\textnormal{thresh} = 0.3$.}
    \label{fig: vel}
    % \vspace{-10pt}
\end{figure*}

\begin{figure*}[t]
\centering
   \begin{subfigure}[h]{0.24\textwidth}
    \includegraphics[width=\textwidth]{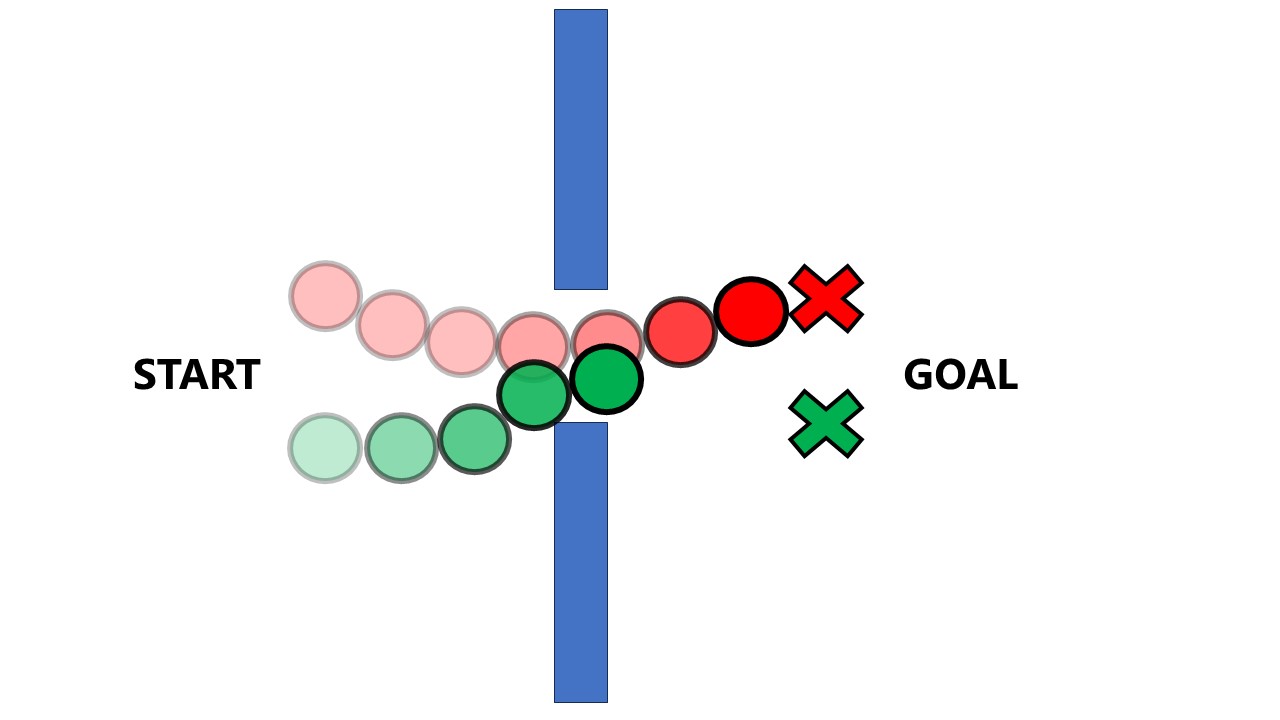}
    \caption{MPC-CBF w. liveness}
    \label{fig: game1}
  \end{subfigure}
 \begin{subfigure}[h]{0.24\textwidth}
    \includegraphics[width=\textwidth]{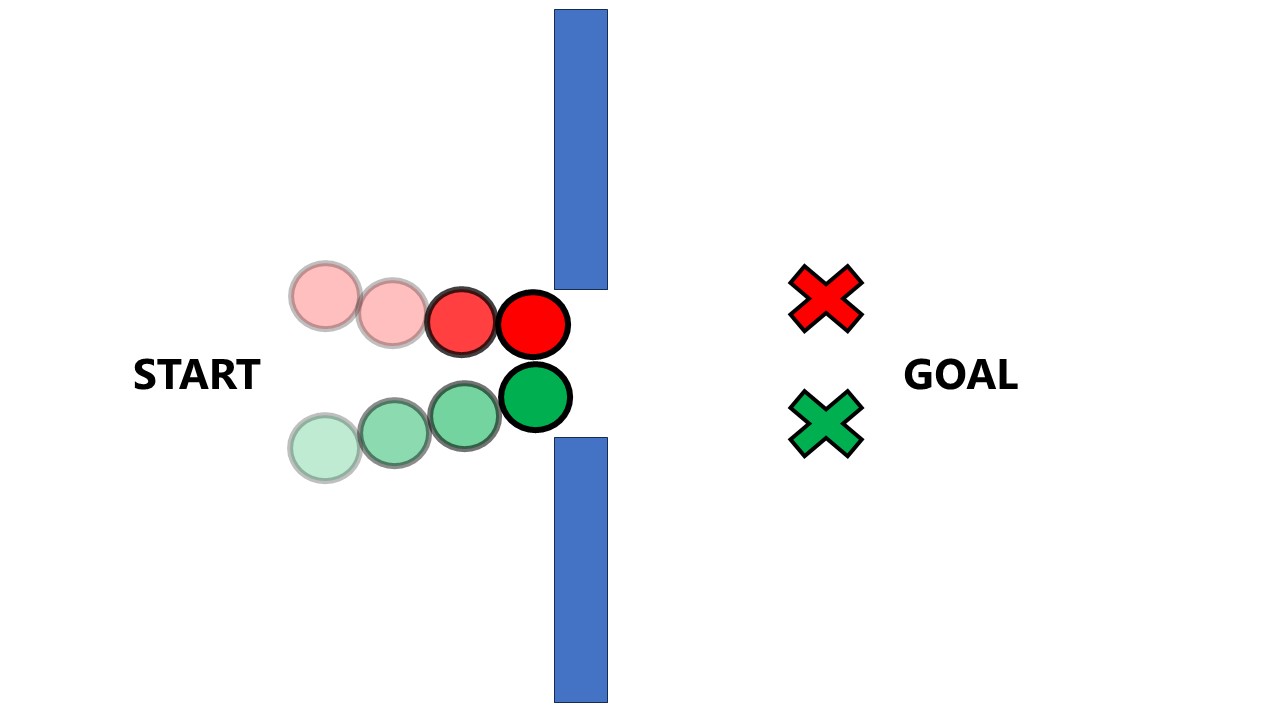}
    \caption{MPC-CBF\cite{zeng2021safety_cbf_mpc}}
    \label{fig: game2}
  \end{subfigure}
\begin{subfigure}[h]{0.24\textwidth}
    \includegraphics[width=\textwidth]{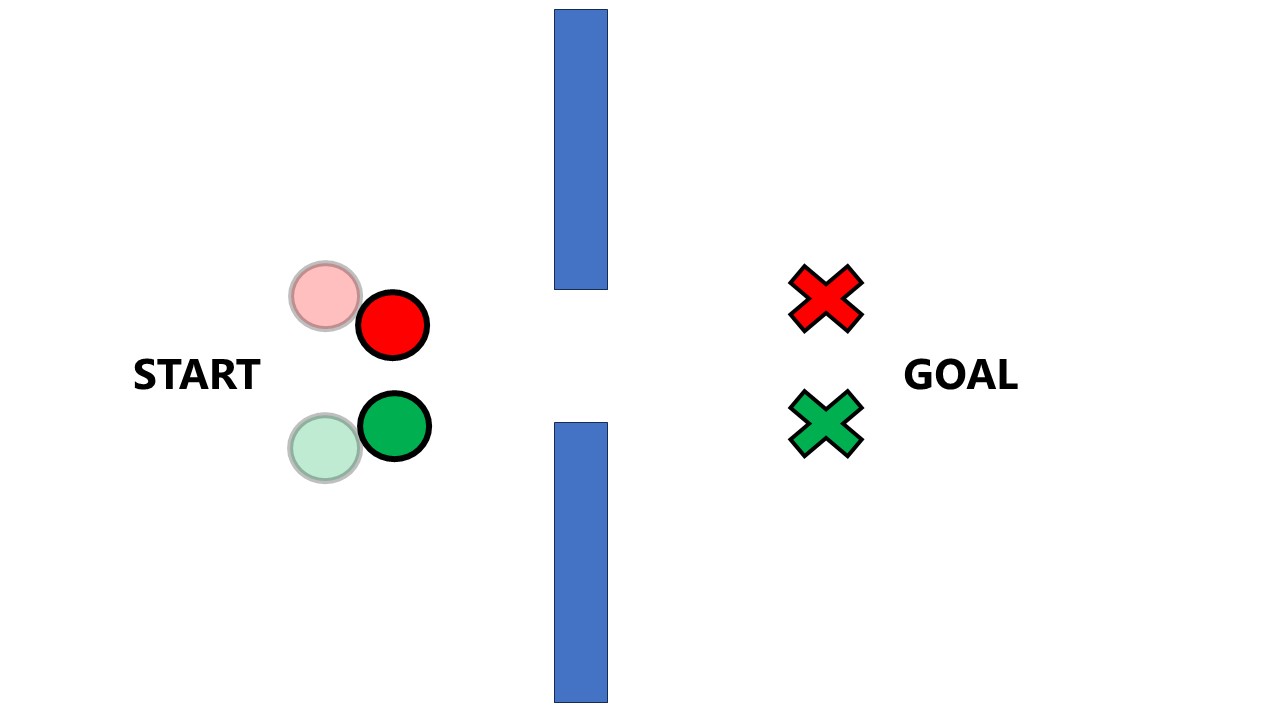}
    \caption{NH-ORCA~\cite{nh-orca}}
    \label{fig: game3}
  \end{subfigure}
   \begin{subfigure}[h]{0.24\textwidth}
    \includegraphics[width=\textwidth]{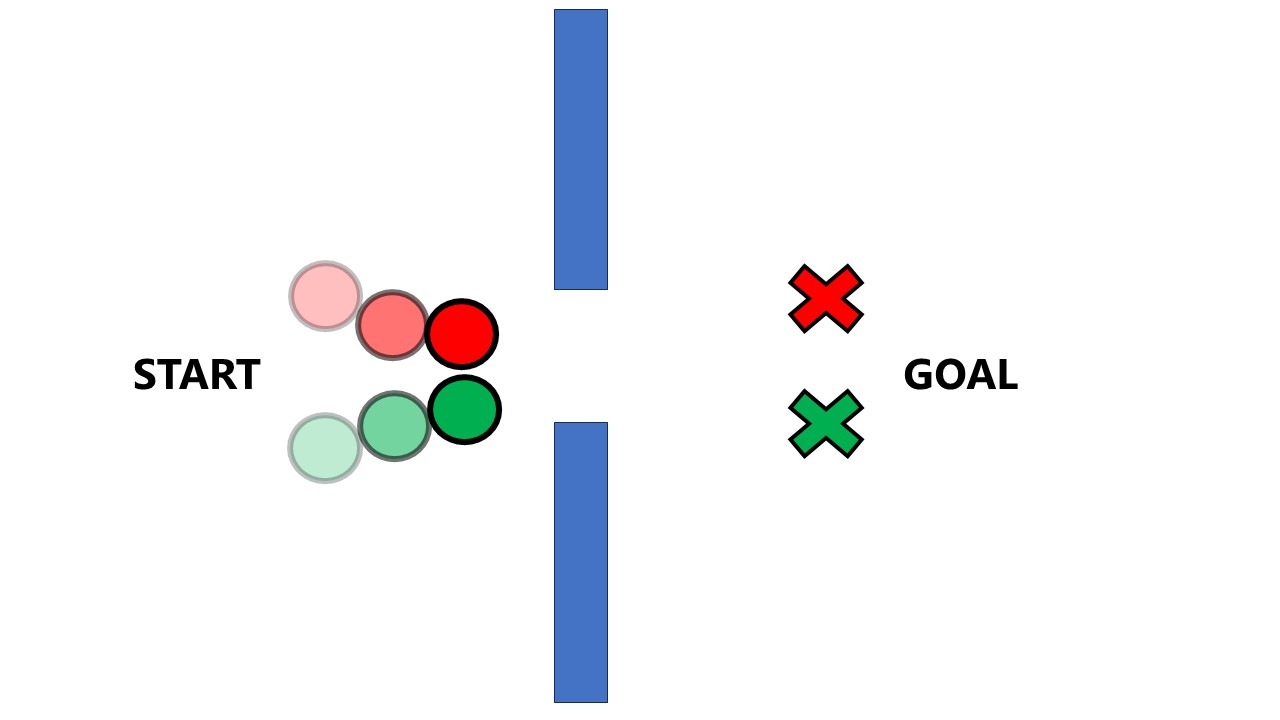}
    \caption{NH-TTC~\cite{davis2019nh}}
    \label{fig: game4}
  \end{subfigure}
     \begin{subfigure}[h]{0.24\textwidth}
    \includegraphics[width=\textwidth]{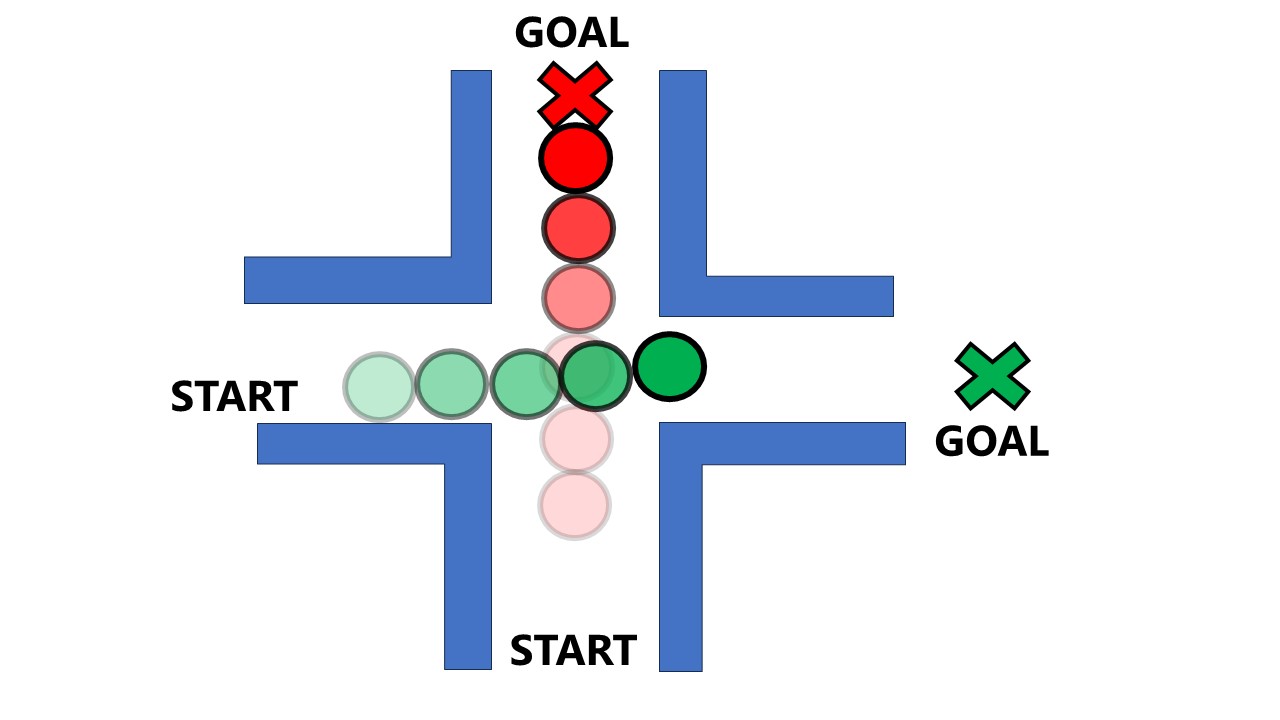}
    \caption{MPC-CBF w. liveness}
    \label{fig: nongame1}
  \end{subfigure}
 \begin{subfigure}[h]{0.24\textwidth}
    \includegraphics[width=\textwidth]{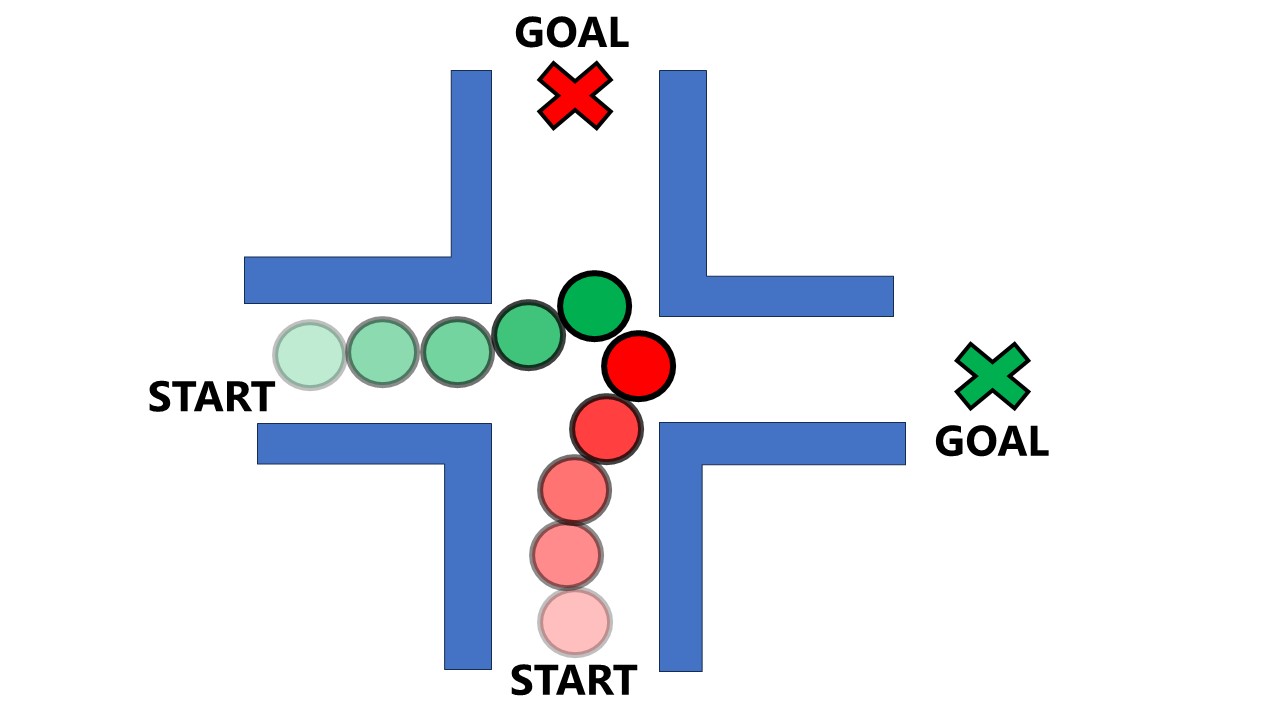}
    \caption{MPC-CBF\cite{zeng2021safety_cbf_mpc}}
    \label{fig: nongame2}
  \end{subfigure}
\begin{subfigure}[h]{0.24\textwidth}
    \includegraphics[width=\textwidth]{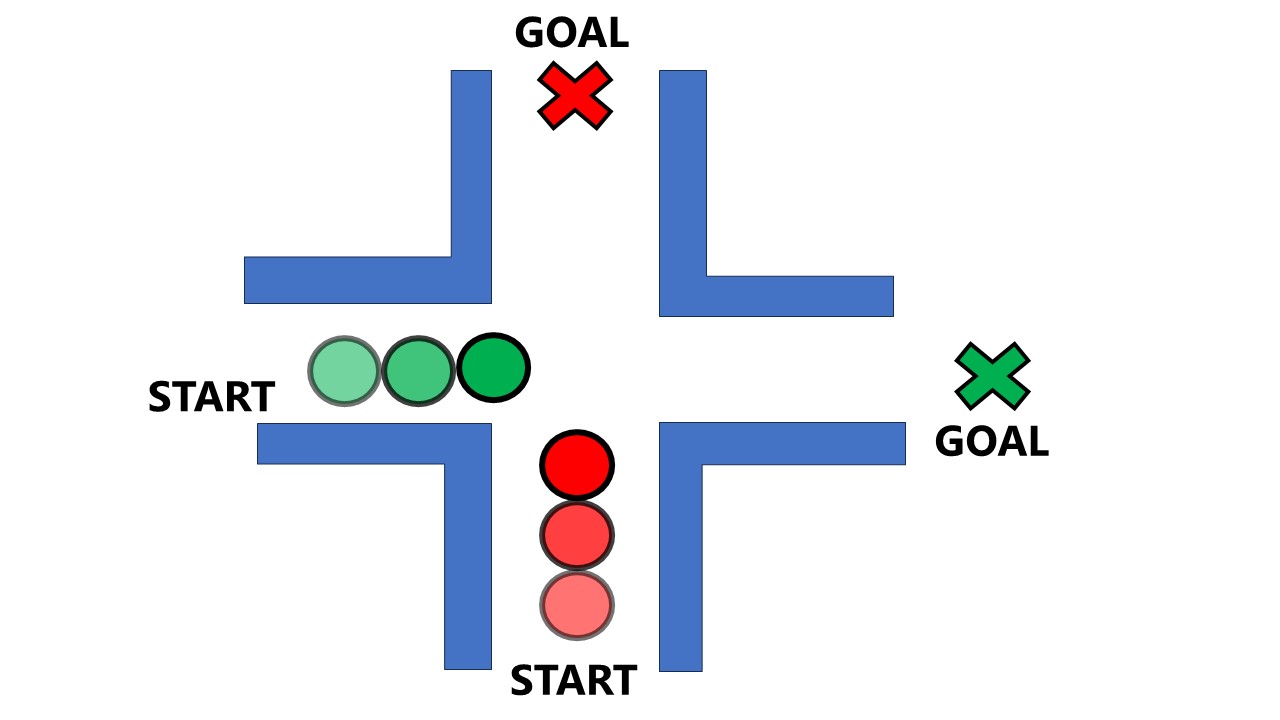}
    \caption{NH-ORCA~\cite{nh-orca}}
    \label{fig: nongame3}
  \end{subfigure}
   \begin{subfigure}[h]{0.24\textwidth}
    \includegraphics[width=\textwidth]{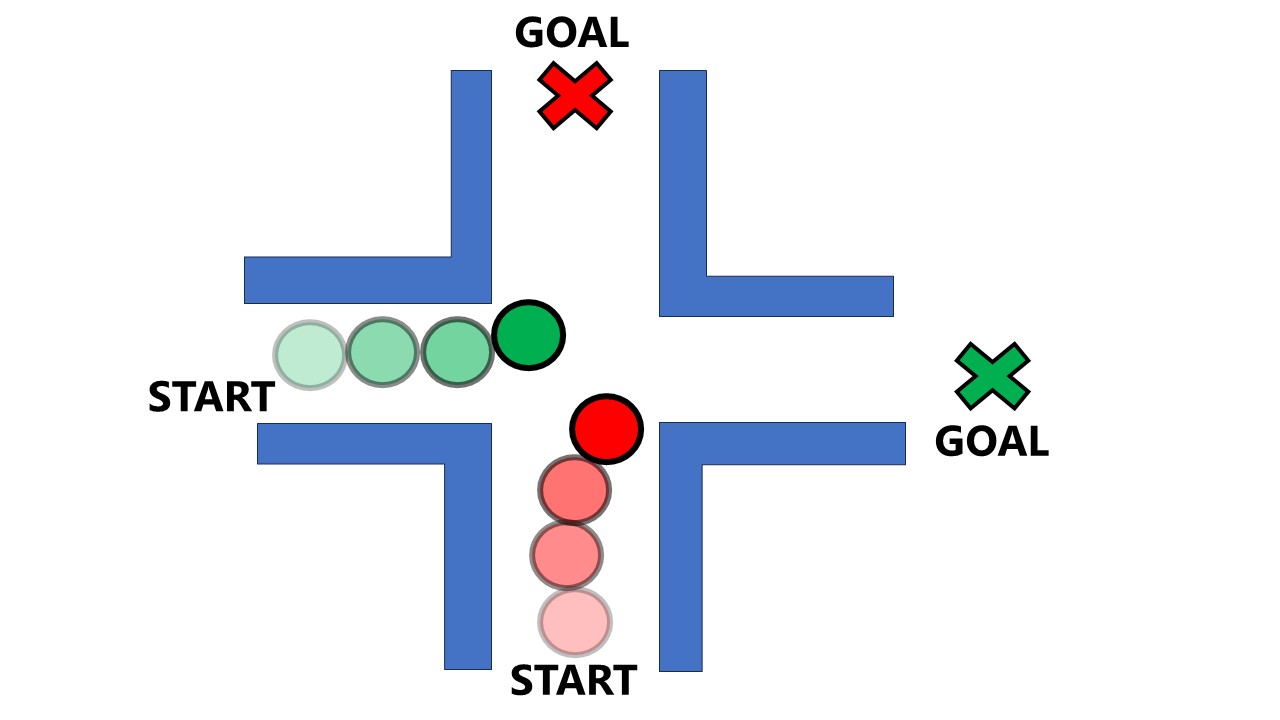}
    \caption{NH-TTC~\cite{davis2019nh}}
    \label{fig: nongame4}
  \end{subfigure}
\caption{\textbf{Comparing with methods \underline{without} deadlock resolution capabilities}--Our approach, control-theoretic MPC-CBF enables the green agent to yield to the red agent, which is in contrast to the baselines where the agents enter a deadlock. Similar observation in the intersection scenario. \textit{Conclusion:} Our control-theoretic controller presented in Section~\ref{sec: GT-resolution} encourages queue formation thereby resolving a deadlock situation in a decentralized and smooth manner.} 
  \label{fig: PySim}
  % \vspace{-10pt}
\end{figure*}  

\begin{figure*}[t]
    \centering
    \includegraphics[width=\textwidth]{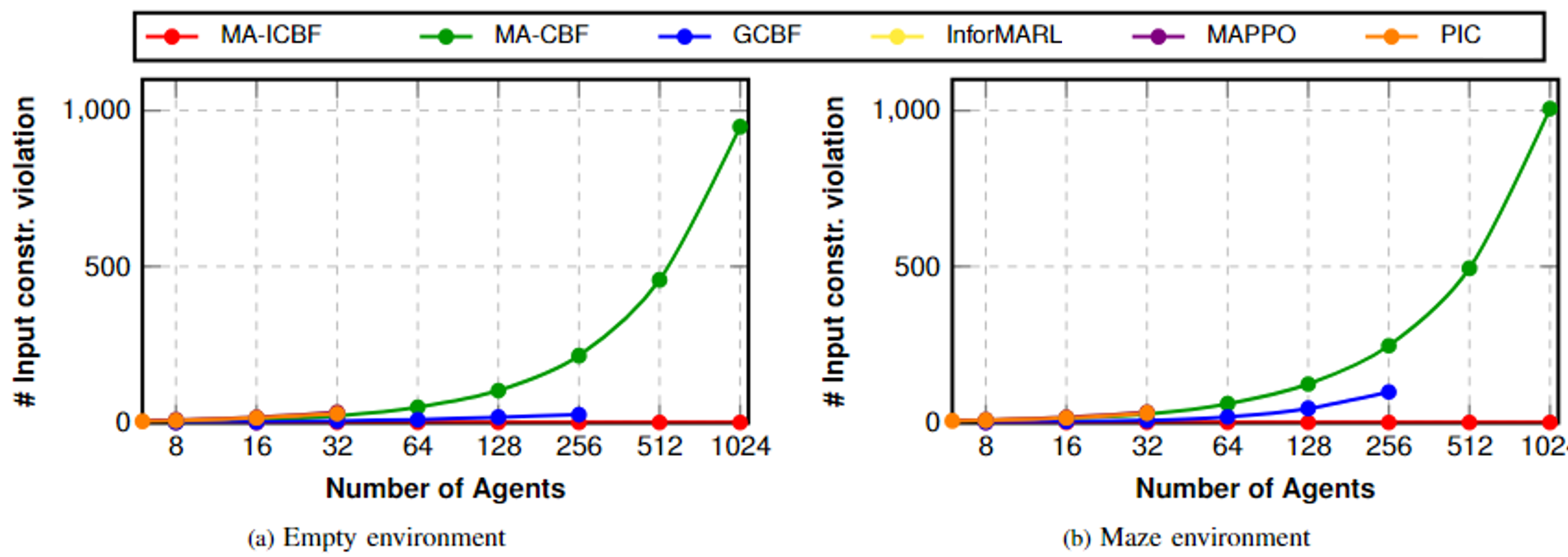}
    \caption{Average number of input constraint violations versus the number of agents in different settings.}
    \label{fig: input}
    % \vspace{-10pt}
\end{figure*}

\begin{figure*}[t]
    \centering
    \begin{subfigure}[t]{0.49\linewidth}
        \centering
        \includegraphics[width=\linewidth]{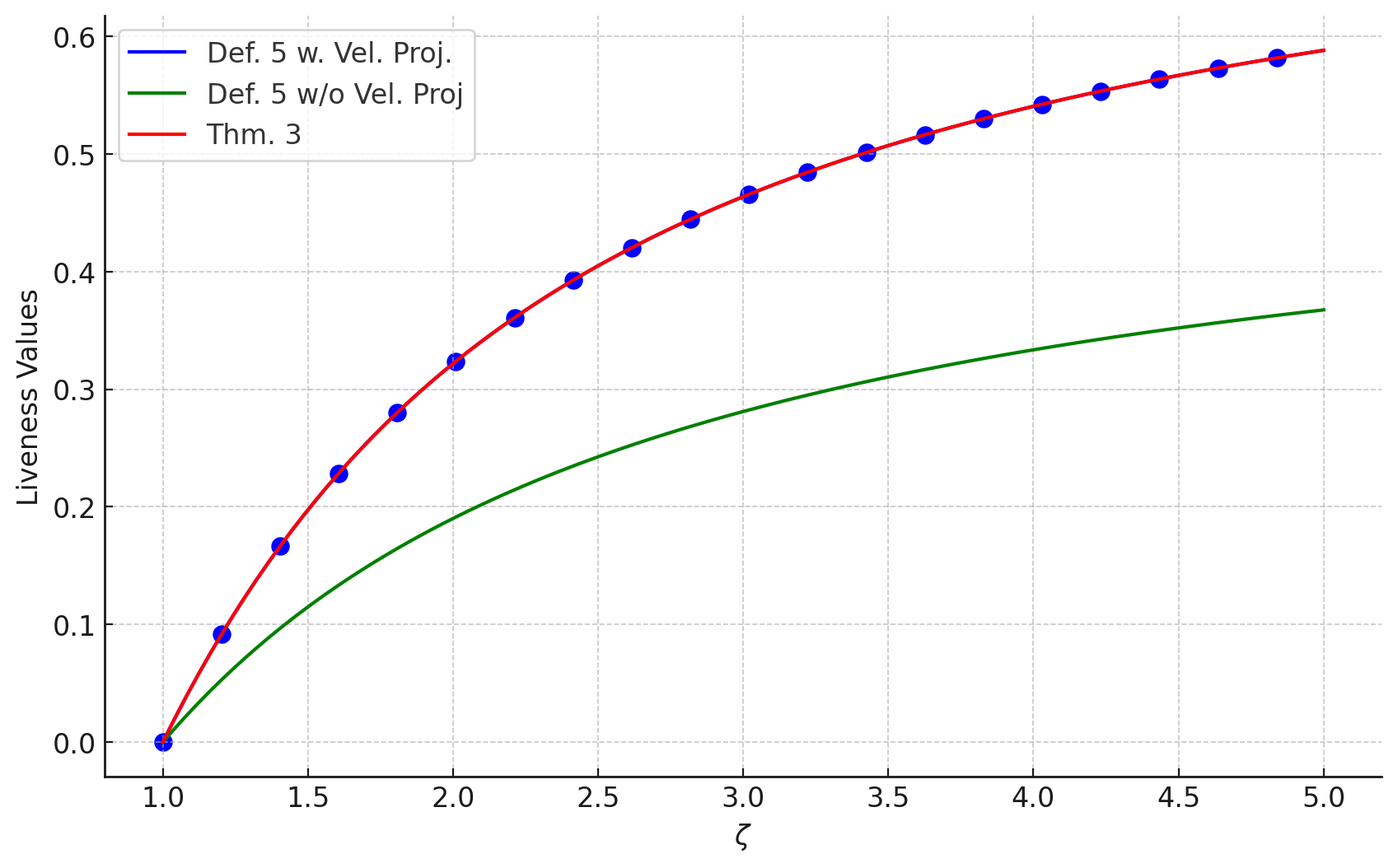}
        \caption{For $\theta = \frac{\pi}{6}$: Showing that Definition~\ref{def: liveness_function} with velocity projection reconciles with Theorem~\ref{thm: smg->l<tau}.}
        \label{fig: reconcile_30}
    \end{subfigure}
    \hfill
    \begin{subfigure}[t]{0.49\linewidth}
        \centering
        \includegraphics[width=\linewidth]{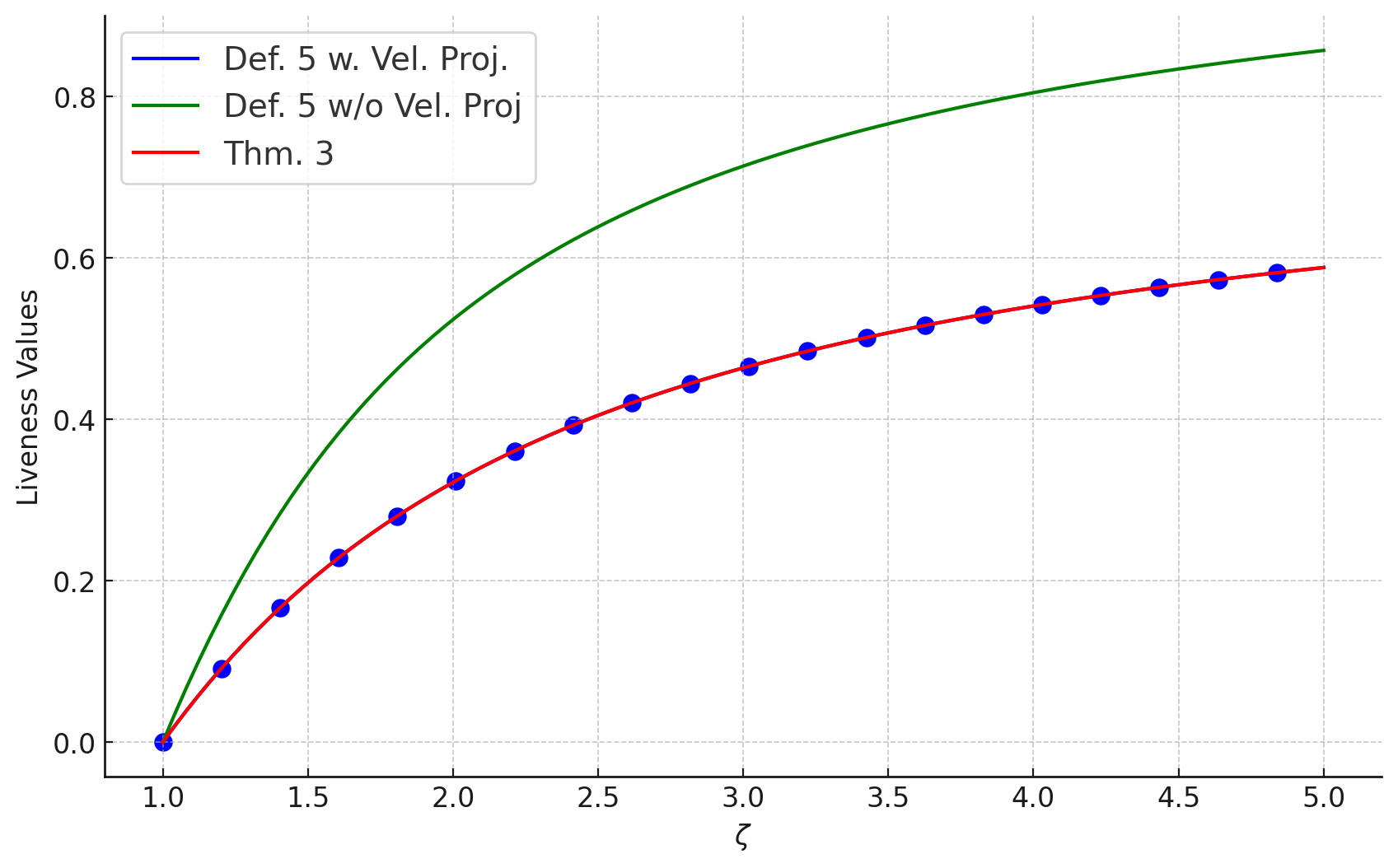}
        \caption{For $\theta = \frac{\pi}{3}$: Showing that Definition~\ref{def: liveness_function} with velocity projection reconciles with Theorem~\ref{thm: smg->l<tau}.}
        \label{fig: reconcile_60}
    \end{subfigure}
    \caption{Comparison of Definition~\ref{def: liveness_function} with velocity projection with Theorem~\ref{thm: smg->l<tau} for different values of $\theta$.}
    \label{fig: reconcile_combined}
\end{figure*}

\noindent\textbf{Human-Robot Setting}—In an extension of our work to human-robot interactions, we recreate the doorway and intersection social mini-game scenarios\footnote{We suggest the reader to watch the full video at \href{https://www.youtube.com/watch?v=fA7BbM8iTwg}{https://www.youtube.com/watch?v=fA7BbM8iTwg}.}, as depicted in Figure~\ref{fig: realworld}. Here, we compare our algorithm against the traditional Dynamic Window Approach (DWA), introducing a human participant into the equation for a more nuanced assessment. Figures~\ref{fig: pass0},~\ref{fig: pass1}, and~\ref{fig: pass2}, corresponding to rows $1$ and $2$ for the doorway situation, eloquently highlight how our control-theoretic, deadlock-avoidance strategy equips the robot with the finesse to decelerate strategically. This allows the robot to defer to the human participant seamlessly, thereby maintaining an uninterrupted flow of motion. This stands in stark contrast to the results achieved by DWA, where the robot ends up making physical contact with the human, as illustrated in Figures~\ref{fig: fail0},~\ref{fig: fail1}, and~\ref{fig: fail2}.

Of particular significance is our algorithm's capability to adapt to varied human behaviors. As Figures~\ref{fig: pass1i} and~\ref{fig: pass2i} demonstrate, our robot is not merely programmed to yield; it can also dynamically accelerate when the situation warrants, particularly when human participants display hesitancy or a conservative pace. This preemptive speeding up averts potential deadlocks that could otherwise ensue. Conversely, DWA lacks the agility for such assertive, situationally-responsive maneuvers, often resulting in a halted state and thereby creating a deadlock scenario.

\subsection{Simulation Results}
We implement the MPC-CBF controller using double-integrator system dynamics in Equation~\eqref{eq: control_affine_dynamics}, cost function in Equation~\eqref{eq: SNUPI_safety_constrained_velocity_scaling_v}, and constraints in Equation~\eqref{eq: constraints}. All experiments have been performed on a 
AMD Ryzen machine@2.90GHz with 16.0GB of RAM.

\noindent\textbf{Analysis}--Figure~\ref{fig: vel} presents simulation results and analysis of interaction between two agents at a narrow doorway. In the first experiment where distance \( d \) is set to $1.11m$ and the difference in goal position \( \Delta x_g \) is $0$, shown in Figure~\ref{fig: vel1}, a deadlock is detected at $t=5$. This is indicated by the liveness value dipping below the threshold ($0.3$) at this time. At this point, agent $2$ already starts to slow down according to Equation~\ref{eq: perturbation}, as indicated by the green curve. The deadlock occurrence is around $t=10$, which is when agent $2$ slows down even further, while agent 1's speed (blue curve) shows a significant peak, in order to prevent the deadlock. Figure~\ref{fig: vel3} demonstrates the same scenario in the absence of deadlock prevention. In this case,  both agents slow down around $t=10$ and eventually come to a halt. Throughout this period, the liveness value remains below the threshold.

Figures~\ref{fig: vel2} and~\ref{fig: vel4} follow a similar pattern with a larger distance \( d = 2.55m \) and a slight difference in goal position \( \Delta x_g = 0.2m \). The goal with this second experiment is to convey that our perturbation algorithm is invariant to changes in the configuration.

\noindent\textbf{Versus baselines \underline{with} deadlock resolution}--We compare our deadlock resolution method (MPC-CBF) with quadratic programming-based controllers (QP-CBF)~\cite{wang2017safety} using random perturbation as well as a variant where we implement our approach (QP-CBF), infinite-horizon MPC and buffered voronoi cells~\cite{impc}, a multi-agent reinforcement learning-based algorithm, CADRL~\cite{cadrl,sacadrl}, and finally, with an ORCA-based approach~\cite{orcamapf} that uses multi-agent path finding (MAPF) to resolve deadlocks. We perform experiments in social mini-games at doorways, hallways, and intersections, and report the mean $\pm$ standard deviation for average change in velocity ($\Delta V$), path deviation (meters), and average time steps across $2$ agents in Table~\ref{tab: perturb_comparison}.
% $\Delta V$ measures how much an agent needs to modulate their velocity in order to prevent a deadlock. The path deviation measures the difference between the preferred path $\widetilde \Gamma^i$ and executed path, $\Gamma^i$.
For all three settings, deadlock resolution results in the smallest average velocity and path deviation, which implies that our approach allows agents to modulate their velocities more efficiently that results in a trajectory that more closely resembles the preferred trajectory. Note that path deviation for ORCA-MAPF will always be zero due to the discrete nature of the MAPF. Additionally, we observe that the average completion time depends on the social mini-game. For instance, random perturbation takes longer in the doorway setting due to the increased number of constraints in space. In the intersection and hallway settings, however, perturbation is slower or comparable since random perturbation agents move faster, albeit along inefficient trajectories. Lastly, we observe that MPC + liveness results in a very small makespan at the cost of high average $\Delta V$ and path deviation. This suggests that simply adding liveness constraints to the MPC solver encourages goal reaching but at the cost of smoothness.

\noindent\textbf{Versus baselines \underline{without} deadlock resolution}—Our study compares our navigation strategy against other multi-robot navigation algorithms that lack any explicit deadlock resolution mechanisms. This includes a range of methods like MPC-CBF~\cite{zeng2021safety_cbf_mpc}, NH-ORCA~\cite{nh-orca}, NH-TTC~\cite{davis2019nh}, CADRL and its variants~\cite{cadrl, cadrl-lstm, ga3c-cadrl}, and DWA. Of these methods, MPC-CBF, NH-ORCA, and NH-TTC fail to reach their goals, either colliding or ending in a deadlock. We visually demonstrate these comparisons in Figure~\ref{fig: PySim}. With our method, green agents yield to red agents, enabling smooth passage through constricted spaces like doorways or intersections. Conversely, approaches like MPC-CBF, NH-ORCA, and NH-TTC falter, leading to either collisions or unresolved deadlocks. For example, Figures~\ref{fig: game2} and~\ref{fig: nongame2} display a deadlock when using MPC-CBF due to the symmetrical challenges posed by the environment. Likewise, methods that emphasize collision-free navigation, such as NH-ORCA and NH-TTC, encounter deadlocks as displayed in Figures~\ref{fig: game3},~\ref{fig: nongame3},~\ref{fig: game4}, and~\ref{fig: nongame4}. The inherent symmetry of the environment leaves these agents with empty feasible action sets, leading to a deadlock unless specific perturbation strategies are deployed.

\subsection{Dealing with Input Constraints}

We performed a simulation using an empty environment and a complex maze environment which is a $10m \times 10m$ grid. For each experiments 5 trials are conducted with each trial being randomized with respect to start and goal position in both of the environments. All experiments have been conducted on an AMD Ryzen machine @$2.90$GHz with $8$ cores and $16.0$ GB of RAM. The maximum number of agents tested in our experiments is $1024$. We assume each agent's control policy satisfies the input constraints i.e. $\left\lvert u^i_t \right \rvert\leq u_{\text{max}}$ where $u_\text{max}>0$. We set the input constraints in the range of $[-0.2, 0.2]$ meters per second for the linear velocity and $[-12, 12]$ degrees for the angular velocity. Figure~\mbox{\ref{fig: input}} shows that the proposed approach successfully constrains all agents within these bounds. In contrast, baselines GCBF~\mbox{\cite{zhang2023neural_gcbf}}, MACBF~\mbox{\cite{qin2021learning_macbf}}, and MARL~\mbox{\cite{yu2022surprising_mappo, liu2020pic}} methods (InfoMARL, MAPPO, PIC) show an exponential increase in the number of input constraint violations. We refer the reader to~\mbox{\cite{zinage2024decentralized}} for additional details.

\subsection{Effect of Velocity Projection}

In Figure~\ref{fig: reconcile_30}, we plot the liveness values obtained via Definition~\ref{def: liveness_function} with velocity transformation (\textcolor{blue}{blue}), Definition~\ref{def: liveness_function} without velocity transformation  (\textcolor{ForestGreen}{green}), and Theorem~\ref{thm: smg->l<tau} (\textcolor{red}{red}) for $\theta = \frac{\pi}{6}$. Similarly, in Figure~\ref{fig: reconcile_60}, we plot the liveness values obtained via Definition~\ref{def: liveness_function} with velocity transformation (\textcolor{blue}{blue}), Definition~\ref{def: liveness_function} without velocity transformation (\textcolor{ForestGreen}{green}), and Theorem~\ref{thm: smg->l<tau} (\textcolor{red}{red}) for $\theta = \frac{\pi}{3}$.

\begin{figure}[t]
    \centering
    \includegraphics[width=\linewidth]{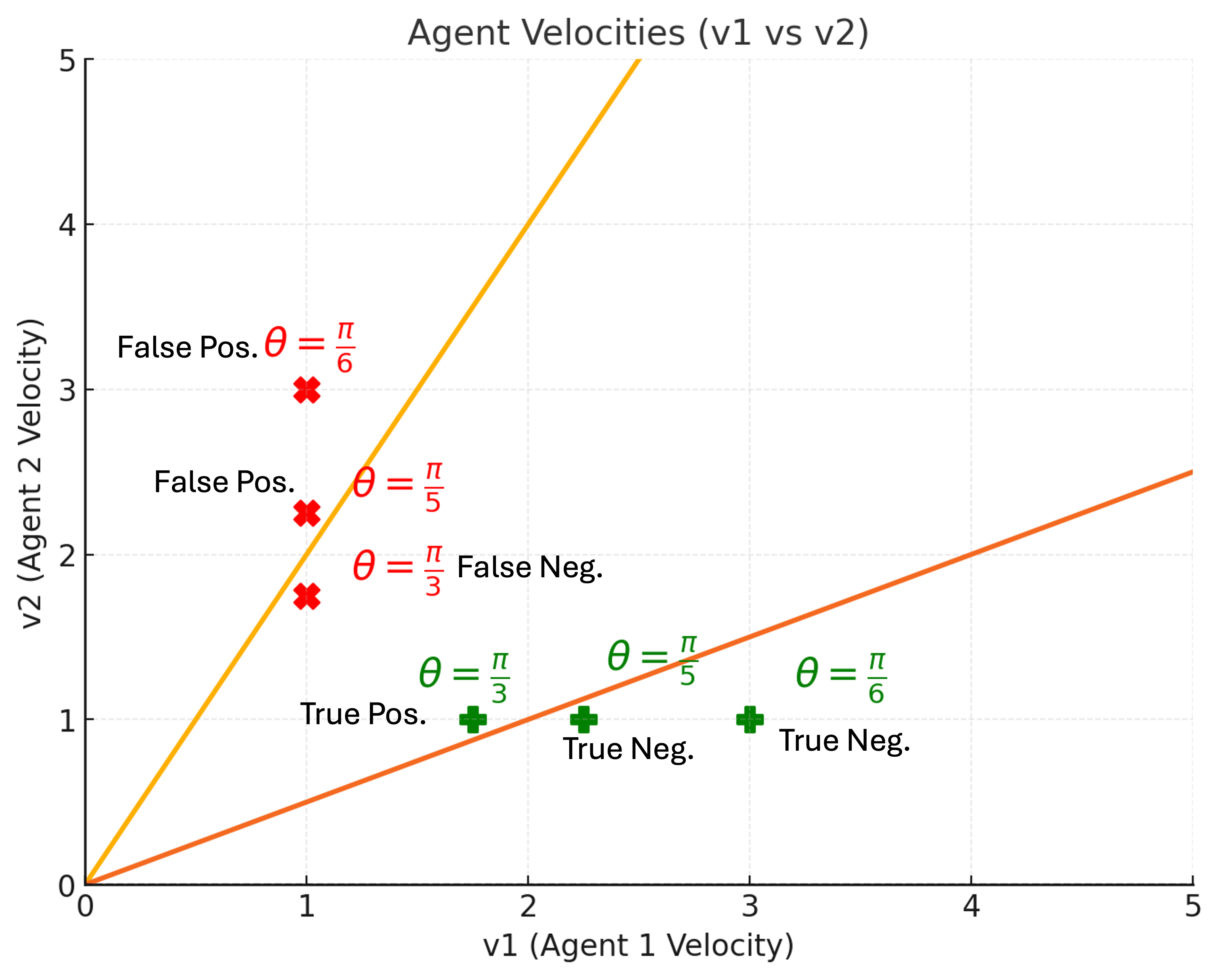}
    \caption{Effect of velocity projection: the red trials correspond to false negatives and false positives of an SMG occurrence using Definition~\ref{def: liveness_function} without velocity projection, whereas the green trials correspond to true positives and true negatives of an SMG occurrence using Definition~\ref{def: liveness_function} with velocity projection.}
    \label{fig: empirical}
\end{figure}

We ran additional experiments where we tested our approach in SMGs where the agents made the following angles with the relative poistion vector: $\left \{\frac{\pi}{3}, \frac{\pi}{6}, \frac{\pi}{5}\right\}$. In Figure~\ref{fig: empirical}, we plot trials conducted with varying speed scaling factors (each ($v1, v2$) point characterizes a speed scaling) and angle. A green ``\textcolor{ForestGreen}{$\textbf{+}$}'' trial indicates when an SMG occurrence was correctly identified (true positives and true negatives) using velocity transformation whereas a red ``\textcolor{red}{\textbf{x}}'' indicates when an SMG was incorrectly identified (false positives and false negatives) without velocity transformation. 

For example, of the three green points, Definition~\ref{def: liveness_function} with velocity transformation identifies a true positive occurrence of an SMG at v1=1.75 m/s and v2=1 m/s ($\zeta = 1.75, \theta = \frac{\pi}{3}$) and two true negative SMGs at v1=2.25 m/s and v2=1 m/s ($\zeta = 2.25, \theta = \frac{\pi}{5}$) and v1=3 m/s and v2=1 m/s ($\zeta = 3, \theta = \frac{\pi}{6}$), respectively. On the other hand, of the three red points, Definition~\ref{def: liveness_function} without velocity transformation identifies a false negative occurrence of an SMG at v2=1.75 m/s and v1=1 m/s ($\zeta = 1.75, \theta = \frac{\pi}{3}$) and two false positive SMGs at v2=2.25 m/s and v1=1 m/s ($\zeta = 2.25, \theta = \frac{\pi}{5}$) and v2=3 m/s and v1=1 m/s ($\zeta = 3, \theta = \frac{\pi}{6}$), respectively.

\section{Conclusion, Limitations, and Future Work}
\label{sec: conc}
In this work, we presented an approach to address the challenges of safe and deadlock-free navigation in decentralized, non-cooperative, and non-communicating multi-robot systems in constrained environments. We introduced the notion of \emph{social mini-games} to formalize the multi-robot navigation problem in constrained spaces. Our navigation approach uses a combined MPC-CBF formulation for optimal and safe long-horizon planning. Our approach to ensuring liveness rests on the insight that \textit{there exists barrier certificates that allow each robot to preemptively perturb their state in a minimally-invasive fashion onto liveness sets i.e. states where robots are deadlock-free}. 

There are a few limitations of our work. First, we only guarantee safety and liveness for asynchronously performing robots at a local level. And second, the current approach is applicable only in certain geometries such as doorways, hallways, intersections. Some other geometries that our approach can handle, but have not been studied in this paper, include L-shaped corners and corridors, T-shaped junctions, blind corners, and roundabouts. For future work, we plan on investigating the following open questions: $(i)$ investigating the integration of machine learning techniques to enhance decision-making capabilities, adaptability of the decentralized controllers, and extending scalability to multiple agents $(ii)$ establishing theoretical connections between game-theoretic solutions (\textit{e.g.} Nash equilibrium) and our control-theoretic solution to establish global optimality, $(iii)$ extending our solution to a broader and more general class of geometries, and $(iv)$ investigating robust control methods to establish a bound on $\zeta$ to address perception or sensor errors in observing neighbors' velocities.

\section*{Declarations}

\subsection*{Ethical Considerations and Conflicts of Interest}
This work was conducted in full accordance with the principles highlighted in the Committee on Publication Ethics. Our experiments involving cameras, lidars and any other sensors did not capture or record identifiable information of other humans. Lastly, the authors declare that there are no financial or personal conflicts of interest regarding the publication of this paper.

\subsection*{Data Availability}
This work primarily used open source software, data, and tools that are available at \href{https://github.com/ut-amrl}{\textbf{https://github.com/ut-amrl}}. The data and software produced by this work are available at \href{https://github.com/Vrushabh27/mpc-cbf}{\textbf{https://github.com/Vrushabh27/mpc-cbf}}.

\subsection*{Funding}
This work has taken place in the Autonomous Mobile Robotics Laboratory (AMRL) and the Learning Agents Research Group (LARG) at UT Austin. AMRL research in this work is supported by NSF (CAREER-2046955). LARG research is supported in part by NSF (FAIN-2019844, NRT-2125858), ONR (N00014-18-2243), ARO (E2061621), Bosch, Lockheed Martin, and UT Austin's Good Systems grand challenge. Peter Stone serves as the Executive Director of Sony AI America and receives financial compensation for this work. The terms of this arrangement have been reviewed and approved by the University of Texas at Austin in accordance with its policy on objectivity in research. The second and third authors acknowledge support from NSF (CMMI-1937957).

\subsection*{Author Contributions}
R.C wrote the main manuscript text, conducted experiments, prepared figures, tables, and theoretical results. V.Z. and E.B. reviewed the paper, prepared theoretical results, and wrote sections of the main paper. J.B. and P.S. reviewed the paper, provided guidance, direction, mentorship, and provided comments.

% \footnotesize{
\bibliography{refs}
\bibliographystyle{ieeetr}
% }

\end{document}